\documentclass{article}

\PassOptionsToPackage{numbers, compress}{natbib}
\usepackage[preprint]{neurips_2026}

 \usepackage{amsmath}
 \usepackage{booktabs}
 \usepackage{multirow}
 \usepackage{makecell}
 \usepackage{graphicx}
 \usepackage{pgfplots}
 \usepackage{subcaption}
 \usepackage[table]{xcolor}
\pgfplotsset{compat=1.18}
\usepackage{algorithm}
\usepackage{algpseudocode}
\usepackage{amsmath}
\usepackage[font=small]{caption}

\usepackage[utf8]{inputenc} 
\usepackage[T1]{fontenc}    
\usepackage{hyperref}       
\usepackage{url}            
\usepackage{booktabs}       
\usepackage{amsfonts}       
\usepackage{nicefrac}       
\usepackage{microtype}      
\usepackage{xcolor}         
\usepackage{booktabs}
\usepackage{multirow}
\usepackage{makecell}
\usepackage{graphicx}

\usepackage[utf8]{inputenc} 
\usepackage[T1]{fontenc}    
\usepackage{hyperref}       
\usepackage{url}            
\usepackage{booktabs}       
\usepackage{amsfonts}       
\usepackage{nicefrac}       
\usepackage{microtype}      
\usepackage{xcolor}         
\usepackage{booktabs} 
\usepackage{microtype}
\usepackage{amsmath}
\usepackage{comment}
\usepackage{enumitem}
\usepackage{graphicx}
\usepackage[export]{adjustbox}[2011/08/13]
\usepackage{wrapfig}
\usepackage{caption}

\title{GeoSym127K: Scalable Symbolically-verifiable Synthesis for Multimodal Geometric Reasoning}

\newcommand\crule[1][black]{\textcolor{#1}{\rule{2mm}{2mm}}}

\definecolor{myred}{HTML}{ff0000}
\definecolor{mygreen}{HTML}{008000}
\definecolor{myblue}{HTML}{0000ff}
\definecolor{myorange}{HTML}{ffa500}
\definecolor{myviolet}{HTML}{ee82ee}
\definecolor{mycyan}{HTML}{00bcd4} 

\author{
Jinhao Jing$^{*\crule[myred]\crule[mygreen]}$
\qquad
Zheng Ma$^{*\crule[myred]}$
\qquad
Jinwei Liang$^{*\ddagger\crule[myred]}$
\qquad
Qiannian Zhao$^{\crule[mygreen]}$
\\[0.15cm]
\textbf{Shawn Chen}$^{\crule[myblue]}$
\qquad
\textbf{Jing Yang}$^{\crule[myorange]}$
\qquad
\textbf{Por Lip Yee}$^{\crule[myorange]}$
\qquad
\textbf{Prayag Tiwari}$^{\crule[mycyan]}$
\\[0.15cm]
\textbf{Jingjing Bai}$^{\crule[myviolet]}$
\qquad
\textbf{Benyou Wang}$^{\crule[mygreen]}$
\qquad
\textbf{Lewei Lu}$^{\dagger\crule[myred]}$
\qquad
\textbf{Zhan Su}$^{\dagger\crule[mycyan]}$
\\[0.25cm]
$^{\crule[myred]}$SenseTime Research
\qquad
$^{\crule[mygreen]}$The Chinese University of Hong Kong, Shenzhen
\\[0.05cm]
$^{\crule[myblue]}$University of California, Los Angeles
\qquad
$^{\crule[myorange]}$Universiti Malaya
\qquad
$^{\crule[myviolet]}$Peking University
\\[0.05cm]
$^{\crule[mycyan]}$School of Information Technology, Halmstad University
\\[0.1cm]
$^{\dagger}$Corresponding Author
\qquad
$^{*}$Equal Contribution
\qquad
$^{\ddagger}$Project Lead
\\[0.1cm]
\texttt{
jinhaojing@link.cuhk.edu.cn,
luotto@sensetime.com,
zhan.su@hh.se
}
}

\begin{document}

\maketitle

\vspace{-0.65cm}
\begin{abstract}

Large Multimodal Models (LMMs) often struggle with geometric reasoning due to visual hallucinations and a lack of mathematically precise Chain-of-Thought (CoT) data. To address this, we propose the \textbf{GeoSym Engine}, an automated and scalable neuro-symbolic framework. By leveraging a type-conditional grammar and an analytic \textbf{SymGT Solver}, it derives exact symbolic ground truths and seamlessly integrates with a robust rendering pipeline to produce high-precision geometric diagrams. Using this engine, we construct \textbf{GeoSym127K}, a difficulty-stratified dataset featuring 51K high-resolution images, 127K questions with symbolic ground truths, and 55K answer-verified CoT QA pairs. We also introduce \textbf{GeoSym-Bench}, an expert-curated suite of 511 complex samples for rigorous evaluation. Through extensive supervised fine-tuning (SFT), we demonstrate that GeoSym drives concentrated improvements specifically on diagram-dependent and multi-step geometry tasks. Our Qwen3-VL-8B model gains an absolute \textbf{+22.21\%} on the MathVerse Vision-Only subset and reaches \textbf{61.52\%} (\textbf{+6.19\%} improvement) on WeMath, mitigating long-horizon logic fragmentation and outperforming advanced closed-source models like Doubao-1.8. Furthermore, applying Reinforcement Learning with Verifiable Rewards (RLVR) via GRPO reveals that initializing from structural SFT checkpoints substantially elevates the performance ceiling over zero-shot RL. Driven by deterministic exact-match signals, this showcases the robust scaling potential of our verifiable reasoning synthesis. Datasets and code are available at \url{https://huggingface.co/datasets/Tomie0506/GeoSym127K} and \url{https://github.com/Tomie56/GeoSym127K}.

\end{abstract}

\begin{figure}[htbp]
    \centering
    \caption{\textbf{Conceptual overview of the GeoSym framework.} \textbf{(Left)} Current bottlenecks in multimodal geometry: visual hallucination, symbolic math bias, and multi-step degradation. \textbf{(Middle)} The synthesis pipeline generates precise diagrams, analytic ground truths (SymGT), and answer-verified CoTs via strict rejection sampling. \textbf{(Right)} The training paradigm combines SFT and Reinforcement Learning with Verifiable Rewards (RLVR) via GRPO, leveraging exact symbolic signals to boost reasoning.}
    \includegraphics[width=\linewidth]{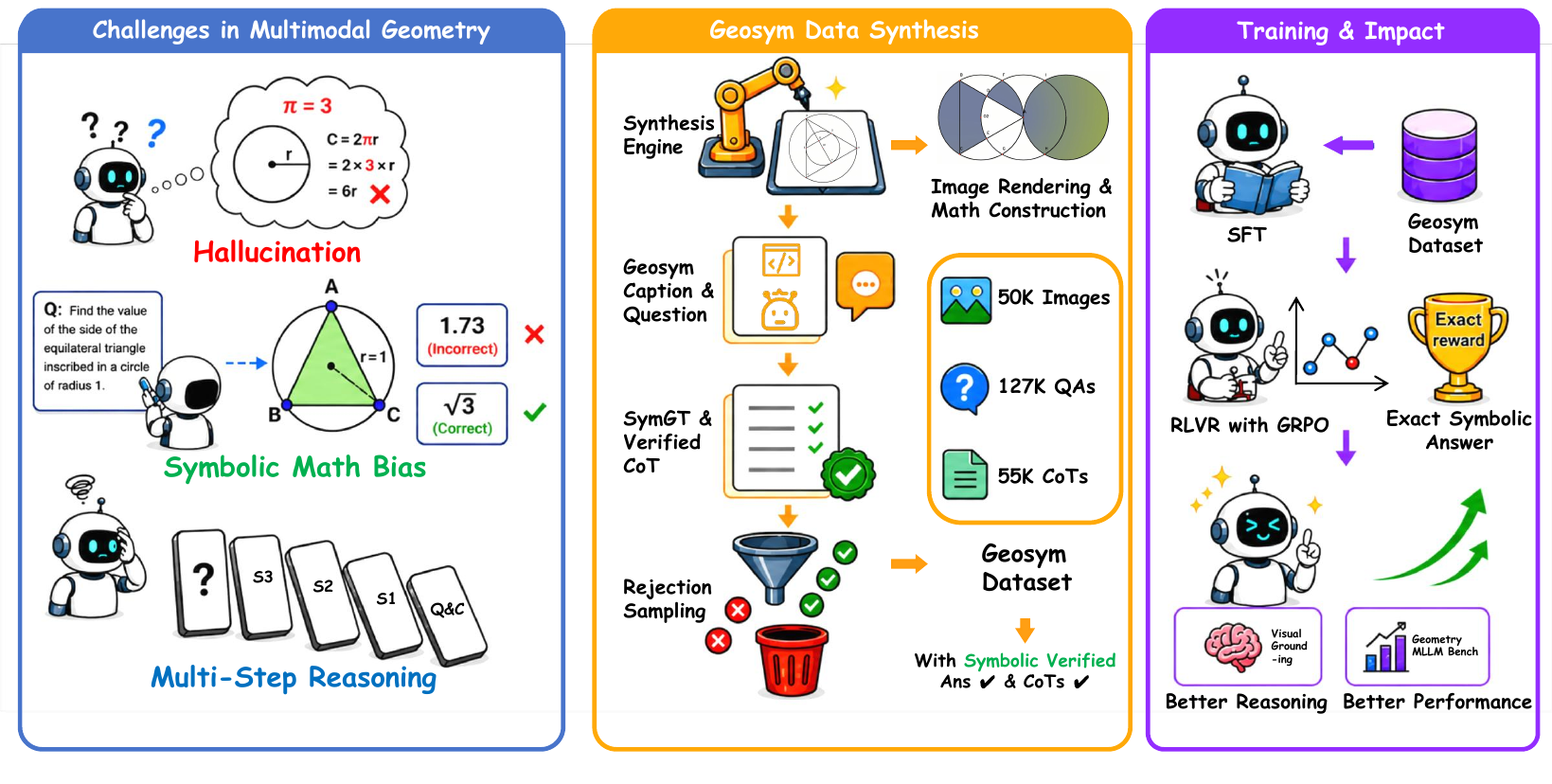} 
    \label{fig:teaser}
    
    \vspace{-1cm}
\end{figure}

\vspace{-0.45cm}
\section{Introduction}
\label{sec:intro}

Multimodal geometric reasoning is a representative touchstone for high-order mathematical intelligence~\cite{ICLR2025_2722a0cc, lu2021intergpsinterpretablegeometryproblem}, which requires Large Multimodal Models (LMMs) to move beyond generic visual-context question answering and execute fine-grained visual anchoring of topologies, rigorously bind these elements to mathematical theorems, and navigate exact logical paths with zero tolerance for numerical error \cite{peng-etal-2023-geodrl, feng2025geobenchrethinkingmultimodalgeometric}. 

While recent state-of-the-art LMMs have shown soaring scores on standard leaderboards \cite{ICLR2024_663bce02}, there are still some limitations\cite{ICLR2025_ec2e7a89, sun2025mathblindfailuresdiagram}. Existing geometry data pipelines largely depend on unreliable LLM-generated pseudo-labels \cite{10.1145/3688866.3689124} or heuristic templates \cite{lu2021intergpsinterpretablegeometryproblem, ICLR2025_db36dcad, deng2025theoremvalidatedreversechainofthoughtproblem}, failing to provide models with sufficiently coherent and formally verifiable supervision. A deeper analysis reveals three critical bottlenecks: visual hallucination during structural grounding~\cite{wang2025largelanguagemodelstruly, 10.1007/978-3-031-73242-3_10}, symbolic math bias caused by reliance on numerical approximations rather than exact mathematical representations~\cite{DBLP:journals/nature/TrinhWLHL24, lu2021intergpsinterpretablegeometryproblem, chen-etal-2022-unigeo}, and a catastrophic performance drop when problems necessitate multi-step deep deduction~\cite{10.1007/978-3-031-73242-3_10, qiao2024wemathdoeslargemultimodal}. This exposes a fundamental failure in precise representation and long-horizon coherence.

To break through these cognitive bottlenecks, we frame our investigation around a single, unified research question: \textbf{(RQ)} Can a massively scalable, symbolically verifiable synthesis paradigm—one that rigorously anchors every visual topology and intermediate logical step to exact mathematical coordinates rather than LLM heuristics—fundamentally eradicate visual hallucinations and overcome the performance degradation inherent in complex, multi-hop geometric reasoning?

To answer these questions, we propose \textbf{GeoSym}, a symbolically verifiable neuro-symbolic synthesis engine and training paradigm. As illustrated in Figure~\ref{fig:teaser}, GeoSym fundamentally reimagines geometric reasoning by establishing a rigorous, closed-loop pipeline: \textit{Symbolic Constraint $\rightarrow$ Exact Analytic Answer $\rightarrow$ Answer-verified Chain-of-Thought (CoT)}\cite{NEURIPS2022_9d560961}. Specifically, the \textbf{GeoSym Engine} consists of: (1) a \textbf{dynamic geometric environment} that evolves topologies upon an arbitrary-precision symbolic manifold; (2) a \textbf{visual-first rendering pipeline} that establishes a rigorous mapping from mathematical coordinates to pixel space, ensuring precise diagram synthesis and grounding for complex geometric elements, including shaded regions; and (3) an analytic \textbf{SymGT Solver} to derive absolute ground truths and reject hallucinated reasoning. Building upon this verifiable engine, we implement a RLVR paradigm using Group Relative Policy Optimization \cite{Guo_2025, shao2024deepseekmathpushinglimitsmathematical}, utilizing deterministic symbolic signals to force policy self-correction. We summarize our core contributions as follows:
\vspace{-0.1cm}
\begin{itemize}
    \item \textbf{Verifiable Synthesis Engine.} We develop the GeoSym engine, a framework that integrates symbolic manifolds with precision-aligned rendering (including complex shaded regions) and an analytic solver. This system effectively minimizes numerical inaccuracies and visual inconsistencies often found in synthetic geometric data.
\vspace{-0.05cm}
    \item \textbf{The GeoSym127K Ecosystem.} We introduce a large-scale, solver-verified ecosystem comprising 127K QA pairs, circumventing the noise inherent in LLM-generated annotations. This suite includes 51K captioning samples for visual alignment, 55K difficulty-stratified SFT pairs, and 20K samples tailored for RLVR, supplemented by an expert-curated 511-sample evaluation benchmark, \textbf{GeoSym-Bench}.
    
\vspace{-0.05cm}
    \item \textbf{Empirical Gains in Multi-Step Reasoning and RL.} Our evaluations demonstrate that GeoSym-driven SFT yields consistent improvements in diagram-dependent and multi-hop reasoning tasks. Furthermore, we show that RLVR leverages deterministic signals to elevate the performance ceiling, addressing the logical fragmentation observed in existing LMMs.
\end{itemize}

\begin{table*}[!b]
    \vspace{-0.2cm}
    \centering
    
    \resizebox{\linewidth}{!}{
    \begin{tabular}{@{} l c c c c c c c c @{}}
    \toprule
    \textbf{Dataset / Engine} & \textbf{Scale} & \textbf{Complex Topology} & \textbf{Area Question} & \textbf{Symbolic GT} & \textbf{Label Source} & \textbf{CoT} & \textbf{Diff. Control} & \textbf{Diagram Format} \\ \midrule
    
    \multicolumn{9}{c}{\textit{Manual Annotation \& Enhanced Real-World Datasets}} \\ \midrule
    Geometry3K \cite{lu2021intergpsinterpretablegeometryproblem} & 3K & -- & \checkmark & \checkmark & Human & $\times$ & $\times$ & Real-world \\
    GeoQA \cite{chen-etal-2021-geoqa} & 3.5K & -- & \checkmark & $\times$ & Human & $\times$ & $\times$ & Real-world \\
    G-LLaVA \cite{10.1145/3688866.3689124} & 8.1K & -- & \checkmark & $\times$ & Human & $\times$ & $\times$ & Real-world \\ \midrule
    
    \multicolumn{9}{c}{\textit{Template \& Rule-Based Synthesis}} \\ \midrule
    MAVIS \cite{ICLR2025_db36dcad} & 800K & $\times$ & $\times$ & \checkmark & Solver & \checkmark & $\times$ & Synthetic \\
    TR-CoT \cite{deng2025theoremvalidatedreversechainofthoughtproblem} & 33K & $\times$ & $\times$ & $\times$ & LLM & \checkmark & \checkmark & Synthetic \\ \midrule
    
    \multicolumn{9}{c}{\textit{Formal Language \& SDF-Based Pipelines}} \\ \midrule
    AutoGeo \cite{huang2024autogeoautomatinggeometricimage} & 100K & $\times$ & $\times$ & $\times$ & LLM & $\times$ & $\times$ & Logical Clauses \\
    NeSyGeo \cite{wu2025nesygeoneurosymbolicframeworkmultimodal} & 85.3K & $\times$ & $\times$ & $\times$ & LLM & \checkmark & \checkmark & Logical Clauses \\
    GeoFM \cite{zhang2025geofmenhancinggeometricreasoning} & -- & \checkmark & $\times$ & \checkmark & Solver & $\times$ & $\times$ & Logical Clauses \\
    GeoSDF \cite{zhang2025geosdfplanegeometrydiagram} & -- & \checkmark & $\times$ & $\times$ & LLM & $\times$ & $\times$ & Logical Clauses \\
    TrustGeoGen \cite{fu2026trustgeogenformalverifieddataengine} & 2.8K & \checkmark & $\times$ & \checkmark & Solver & \checkmark & \checkmark & Logical Clauses \\ \midrule
    
    \multicolumn{9}{c}{\textit{MLLM-Generated Code Synthesis}} \\ \midrule
    GeoGPT4V \cite{cai2024geogpt4vgeometricmultimodallarge} & 10K & $\times$ & $\times$ & $\times$ & LLM & $\times$ & $\times$ & Wolfram Code \\ \midrule
    
    \multicolumn{9}{c}{\textbf{\textit{The GeoSym Paradigm (Ours)}}} \\ \midrule
    \textbf{GeoSym (Ours)} & \textbf{127} & \textbf{\checkmark} & \textbf{\checkmark} & \textbf{\checkmark} & \textbf{Solver} & \textbf{\checkmark} & \textbf{\checkmark} & \textbf{Synthetic} \\ \bottomrule
    \end{tabular}
    
    }
    \caption{\textbf{Comprehensive Comparison of Multimodal Geometry Datasets and Synthesis Engines.} GeoSym uniquely achieves a complete feature set, combining high-complexity topological generation, precise shaded area processing, analytic ground truths (SymGT), and answer-verified CoT with rigorous difficulty stratification. Detailed visual comparisons of dataset instances are provided in Appendix~\ref{app:samples_and_comparison}. (\checkmark indicates supported, $\times$ indicates not supported or missing, -- indicates not applicable).}
    \label{tab:dataset_comparison}
\end{table*}

\section{Related Work}
\label{sec:related}

\textbf{Manual Annotation and LLM/MLLM Generation.} Manual datasets (e.g., Geometry3K \cite{lu2021intergpsinterpretablegeometryproblem}, GeoQA \cite{chen-etal-2021-geoqa}) offer natural phrasing but lack the scalability required for high-complexity, multi-hop reasoning. To scale, methods like G-LLaVA \cite{10.1145/3688866.3689124} utilize LLMs to synthesize text-based reasoning trajectories. Taking a different approach, GeoGPT4V \cite{cai2024geogpt4vgeometricmultimodallarge} employs Multimodal LLMs (MLLMs) to generate executable Wolfram code for geometric image and data synthesis. However, both trajectories remain fundamentally unverifiable: purely text-based LLM generation injects latent logical hallucinations, while MLLM code synthesis struggles to consistently guarantee mathematical exactness and structural stability when dealing with highly complex or overlapping topologies.

\textbf{Formal Language and SDF-Based Pipelines.} Pioneering systems like AlphaGeometry~\cite{DBLP:journals/nature/TrinhWLHL24} have demonstrated profound theorem-proving capabilities using formal logical clauses; however, they operate exclusively in the symbolic text domain and fundamentally lack multimodal visual grounding. To bridge this gap, recent neuro-symbolic methods utilize formal representations (e.g., AutoGeo \cite{huang2024autogeoautomatinggeometricimage}, NeSyGeo \cite{wu2025nesygeoneurosymbolicframeworkmultimodal}) or Signed Distance Fields (e.g., GeoSDF \cite{zhang2025geosdfplanegeometrydiagram}) to improve image-math alignment. However, SDF-based methods struggle to render highly complex compound geometries, while frameworks like NeSyGeo still rely on LLMs for final answers, risking pseudo-label errors. Conversely, although TrustGeoGen \cite{fu2026trustgeogenformalverifieddataengine} achieves full-chain formal verification, its rigid logical clause generation is computationally inefficient and yields low-quality, unnatural diagrams. More fundamentally, these formal language systems inherently struggle to model and analytically compute complex overlapping area relationships.

\textbf{Template-Based and Rule-Driven Synthesis.} Approaches like MAVIS \cite{ICLR2025_db36dcad} and TR-CoT \cite{deng2025theoremvalidatedreversechainofthoughtproblem} employ rule-based engines and heuristic text templates. However, rigid manual engines struggle to generate diverse or complex topological variants (e.g., dynamic circumcircles or arbitrary shaded areas). Furthermore, their reliance on rigid templates restricts the linguistic and structural diversity necessary for robust model generalization. In conclusion, current data construction paradigms can be broadly categorized into three trajectories, each facing inherent limitations as summarized in Table~\ref{tab:dataset_comparison}.

\textbf{Verifiable RL for Reasoning.}
As highlighted by GSM-Symbolic \cite{ICLR2025_ec2e7a89}, the reliance of LLMs on pattern replication rather than genuine reasoning leads to extreme fragility under numerical or structural variations, underscoring the critical necessity of symbolic verifiability to prevent reward hacking\cite{NEURIPS2022_3d719fee} in Reinforcement Learning. Currently, RL has catalyzed breakthroughs in mathematical domains, with DeepSeekMath \cite{shao2024deepseekmathpushinglimitsmathematical} and WizardMath \cite{luo2025wizardmathempoweringmathematicalreasoning} pioneering process supervision in text-only mathematics. In multimodal contexts, Vision-R1 \cite{huang2026visionr1incentivizingreasoningcapability} demonstrated that RL requires high-quality cold-start CoTs to activate complex reasoning. While powerful, these works typically rely on LLM-based reward models (RMs) prone to reward hacking in geometrically ambiguous scenarios.

\section{Methodology: Scalable and Symbolically-Verifiable Synthesis}
\label{sec:methodology}

Landmark systems such as \textit{AlphaGeometry}~\cite{DBLP:journals/nature/TrinhWLHL24} have established that complex mathematical deduction requires a neuro-symbolic alliance: pairing a neural model's intuitive pattern recognition with a symbolic engine's rigorous derivation. Extending this paradigm to the multimodal domain, however, introduces unique challenges, as models must precisely align unstructured pixels with strict geometric theorems. 

\textbf{Necessity of Analytic Solvers.} 
Recent analytical work, such as \textit{GSM-Symbolic}~\cite{ICLR2025_ec2e7a89}, exposes the severe limitations of current LMMs: operating primarily as empirical pattern matchers, they suffer catastrophic failures upon minor numerical or topological perturbations. This fragility is exacerbated by traditional scaling paradigms that rely on LLMs to stochastically generate pseudo-labels or reasoning chains, which inevitably injects logical hallucinations. True geometric reasoning cannot be scaled through probabilistic text generation; it inherently relies on powerful mathematical parsers. To instill genuine geometric intelligence, dataset scaling must shift to \textit{correct-by-construction} generation, where every visual state and reasoning step is strictly derived and verified by a deterministic analytic solver.

\textbf{The Guiding Philosophy.} 
Synthesizing these motivations, we propose the philosophy of \textbf{Symbolically-Verifiable Synthesis}. This principle dictates that all generated geometric topologies, visual renderings, and textual reasoning trajectories must not be stochastic creations, but rather strict dual projections of a shared, arbitrary-precision mathematical manifold. By replacing LLM-generated heuristics with mathematical truths, this approach naturally yields the massive, flawless Chain-of-Thought (CoT) trajectories necessary for robust Supervised Fine-Tuning (SFT), while simultaneously providing the exact-match, deterministic reward signals required to safely drive Reinforcement Learning (RLVR) without reward hacking.

\section{The GeoSym Synthesis Framework}
\label{sec:method}

\subsection{Symbolic Geometric Manifold}
\label{subsec:manifold}

To eradicate the cumulative floating-point errors inherent in conventional synthesis, we define the geometric environment as an arbitrary-precision state space $\mathcal{G} = \langle \mathcal{P}, \mathcal{E}, \Phi, \mathcal{L}, \mathcal{T} \rangle$. Specifically, we enforce a strict \textit{Atomicity} principle where the point set $\mathcal{P}$ serves as the sole atoms of the manifold; higher-order entities $\mathcal{E}$ (e.g., segments, arcs) maintain only topological references to $\mathcal{P}$. The coordinate system $\Phi$ is strictly attached to these atomic points, with spatial values $(x, y)$ maintained as analytic expression trees via SymPy to ensure absolute mathematical precision during complex transformations. Furthermore, the system quantifies the logical depth $\mathcal{L}$ of the data, where derived entities are assigned a level of $\max(\mathcal{L}_{parents}) + 1$. This structural information, paired with the ordered generative trajectory $\mathcal{T}$, serves as the logical backbone for subsequent Chain-of-Thought (CoT) synthesis.

\begin{figure}[t]
    \centering
    \caption{\textbf{Overview of the GeoSym Synthesis Framework.} \textbf{Left:} The \textit{Symbolic Manifold} anchors analytic expressions ($\Phi$) to atomic points ($\mathcal{P}$) for arbitrary precision. \textbf{Right:} A strictly verified 4-stage pipeline (\textbf{Builder}, \textbf{Drawer}, \textbf{GT Solver}, \textbf{Generator}) evolves complex topologies, bridges abstract entities to visual pixels via CCA, derives SymPy metrics, and ensures structural integrity via $\text{Simplify}(A_{pred} - A_{GT}) \equiv 0$ CoT verification.}
    \label{fig:pipeline} 
    \vspace{-0.1cm}
    \includegraphics[width=\linewidth]{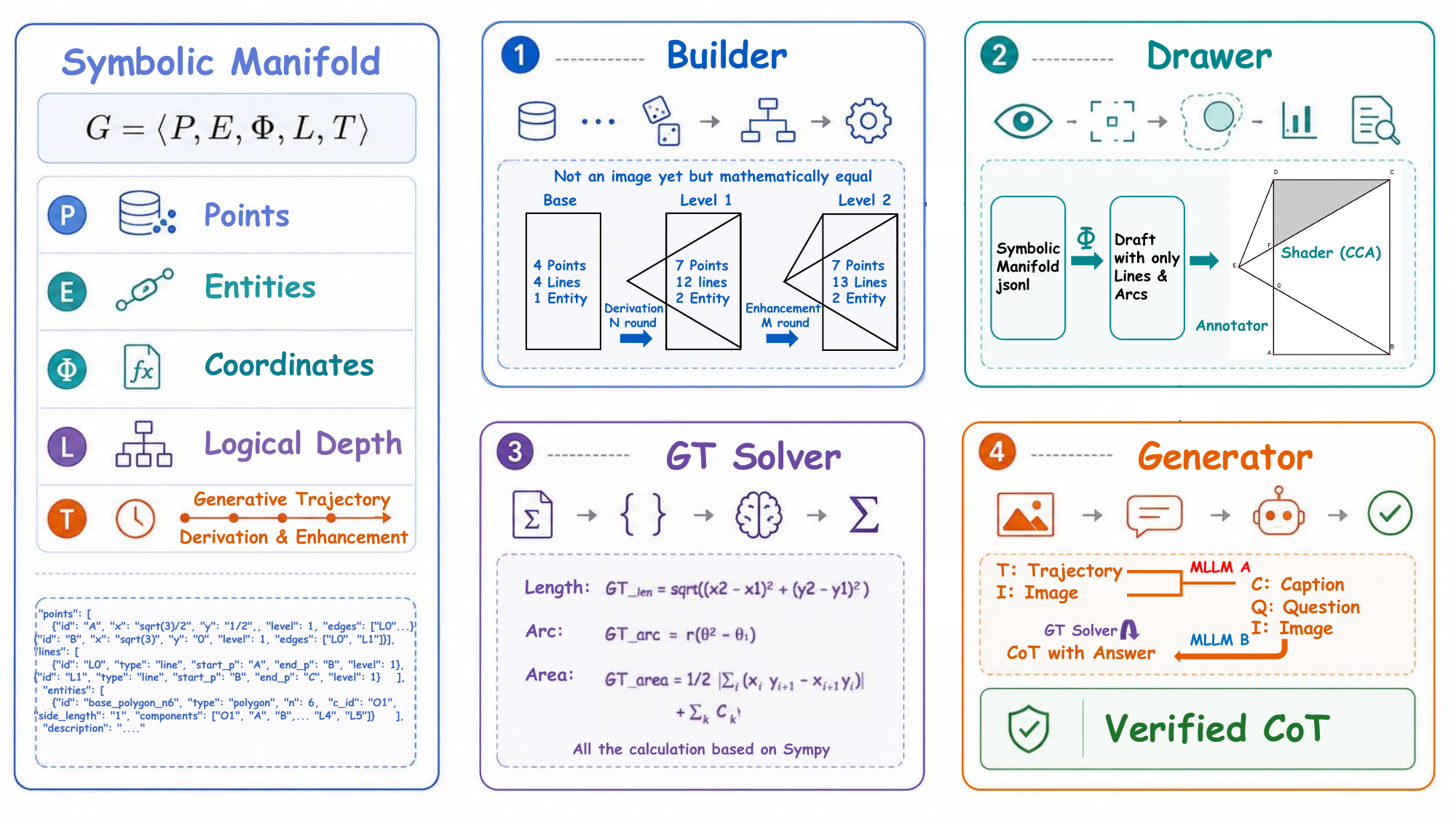} 
    
    \vspace{-0.2cm}
\end{figure}

\subsection{The GeoSym Synthesis Pipeline}
\label{subsec:geosym_pipeline}

The GeoSym pipeline instantiates a rigorous closed-loop from symbolic manifolds to natural language instructions, ensuring absolute mathematical integrity through four key stages, as illustrated in Figure~\ref{fig:pipeline} and algorithm details are provided in Appendix~\ref{app:Algorithm}.

\textbf{Type-Conditional Topological Evolution.} Geometric construction is modeled as a sequential decision process governed by a type-conditional probabilistic grammar (\ref{app:grammar_spec}). The sampling of evolutionary operators $\mathcal{OP}$ is strictly conditioned on parent entity types (e.g., circular bases favor concentric scaling), ensuring diverse yet intuitive topologies. To simulate human drafting, an integrated \textit{Builder} module executes auxiliary constructions (e.g., perpendiculars) and analytically instantiates intersections as new atoms by solving algebraic equations, maintaining strict manifold closure (\ref{alg:generation_process}).

\textbf{Visual-First Grounding and Alignment.} To resolve the challenge of abstract area deduction, we apply Connected Component Analysis (CCA) \cite{ROSENFELD196833} to binarized line art to extract independent closed regions. These contours are strictly mapped back to symbolic entity sequences (e.g., ``Arc A + Segment B'') and subjected to a rigorous geometric closure check. Only regions reconstructible as mathematically self-consistent symbolic loops are instantiated as \textit{Shaded Block} entities, ensuring every visual region possesses an exact symbolic definition (\ref{app:rendering_algo}).

\textbf{SymGT Solver and Task Formulation.} We implement \textit{Tail-Biased Querying} to target entities with higher dependency values, forcing models to implicitly backtrack the generative trajectory and inducing multi-step reasoning. Ground truths are derived via algebraic expressions to avoid numerical approximation. For regions bounded by mixed curves, we introduce a \textit{Generalized Symbolic Shoelace Algorithm} that decomposes area calculations into rectilinear polygonal baselines and non-linear topological compensations based on arc winding directions (\ref{app:area_solver}).

\textbf{Instruction Synthesis and Logical Verification.} We implement a \textit{Generate-and-Verify} pipeline to ensure cross-modal consistency. First, a teacher MLLM translates the generative trajectory $\mathcal{T}$ into a \textit{GeoSym-Caption}, establishing a precise structural description. Subsequently, the teacher generates CoT rationales using the \textbf{image, caption, and question} as joint context(\ref{app:prompt_templates}). To minimize hallucinations, we apply a symbolic filter via SymPy: only samples where the predicted answer $A_{pred}$ satisfies $\text{Simplify}(A_{pred} - A_{GT}) \equiv 0$ are retained. This deterministic verification ensures high-fidelity reasoning across (Image, Question, CoT) triplets (\ref{app:Algorithm}).

\vspace{-0.1cm}
\section{The GeoSym Dataset and Benchmark}
\label{sec:dataset}

\subsection{The GeoSym127K Data Ecosystem}
\label{subsec:ecosystem}

We curate \textbf{GeoSym127K}, a solver-verified ecosystem of 127K QA pairs, circumventing the noise of LLM-based labeling. The dataset is fundamentally constructed using a \textit{Generation-Driven Complexity Stratification} framework: by manipulating hyperparameters such as maximum recursion depth (Appendix~\ref{app:config}), we synthesize geometric problems across three distinct difficulty tiers (\textit{Entry, Hard, Expert}), as detailed in Table~\ref{tab:macro_stats}. 

\begin{table}[b!]
\vspace{-0.5cm}
\centering
\small
\caption{\textbf{Macro-Tier Statistics and Tier-wise Characteristics.} }
\label{tab:macro_stats}
\begin{tabular}{@{} l r r r r l @{}}
\toprule
\textbf{Tier} & \textbf{States/Caption} & \textbf{QA Pairs} & \textbf{Ans-Verified CoTs} & \textbf{Pass Rate} & \textbf{Core Features} \\ \midrule
Entry  & 20,177 & 41,844 & 23,440 & 56.09\% & Basic reasoning, 1--2 steps \\
Hard   & 23,410 & 60,157 & 23,835 & 39.78\% & Nested topology, multi-hop \\
Expert & 7,893  & 25,363 & 8,302  & 32.73\% & Expert level, hard for MLLMs \\ \midrule
\textbf{All} & \textbf{51,480} & \textbf{127,364} & \textbf{55,577} & \textbf{43.64\%} & \textbf{--} \\ \bottomrule
\end{tabular}
\end{table}

\begin{figure*}[t!]
    \centering
    \includegraphics[width=0.95\linewidth]{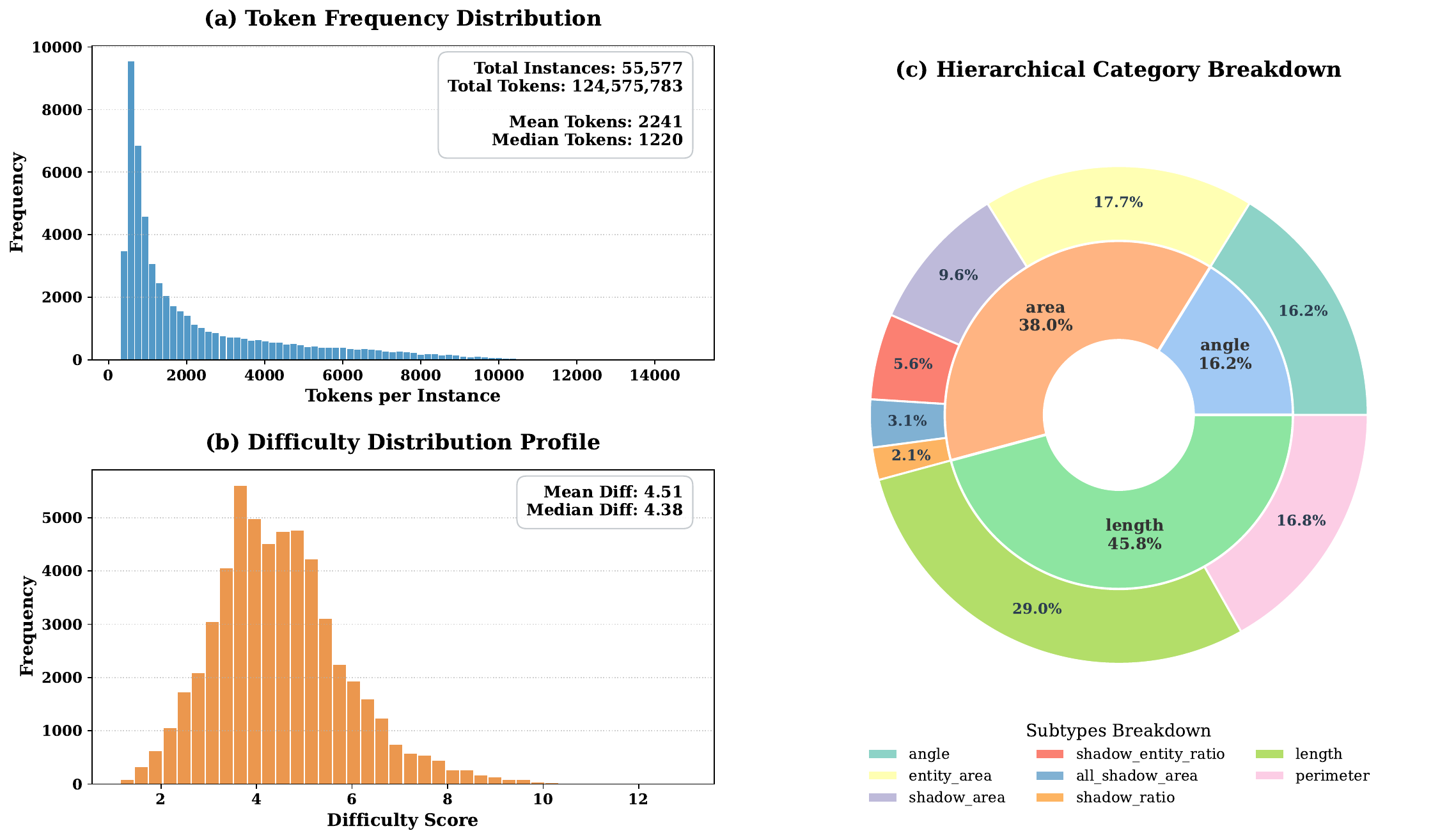}
    \caption{\textbf{GeoSym Instruct Dataset Overview.} \textbf{(a-b)} Distributions of total tokens per instance and difficulty scores, demonstrating the dataset's broad logical depth and text-rich reasoning chains. \textbf{(c)} A hierarchical nested ring chart illustrating the proportion of different geometric types (inner ring) and subtypes (outer ring), with core overall statistics embedded in the center.}
    \label{fig:dataset_overview}
    \vspace{-0.3cm}
\end{figure*}

From this stratified generative pool, only the instances that strictly pass our deterministic answer verification are curated into \textbf{GeoSym-Instruct-55K} for supervised fine-tuning, as illustrated in Figure~\ref{fig:dataset_overview}. The broader ecosystem further comprises \textbf{GeoSym-Caption-51K} for robust visual alignment, and \textbf{GeoSym-RL-20K} (10k Entry, 10k Hard), which is formatted exclusively with symbolic ground truths to provide deterministic rewards for policy optimization without SFT pool contamination. The ecosystem is finalized by \textbf{GeoSym-Bench}, an expert-curated 511-sample suite for high-order reasoning evaluation.

\begin{table*}[b!]
    \centering
    \begin{minipage}[t]{0.45\linewidth}
            \centering
            \caption{Human expert validation pass rates on a stratified random pool of 1,000 samples from GeoSym-127K.}
            \label{tab:expert_validation}
            \vspace{0.1cm} 
            \begin{tabular}{lc}
                \toprule
                \textbf{Validation Metric} & \textbf{Result} \\
                \midrule
                Audited Sample Volume        & 1,000 \\
                \midrule
                Topological Validity (Image) & 100.0\% \\
                Symbolic Ground Truth (Answer)  & 100.0\% \\
                Full CoT Derivation (CoT)    & 98.4\%  \\
                \bottomrule
            \end{tabular}
        \end{minipage}\hfill
    \begin{minipage}[t]{0.48\linewidth}
        \centering
        \caption{Baseline accuracy on GeoSym-Bench. Our 8B model with GeoSym significantly outperforms massive proprietary and open-weight LMMs.}
        \label{tab:geosym_bench_results}
        \vspace{0.1cm}
        \begin{tabular}{lc}
            \toprule
            \textbf{Model} & \textbf{Accuracy (\%)} \\
            \midrule
            Doubao-1.8\cite{seed2026seed18modelcardgeneralized} & 11.55 \\
            Qwen3-VL-235B\cite{bai2025qwen3vltechnicalreport} & 14.68 \\
            Gemini-3-Pro & 15.66 \\
            \midrule
            \textbf{Qwen3-VL-8B\cite{bai2025qwen3vltechnicalreport} + GeoSym} & \textbf{18.79} \\
            \bottomrule
        \end{tabular}
    \end{minipage}
\end{table*}

\subsection{GeoSym-Bench}
\label{subsec:bench}

To definitively establish the mathematical rigor of our synthetic ecosystem, a panel of human experts systematically audited a stratified random pool of 1,000 instances from GeoSym-127K. As summarized in Table~\ref{tab:expert_validation}, this evaluation confirmed a \textbf{100\% accuracy rate} for both topological validity and symbolic ground truths, alongside an exceptional \textbf{98.4\% pass rate} for MLLM-generated CoT rationales, empirically validating our generation pipeline. From this strictly verified pool, the experts meticulously curated 511 highly representative, error-free instances to construct \textbf{GeoSym-Bench}. Featuring extreme topological density, complex shaded regions, and competition-level multi-step logic, this benchmark serves as a definitive stress test for LMMs. Consequently, baseline evaluations in Table~\ref{tab:geosym_bench_results} demonstrate that while massive proprietary and open-weight models struggle on this benchmark, our GeoSym-enhanced 8B architecture achieves leading performance. A detailed analysis of minor CoT error modes and comprehensive baseline logs are further provided in Appendix~\ref{app:detailed_samples}.

\begin{table*}[h!]
    \centering
    \caption{\textbf{Main SFT Results} The table details the macro-average scores and specific sub-category breakdowns. The highest values within each parameter-scale comparison group are highlighted with a \colorbox{blue!10}{\textbf{light blue background and bold text}}. Asterisks ($^*$) indicate models evaluated via API calls using identical evaluation settings. The dagger ($^{\dagger}$) denotes a specific anomalous result observed for the 8B baseline on MathVerse Vision-only, with a detailed discussion deferred to Appendix~\ref{app:MathVerse}. The double dagger ($^{\ddagger}$) indicates selected open source baseline synthesis methods for fair comparison with details reported in Appendix~\ref{app:epoch_matching}. Note that our reported GeoSym scores represent the peak-performing training epoch; an ablation on epoch saturation is provided in Appendix~\ref{app:full_logs}}
    \label{tab:sft_main}
    \resizebox{\linewidth}{!}{
    \setlength{\tabcolsep}{4pt} 
    \begin{tabular}{@{} l c | c c | c c c c | c c c | c c c c c c @{}}
        \toprule
        \multirow{3}{*}{\textbf{Model \& Method}} & \multirow{3}{*}{\textbf{Overall}} & \multicolumn{2}{c|}{\makecell{\textbf{MathVista\cite{ICLR2024_663bce02}}\\1000}} & \multicolumn{4}{c|}{\makecell{\textbf{MathVerse\cite{10.1007/978-3-031-73242-3_10}} \textit{Vision only}\\788 (3940)}} & \multicolumn{3}{c|}{\makecell{\textbf{MathVision\cite{NEURIPS2024_ad0edc7d}}\\3040}} & \multicolumn{6}{c}{\makecell{\textbf{WeMath\cite{qiao2024wemathdoeslargemultimodal}}\\1740}} \\
        \cmidrule(lr){3-4} \cmidrule(lr){5-8} \cmidrule(lr){9-11} \cmidrule(lr){12-17}
        & & \makecell{geometry\\solving} & \makecell{geometry\\reasoning} & Angle & Length & Area & Plane & Angle & Area & Length & \makecell{Angles\\\& Length} & \makecell{Calc. of\\Plane} & \makecell{Under.\\of Plane} & \makecell{One-\\step} & \makecell{Two-\\step} & \makecell{Three-\\step} \\
        \cmidrule(lr){3-4} \cmidrule(lr){5-8} \cmidrule(lr){9-11} \cmidrule(lr){12-17}
        & & 208 & 239 & 193 & 182 & 91 & 510 & 173 & 500 & 449 & 34 & 340 & 256 & 1215 & 360 & 165 \\
        \midrule
        
        \multicolumn{17}{c}{\textbf{\textit{Closed-source LMMs}}} \\ 
        \midrule
        \textbf{Gemini-3-Pro}$^*$ & \cellcolor{blue!10}\textbf{79.51} & \multicolumn{2}{c|}{\cellcolor{blue!10}\textbf{88.50}} & \multicolumn{4}{c|}{\cellcolor{blue!10}\textbf{85.91}} & \multicolumn{3}{c|}{\cellcolor{blue!10}\textbf{83.36}} & \multicolumn{6}{c}{\cellcolor{blue!10}\textbf{73.81}} \\
        \textbf{GPT-5\cite{singh2026openaigpt5card}}$^*$ & 76.55 & \multicolumn{2}{c|}{81.90} & \multicolumn{4}{c|}{81.20} & \multicolumn{3}{c|}{72.00} & \multicolumn{6}{c}{71.10} \\
        \textbf{Doubao-Seed-1.8\cite{seed2026seed18modelcardgeneralized}}$^*$ & 73.42 & \multicolumn{2}{c|}{86.30} & \multicolumn{4}{c|}{82.49} & \multicolumn{3}{c|}{69.61} & \multicolumn{6}{c}{58.29} \\
        \midrule
        
        \multicolumn{17}{c}{\textbf{\textit{Open-source LMMs (Large Scale > 30B)}}} \\ 
        \midrule
        \textbf{Qwen3.5-397B-A17B\cite{qwen35blog}}$^*$ & \cellcolor{blue!10}\textbf{87.47} & \multicolumn{2}{c|}{\cellcolor{blue!10}\textbf{90.20}} & \multicolumn{4}{c|}{\cellcolor{blue!10}\textbf{86.93}} & \multicolumn{3}{c|}{\cellcolor{blue!10}\textbf{85.79}} & \multicolumn{6}{c}{\cellcolor{blue!10}\textbf{86.95}} \\
        \textbf{Qwen3-VL-235B-A22B\cite{bai2025qwen3vltechnicalreport}}$^*$ & 77.34 & \multicolumn{2}{c|}{84.90} & \multicolumn{4}{c|}{73.75} & \multicolumn{3}{c|}{75.00} & \multicolumn{6}{c}{75.71} \\
        \textbf{Qwen3-VL-30B-A3B\cite{bai2025qwen3vltechnicalreport}}$^*$ & 72.76 & \multicolumn{2}{c|}{81.20} & \multicolumn{4}{c|}{73.10} & \multicolumn{3}{c|}{67.70} & \multicolumn{6}{c}{69.05} \\
        \midrule
        
        \multicolumn{17}{c}{\textbf{\textit{Base Model: Qwen3-VL Series (8B \& 4B)}}} \\ 
        \midrule
        \multirow{2}{*}{\textbf{Qwen3VL-8B-instruct}} & \multirow{2}{*}{55.94} & \multicolumn{2}{c|}{75.80} & \multicolumn{4}{c|}{38.32$^{\dagger}$} & \multicolumn{3}{c|}{\cellcolor{blue!10}\textbf{54.54}} & \multicolumn{6}{c}{55.33} \\
        & & 87.50 & 85.77 & 36.27 & 40.66 & 25.27 & 38.04 & \cellcolor{blue!10}\textbf{67.05} & 59.80 & \cellcolor{blue!10}\textbf{69.49} & 39.12 & 85.50 & 77.20 & 79.84 & 71.11 & 64.24 \\
        \cmidrule(lr){1-17}
        \multirow{2}{*}{\makecell[l]{\quad + TR-GeoMM~\cite{deng2025theoremvalidatedreversechainofthoughtproblem}$^{\ddagger}$}} & \multirow{2}{*}{38.50} & \multicolumn{2}{c|}{61.80} & \multicolumn{4}{c|}{37.69} & \multicolumn{3}{c|}{24.11} & \multicolumn{6}{c}{30.38} \\
        & & 64.90 & 64.85 & 38.34 & 42.31 & 21.98 & 39.61 & 28.32 & 25.60 & 23.83 & 34.56 & 71.89 & 64.03 & 66.17 & 46.67 & 32.12 \\
        \cmidrule(lr){1-17}
        \multirow{2}{*}{\makecell[l]{\quad + GeoTrust-train~\cite{fu2026trustgeogenformalverifieddataengine}$^{\ddagger}$}} & \multirow{2}{*}{40.31} & \multicolumn{2}{c|}{63.20} & \multicolumn{4}{c|}{44.54} & \multicolumn{3}{c|}{26.25} & \multicolumn{6}{c}{32.19} \\
        & & 78.85 & 75.73 & 40.93 & 61.54 & 28.57 & 47.84 & 39.31 & 22.20 & 25.39 & 35.79 & 72.65 & 57.03 & 64.03 & 52.78 & 61.21 \\
        
        \cmidrule(lr){1-17}
        \multicolumn{17}{c}{\textit{\textbf{--- Our Methods ---}}} \\
        \cmidrule(lr){1-17}
        
        \multirow{2}{*}{\makecell[l]{\quad \textbf{+ GeoSym Entry}}} & \multirow{2}{*}{62.49} & \multicolumn{2}{c|}{\cellcolor{blue!10}\textbf{76.60}} & \multicolumn{4}{c|}{\cellcolor{blue!10}\textbf{60.53}} & \multicolumn{3}{c|}{53.47} & \multicolumn{6}{c}{59.33} \\
        & & 92.31 & 90.38 & 62.69 & \cellcolor{blue!10}\textbf{75.83} & \cellcolor{blue!10}\textbf{51.65} & 65.29 & 61.27 & 61.80 & 64.81 & 43.16 & 87.37 & 79.27 & 82.06 & 76.11 & 72.12 \\
        \cmidrule(lr){1-17}
        \multirow{2}{*}{\makecell[l]{\quad \textbf{+ GeoSym Hard}}} & \multirow{2}{*}{\cellcolor{blue!10}\textbf{63.18}} & \multicolumn{2}{c|}{\cellcolor{blue!10}\textbf{76.60}} & \multicolumn{4}{c|}{60.41} & \multicolumn{3}{c|}{54.21} & \multicolumn{6}{c}{\cellcolor{blue!10}\textbf{61.52}} \\
        & & \cellcolor{blue!10}\textbf{92.79} & \cellcolor{blue!10}\textbf{90.80} & \cellcolor{blue!10}\textbf{65.28} & 72.53 & \cellcolor{blue!10}\textbf{51.65} & \cellcolor{blue!10}\textbf{65.49} & 63.58 & \cellcolor{blue!10}\textbf{62.40} & \cellcolor{blue!10}\textbf{69.49} & \cellcolor{blue!10}\textbf{51.75} & \cellcolor{blue!10}\textbf{88.04} & \cellcolor{blue!10}\textbf{83.35} & \cellcolor{blue!10}\textbf{83.62} & \cellcolor{blue!10}\textbf{77.22} & \cellcolor{blue!10}\textbf{75.15} \\
        \midrule
        
        \multirow{2}{*}{\textbf{Qwen3VL-4B-Instruct}} & \multirow{2}{*}{52.39} & \multicolumn{2}{c|}{73.40} & \multicolumn{4}{c|}{30.33} & \multicolumn{3}{c|}{\cellcolor{blue!10}\textbf{51.09}} & \multicolumn{6}{c}{54.76} \\
        & & 83.65 & 82.43 & 31.09 & 32.42 & 15.38 & 30.00 & \cellcolor{blue!10}\textbf{66.47} & \cellcolor{blue!10}\textbf{57.60} & \cellcolor{blue!10}\textbf{65.92} & 36.49 & 85.26 & \cellcolor{blue!10}\textbf{76.12} & 78.77 & 68.61 & 63.03 \\
        \cmidrule(lr){1-17}
        \multirow{2}{*}{\makecell[l]{\quad \textbf{+ GeoSym Entry}}} & \multirow{2}{*}{55.15} & \multicolumn{2}{c|}{72.80} & \multicolumn{4}{c|}{46.07} & \multicolumn{3}{c|}{48.49} & \multicolumn{6}{c}{53.24} \\
        & & 87.02 & 84.52 & 45.08 & 59.34 & 37.36 & 49.22 & 60.12 & 54.60 & 61.47 & \cellcolor{blue!10}\textbf{42.46} & \cellcolor{blue!10}\textbf{86.52} & 75.81 & 78.68 & 67.78 & 66.06 \\
        \cmidrule(lr){1-17}
        \multirow{2}{*}{\makecell[l]{\quad \textbf{+ GeoSym Hard}}} & \multirow{2}{*}{\cellcolor{blue!10}\textbf{58.47}} & \multicolumn{2}{c|}{\cellcolor{blue!10}\textbf{74.80}} & \multicolumn{4}{c|}{\cellcolor{blue!10}\textbf{55.20}} & \multicolumn{3}{c|}{48.82} & \multicolumn{6}{c}{\cellcolor{blue!10}\textbf{55.05}} \\
        & & \cellcolor{blue!10}\textbf{88.94} & \cellcolor{blue!10}\textbf{86.19} & \cellcolor{blue!10}\textbf{60.10} & \cellcolor{blue!10}\textbf{66.48} & \cellcolor{blue!10}\textbf{46.15} & \cellcolor{blue!10}\textbf{61.18} & 57.80 & 54.80 & 62.58 & 37.19 & 85.34 & 75.23 & \cellcolor{blue!10}\textbf{79.34} & \cellcolor{blue!10}\textbf{70.83} & \cellcolor{blue!10}\textbf{67.27} \\
        \midrule
        
        \multicolumn{17}{c}{\textbf{\textit{Base Model: Qwen2.5-VL Series (7B \& 3B)}}} \\ 
        \midrule
        \multirow{2}{*}{\textbf{Qwen2.5VL-7B-Instruct}} & \multirow{2}{*}{39.19} & \multicolumn{2}{c|}{67.90} & \multicolumn{4}{c|}{38.07} & \multicolumn{3}{c|}{23.36} & \multicolumn{6}{c}{27.43} \\
        & & 70.19 & \cellcolor{blue!10}\textbf{69.87} & 36.27 & 41.76 & 29.67 & 39.41 & 29.48 & 25.60 & 26.28 & 47.72 & 68.49 & 53.27 & 62.63 & 47.22 & 40.00 \\
        \cmidrule(lr){1-17}
        \multirow{2}{*}{\makecell[l]{\quad + TR-GeoMM$^{\ddagger}$}} & \multirow{2}{*}{32.82} & \multicolumn{2}{c|}{61.70} & \multicolumn{4}{c|}{27.03} & \multicolumn{3}{c|}{17.43} & \multicolumn{6}{c}{19.24} \\
        & & 53.85 & 54.81 & 30.57 & 24.18 & 20.88 & 28.43 & 16.76 & 17.00 & 16.70 & 38.60 & 58.75 & 54.89 & 53.58 & 33.61 & 18.79 \\
        \cmidrule(lr){1-17}
        \multirow{2}{*}{\makecell[l]{\quad + GeoTrust-train$^{\ddagger}$}} & \multirow{2}{*}{31.35} & \multicolumn{2}{c|}{63.10} & \multicolumn{4}{c|}{31.98} & \multicolumn{3}{c|}{18.98} & \multicolumn{6}{c}{17.24} \\
        & & 65.38 & 64.02 & 32.64 & 35.16 & 27.47 & 18.98 & 23.12 & 14.80 & 17.82 & 23.33 & 59.42 & 48.78 & 51.85 & 38.06 & 35.76 \\
        
        \cmidrule(lr){1-17}
        \multicolumn{17}{c}{\textit{\textbf{--- Our Methods ---}}} \\
        \cmidrule(lr){1-17}
        
        \multirow{2}{*}{\makecell[l]{\quad \textbf{+ GeoSym Entry}}} & \multirow{2}{*}{\cellcolor{blue!10}\textbf{44.23}} & \multicolumn{2}{c|}{68.50} & \multicolumn{4}{c|}{\cellcolor{blue!10}\textbf{44.41}} & \multicolumn{3}{c|}{\cellcolor{blue!10}\textbf{27.04}} & \multicolumn{6}{c}{36.95} \\
        & & 67.79 & 66.53 & \cellcolor{blue!10}\textbf{42.78} & \cellcolor{blue!10}\textbf{52.70} & \cellcolor{blue!10}\textbf{35.88} & \cellcolor{blue!10}\textbf{47.45} & \cellcolor{blue!10}\textbf{38.73} & \cellcolor{blue!10}\textbf{31.60} & \cellcolor{blue!10}\textbf{29.40} & 51.05 & 74.31 & 61.93 & \cellcolor{blue!10}\textbf{68.72} & 51.39 & 46.06 \\
        \cmidrule(lr){1-17}
        \multirow{2}{*}{\makecell[l]{\quad \textbf{+ GeoSym Hard}}} & \multirow{2}{*}{43.50} & \multicolumn{2}{c|}{\cellcolor{blue!10}\textbf{69.40}} & \multicolumn{4}{c|}{41.62} & \multicolumn{3}{c|}{25.66} & \multicolumn{6}{c}{\cellcolor{blue!10}\textbf{37.33}} \\
        & & \cellcolor{blue!10}\textbf{70.67} & 68.62 & 41.96 & 51.65 & 32.97 & 44.31 & 34.10 & 31.00 & 27.84 & \cellcolor{blue!10}\textbf{57.72} & \cellcolor{blue!10}\textbf{74.48} & \cellcolor{blue!10}\textbf{62.75} & 68.64 & \cellcolor{blue!10}\textbf{52.50} & \cellcolor{blue!10}\textbf{46.67} \\
        \midrule
        
        \multirow{2}{*}{\textbf{Qwen2.5VL-3B-Instruct}} & \multirow{2}{*}{29.34} & \multicolumn{2}{c|}{59.60} & \multicolumn{4}{c|}{25.63} & \multicolumn{3}{c|}{19.96} & \multicolumn{6}{c}{12.19} \\
        & & 50.48 & 50.63 & 26.42 & 26.37 & 15.38 & 26.47 & 23.70 & 21.40 & 24.50 & 30.53 & 50.20 & 38.14 & 44.28 & 27.22 & 25.45 \\
        \cmidrule(lr){1-17}
        \multirow{2}{*}{\makecell[l]{\quad \textbf{+ GeoSym Entry}}} & \multirow{2}{*}{\cellcolor{blue!10}\textbf{31.55}} & \multicolumn{2}{c|}{\cellcolor{blue!10}\textbf{60.10}} & \multicolumn{4}{c|}{\cellcolor{blue!10}\textbf{29.55}} & \multicolumn{3}{c|}{\cellcolor{blue!10}\textbf{20.40}} & \multicolumn{6}{c}{20.15} \\
        & & 53.84 & 54.18 & 29.79 & 29.67 & 18.68 & 28.72 & \cellcolor{blue!10}\textbf{24.60} & \cellcolor{blue!10}\textbf{22.00} & \cellcolor{blue!10}\textbf{24.80} & 34.82 & 57.70 & 43.87 & 49.01 & 31.39 & 26.66 \\
        \cmidrule(lr){1-17}
        \multirow{2}{*}{\makecell[l]{\quad \textbf{+ GeoSym Hard}}} & \multirow{2}{*}{30.20} & \multicolumn{2}{c|}{52.10} & \multicolumn{4}{c|}{28.43} & \multicolumn{3}{c|}{19.21} & \multicolumn{6}{c}{\cellcolor{blue!10}\textbf{21.05}} \\
        & & \cellcolor{blue!10}\textbf{57.21} & \cellcolor{blue!10}\textbf{57.74} & \cellcolor{blue!10}\textbf{33.16} & \cellcolor{blue!10}\textbf{32.97} & \cellcolor{blue!10}\textbf{21.98} & \cellcolor{blue!10}\textbf{30.98} & 21.97 & 21.20 & 20.04 & \cellcolor{blue!10}\textbf{39.12} & \cellcolor{blue!10}\textbf{65.20} & \cellcolor{blue!10}\textbf{49.60} & \cellcolor{blue!10}\textbf{53.74} & \cellcolor{blue!10}\textbf{35.56} & \cellcolor{blue!10}\textbf{27.88} \\
        \bottomrule
    \end{tabular}
    }
    \vspace{-0.3cm}
\end{table*}

\begin{figure*}[htbp]
    \centering
    \caption{\textbf{GRPO Training Dynamics across Different SFT Initializations.} 
    \textbf{(a) Average Reward} exhibits a steady and robust ascent across all configurations, confirming that the deterministic exact-match RLVR schema effectively guides policy optimization. 
    \textbf{(b) Response Length} indicates that the models actively explore and sustain extended Chain-of-Thought (CoT) reasoning to secure higher rewards, avoiding shortcut heuristic guessing. 
    \textbf{(c) Policy Entropy} displays a smooth and stable decay, illustrating a healthy transition from stochastic exploration to confident exploitation without suffering from premature mode collapse.}
    \label{fig:rl_training_curves}
    
    \includegraphics[width=\linewidth]{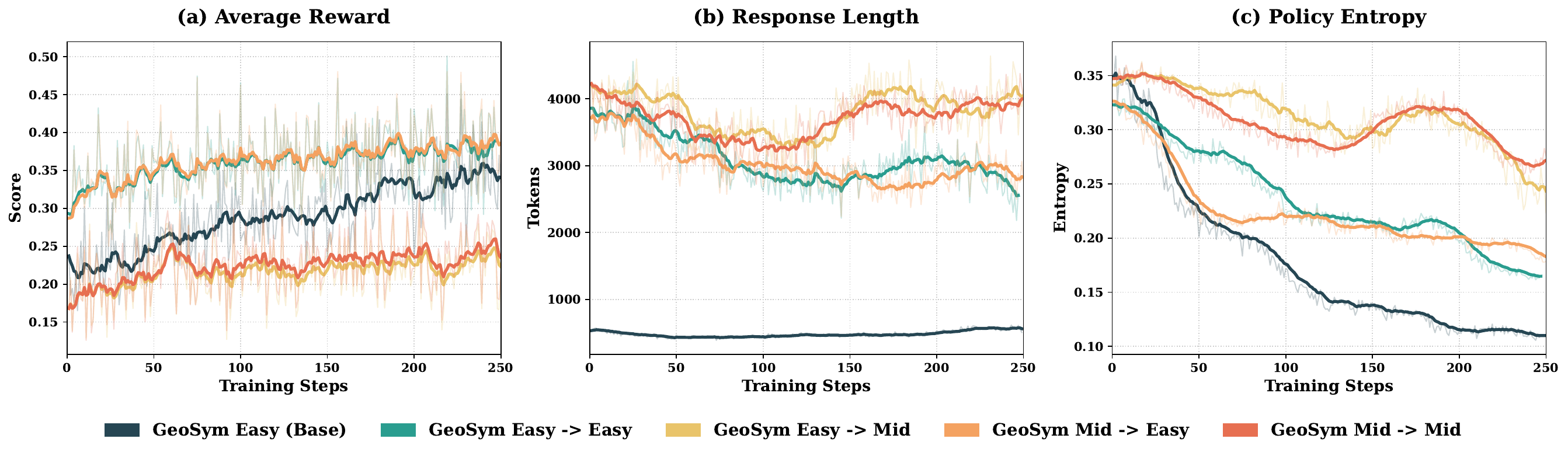} 
    
    \resizebox{\linewidth}{!}{
    \begin{tabular}{l | c c c c c}
        \toprule
        \textbf{Training Phase} & \makecell{\textbf{\hspace{4mm}Overall\hspace{4mm}}} & \textbf{MathVista} & \makecell{\textbf{MathVerse}\\ \textit{Vision only}} & \textbf{MathVision} & \textbf{WeMath} \\
        \midrule
        
        \multicolumn{6}{c}{\textit{Zero-shot GRPO (Directly on Base Model)}} \\
        \midrule
        \textbf{Qwen2.5-VL-7B-Instruct (Base)} & 39.19 & 67.90 & 38.07 & 23.36 & 27.43 \\
        \quad + GRPO-Entry & 42.60 {\scriptsize\textcolor{green!60!black}{(+3.41)}} & \textbf{70.40 {\scriptsize\textcolor{green!60!black}{(+2.50)}}} & 39.85 {\scriptsize\textcolor{green!60!black}{(+1.78)}} & 25.49 {\scriptsize\textcolor{green!60!black}{(+2.13)}} & 34.67 {\scriptsize\textcolor{green!60!black}{(+7.24)}} \\
        \midrule
        
        \multicolumn{6}{c}{\textit{GRPO on SFT Checkpoints (GeoSym Entry)}} \\
        \midrule
        \textbf{GeoSym Entry SFT} & 42.79 & 67.50 & 41.62 & 25.66 & 36.29 \\
        \quad + GRPO-Entry & 44.51 {\scriptsize\textcolor{green!60!black}{(+1.72)}} & 69.20 {\scriptsize\textcolor{green!60!black}{(+1.70)}} & 43.15 {\scriptsize\textcolor{green!60!black}{(+1.53)}} & 25.69 {\scriptsize\textcolor{green!60!black}{(+0.03)}} & \textbf{40.00 {\scriptsize\textcolor{green!60!black}{(+3.71)}}} \\
        \quad + GRPO-Hard & 43.59 {\scriptsize\textcolor{green!60!black}{(+0.80)}} & 69.70 {\scriptsize\textcolor{green!60!black}{(+2.20)}} & 41.62 & 25.69 {\scriptsize\textcolor{green!60!black}{(+0.03)}} & 37.33 {\scriptsize\textcolor{green!60!black}{(+1.04)}} \\
        \midrule
        
        \multicolumn{6}{c}{\textit{GRPO on SFT Checkpoints (GeoSym Hard)}} \\
        \midrule
        \textbf{GeoSym Hard SFT} & 43.71 & 68.60 & 42.51 & 25.63 & 38.48 \\
        \quad + GRPO-Entry & \textbf{44.99 {\scriptsize\textcolor{green!60!black}{(+1.28)}}} & \textbf{70.40 {\scriptsize\textcolor{green!60!black}{(+1.80)}}} & 41.50 {\scriptsize\textcolor{red}{(-1.01)}} & \textbf{28.45 {\scriptsize\textcolor{green!60!black}{(+2.82)}}} & 39.62 {\scriptsize\textcolor{green!60!black}{(+1.14)}} \\
        \quad + GRPO-Hard & 44.58 {\scriptsize\textcolor{green!60!black}{(+0.87)}} & 68.70 {\scriptsize\textcolor{green!60!black}{(+0.10)}} & \textbf{43.40 {\scriptsize\textcolor{green!60!black}{(+0.89)}}} & 26.41 {\scriptsize\textcolor{green!60!black}{(+0.78)}} & 39.81 {\scriptsize\textcolor{green!60!black}{(+1.33)}} \\
        \bottomrule
    \end{tabular}
    }
    
    \captionof{table}{\textbf{Impact of GRPO on Geometric Reasoning.} Relative improvements ($\uparrow$) and drops ($\downarrow$) are calculated against the respective preceding baseline (Base or SFT checkpoint). Note that the baseline SFT scores reported here differ slightly from those in Table~\ref{tab:sft_main}. To ensure a strictly fair evaluation of the GRPO stage, we utilize SFT checkpoints trained with an equivalent epoch and data volume rather than the absolute peak-performing epochs (detailed epoch comparisons explaining these variations are deferred to Table~\ref{tab:grpo_full_results} in Appendix~\ref{app:full_logs} . The highest scores across all configurations are highlighted in \textbf{bold}.}
    \label{tab:rl_main}
    \vspace{-0.5cm}
\end{figure*}

\section{Experiments and Analysis}
\label{sec:experiments}

We evaluate GeoSym's supervised fine-tuning (SFT) and GRPO alignment across diverse multimodal benchmarks. We build our models upon Qwen3-VL-8B-Instruct \cite{bai2025qwen3vltechnicalreport} and Qwen2.5-VL-7B-Instruct \cite{bai2025qwen25vltechnicalreport}, comparing them against closed-source models (e.g., Gemini-3-Pro, GPT-5\cite{singh2026openaigpt5card}), open-source LMMs, and state-of-the-art synthesis pipelines (TR-GeoMM, GeoTrust-train). All models are evaluated fairly via \textbf{VLMEvalKit}\cite{duan2025vlmevalkitopensourcetoolkitevaluating} on MathVista\cite{ICLR2024_663bce02}, MathVision\cite{NEURIPS2024_ad0edc7d}, MathVerse\cite{10.1007/978-3-031-73242-3_10} (\textit{Vision-only} to strictly assess graphical grounding), and WeMath\cite{qiao2024wemathdoeslargemultimodal} (\textit{Strict Score} to penalize guessing). Extensive details regarding hyperparameters, identical decoding parameters, full evaluation logs, and dataset statistics are deferred to Appendices~\ref{app:hyperparams_and_setup}--\ref{app:full_logs} and ~\ref{app:detailed_stats_and_bottlenecks}.

\subsection{Quantitative Performance of SFT}
\label{subsec:exp_sft}

As shown in Table~\ref{tab:sft_main}, GeoSym achieves leading overall scores among equivalent-scale methods. While our geometry specialization incurs a minor alignment tax on broad-domain benchmarks like MathVision, it yields a highly concentrated gain profile on diagram-dependent tasks. Notably, the GeoSym Entry model surpasses the 8B baseline by an absolute \textbf{+23.10\%} on MathVerse Vision-only, empirically proving the efficacy of our rigorous pixel-to-symbol grounding in mitigating visual hallucinations.

Furthermore, GeoSym-driven training significantly preserves logical coherence during deep deduction. The GeoSym Hard configuration attains absolute gains of \textbf{+6.19\%} (8B base) and \textbf{+9.90\%} (7B base) on WeMath, substantially outperforming existing synthesis baselines (e.g., GeoTrust-train). This demonstrates that our verified symbolic reasoning chains effectively mitigate the long-horizon logic fragmentation inherent in traditional models. Complete class-wise logs isolating geometry sub-categories are provided in Appendix~\ref{app:full_logs}.

\subsection{The Impact of GRPO: Pushing the Ceiling}
\label{subsec:exp_rl}

To investigate if deterministic exact-match rewards can elevate reasoning beyond supervised cloning, we deploy GRPO (Table~\ref{tab:rl_main}). Zero-shot GRPO on the 7B base unlocks the multi-step upper bound, yielding a remarkable \textbf{+7.24\%} absolute gain on strict WeMath scores. Moreover, we observe an exceptional synergy between structural SFT and RL. Applying GRPO-Entry atop the GeoSym Hard SFT checkpoint achieves a peak overall score of \textbf{44.99}. These enhancements are most pronounced in deep multi-hop reasoning categories, confirming that our verifiable exact-match rewards actively and safely guide the policy search toward mathematically sound trajectories (Table~\ref{tab:grpo_full_results}).

\subsection{Ablation Studies}
\label{subsec:ablations}
\begin{figure*}[htbp]
    \centering
    
    \includegraphics[width=\linewidth]{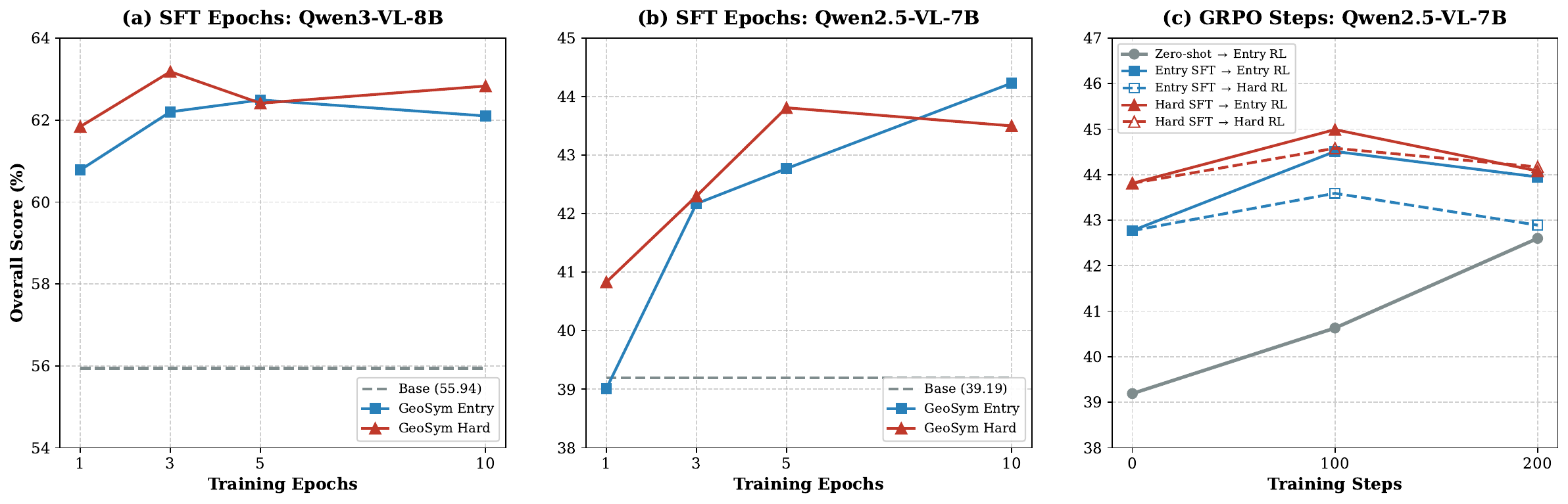}
    
    \vspace{0cm} 
    
    \includegraphics[width=\linewidth]{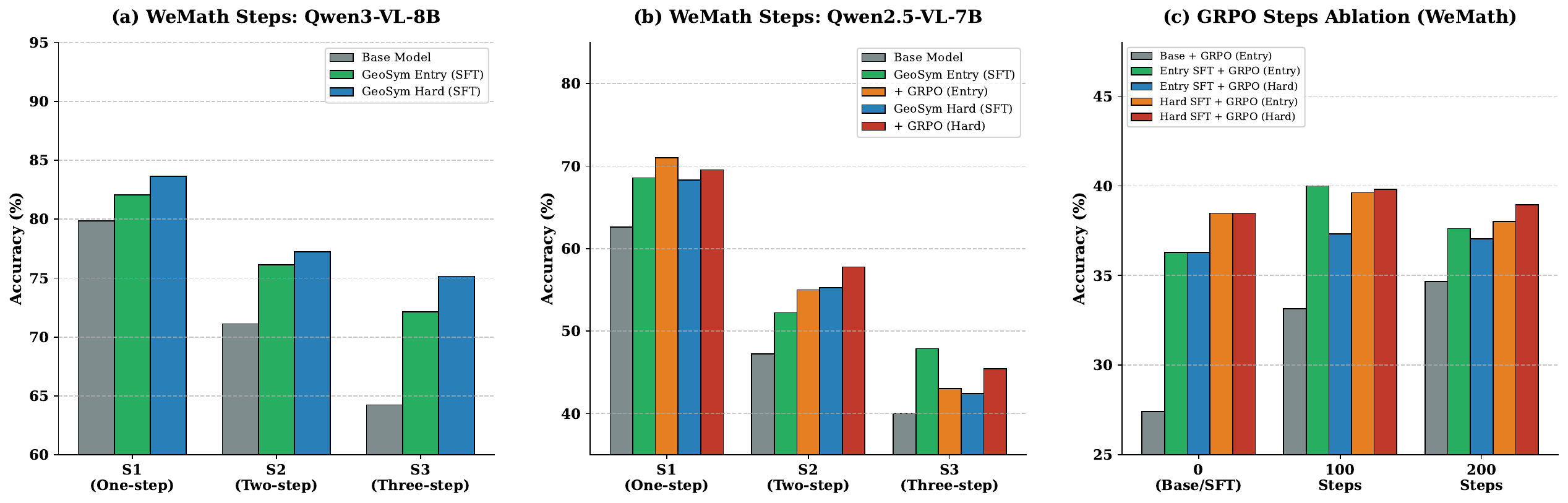}
    
    \caption{\textbf{Comprehensive Training Dynamics and Multi-Step Robustness.} 
    \textbf{Top Row:} Ablation on SFT epochs and GRPO optimization steps. The evaluation identifies an optimal training \textit{sweet spot} around 3--5 SFT epochs and 100 GRPO steps, beyond which the models generally experience diminishing returns or over-optimization regression. Furthermore, initializing RL with structural SFT checkpoints significantly elevates the performance ceiling compared to zero-shot RL.
    \textbf{Bottom Row:} Performance degradation across one-step (S1) to three-step (S3) geometric problems in WeMath. While the zero-shot Base models suffer from severe performance decay in long-horizon tasks, our GeoSym-driven SFT and subsequent GRPO phases grant substantial deductive robustness, drastically narrowing the performance gap in deep multi-hop reasoning.}
    \label{fig:combined_ablation_degradation}
    
    \vspace{-0.3cm} 
    
\end{figure*}

To investigate GeoSym's scalability and robustness, we conduct comprehensive ablations, revealing three key insights: \textbf{(1) Architecture Scaling:} GeoSym's efficacy is independent of parameter capacity. Scaling down to 4B and 3B models (Table~\ref{tab:sft_main}) consistently lifts performance, notably boosting the 4B base by \textbf{+6.08\%} overall and a massive \textbf{+24.87\%} on MathVerse Vision-only. \textbf{(2) Mitigating Multi-Step Degradation:} As illustrated in Figure~\ref{fig:combined_ablation_degradation} (Bottom Row), GeoSym's accuracy gains on WeMath compound as reasoning steps increase, peaking at a \textbf{+10.91\%} absolute improvement on the most complex `S3' subset for the 7B architecture. \textbf{(3) Training Dynamics and RL Initialization:} Figure~\ref{fig:combined_ablation_degradation} (Top Row) identifies 3--5 SFT epochs and 100 GRPO steps as the optimal training \textit{sweet spot}. Critically, initializing GRPO from SFT checkpoints substantially elevates the performance ceiling compared to zero-shot RL, demonstrating that foundational neuro-symbolic alignment is a prerequisite for maximizing RL efficacy. Exhaustive data logs for these ablations are deferred to Appendix~\ref{app:full_logs} (Tables~\ref{tab:ablation_master} and \ref{tab:grpo_full_results}).

\section{Conclusion}
\label{sec:conclusion}

We introduced \textbf{GeoSym}, a neuro-symbolic framework for scalable and verifiable multimodal geometric reasoning. By combining difficulty-stratified synthesis, exact symbolic derivations, and verified CoT supervision, GeoSym improves visual grounding and multi-hop logical consistency in LMMs. We further integrate GRPO with deterministic exact-match rewards to enhance structural reasoning while reducing reward-hacking risks. Experiments on MathVista, MathVerse, MathVision, and WeMath show that GeoSym consistently improves open-source models at comparable scales, mitigating visual hallucination and logic fragmentation. These results suggest that strict neuro-symbolic alignment offers a promising path toward more reliable multimodal mathematical agents. A detailed discussion regarding the current limitations of our framework is deferred to the Appendix~\ref{app:limitations}.

\small
\bibliographystyle{plainnat}
\bibliography{custom}

\clearpage

\appendix

\section{GeoSym127K Dataset Samples and Comparison}
\label{app:samples_and_comparison}

\begin{figure*}[htbp] 
    \vspace{0.3cm} 
    \centering
    
    \includegraphics[width=0.242\linewidth]{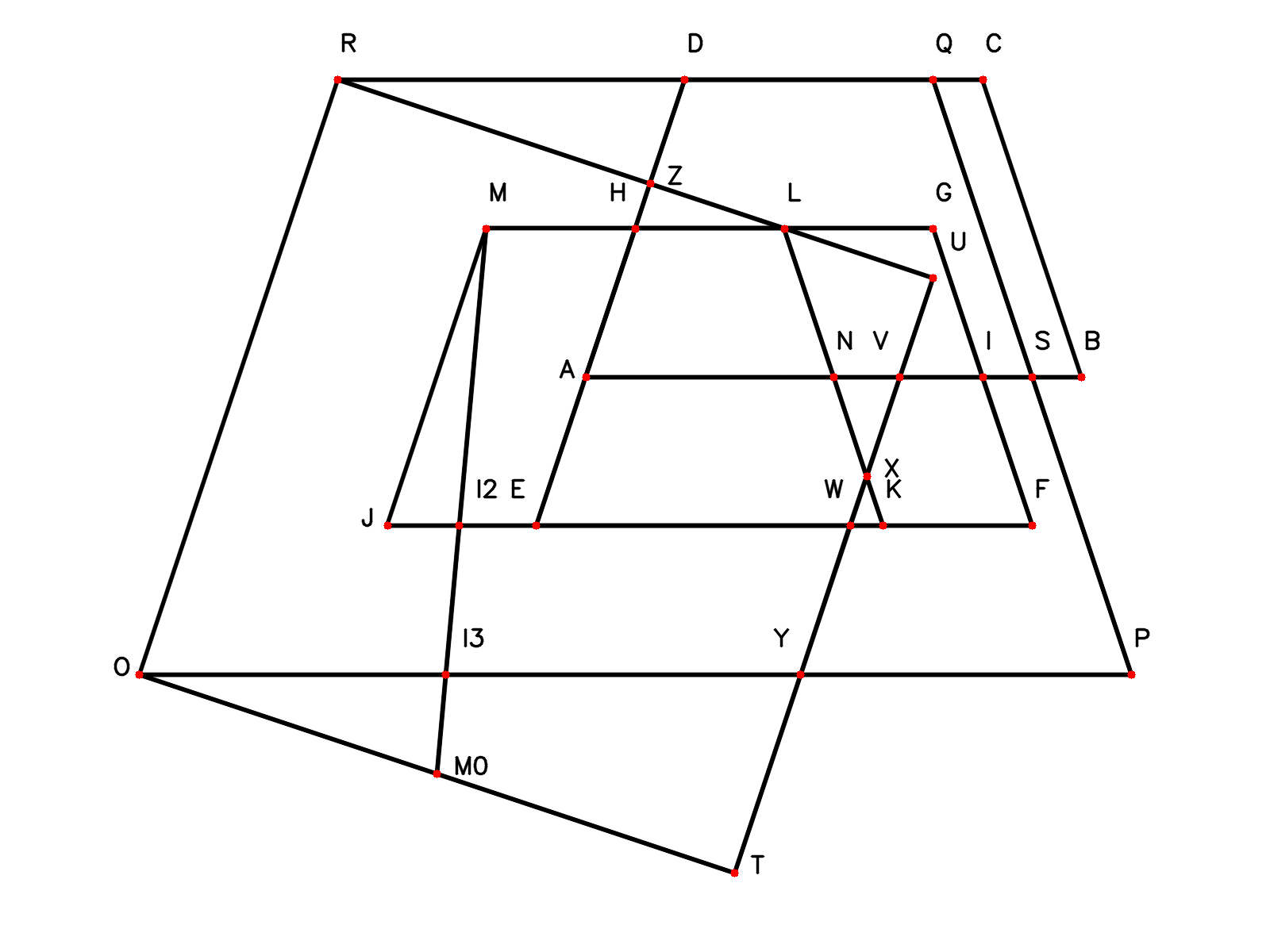}\hfill
    \includegraphics[width=0.242\linewidth]{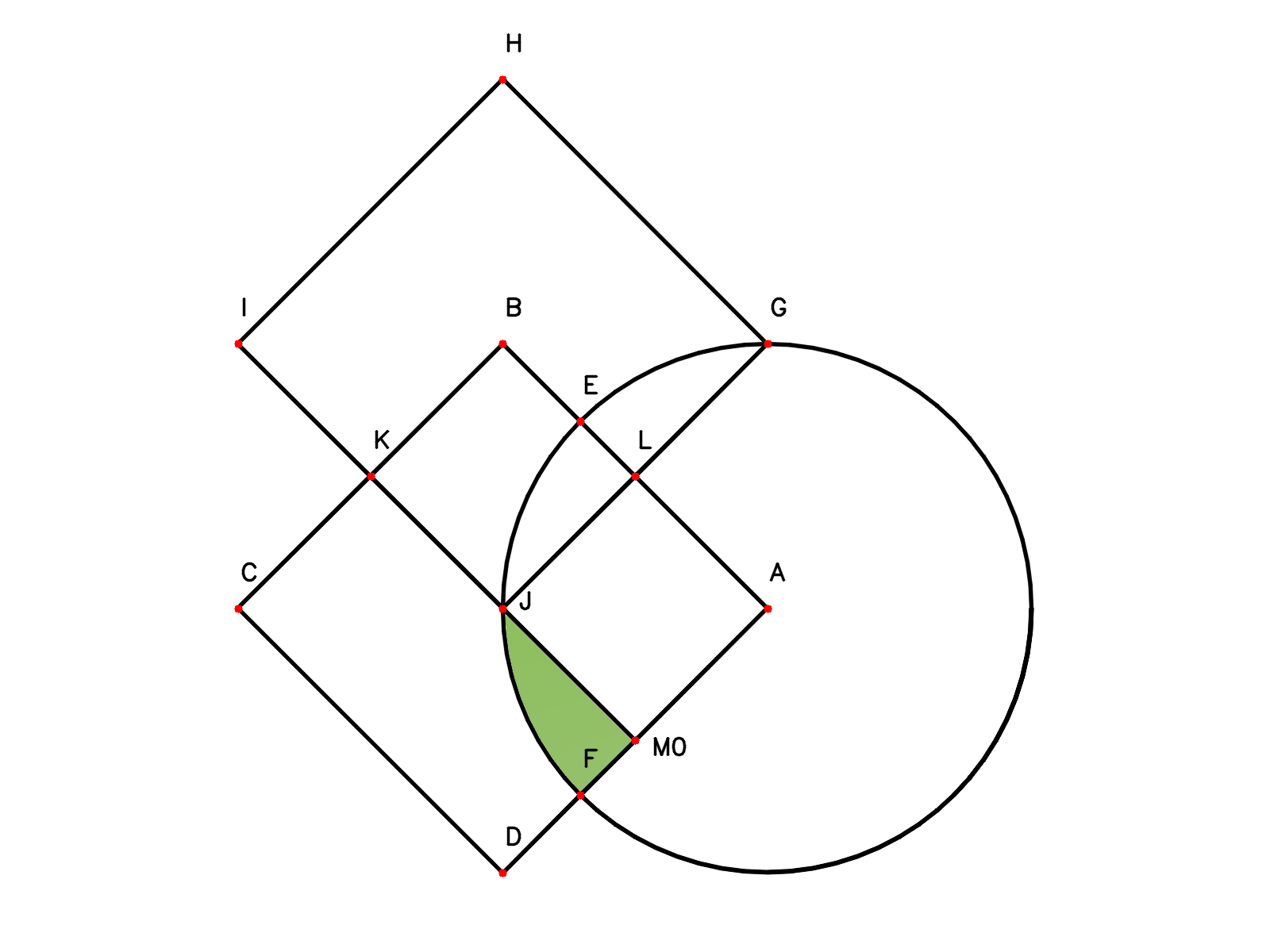}\hfill
    \includegraphics[width=0.242\linewidth]{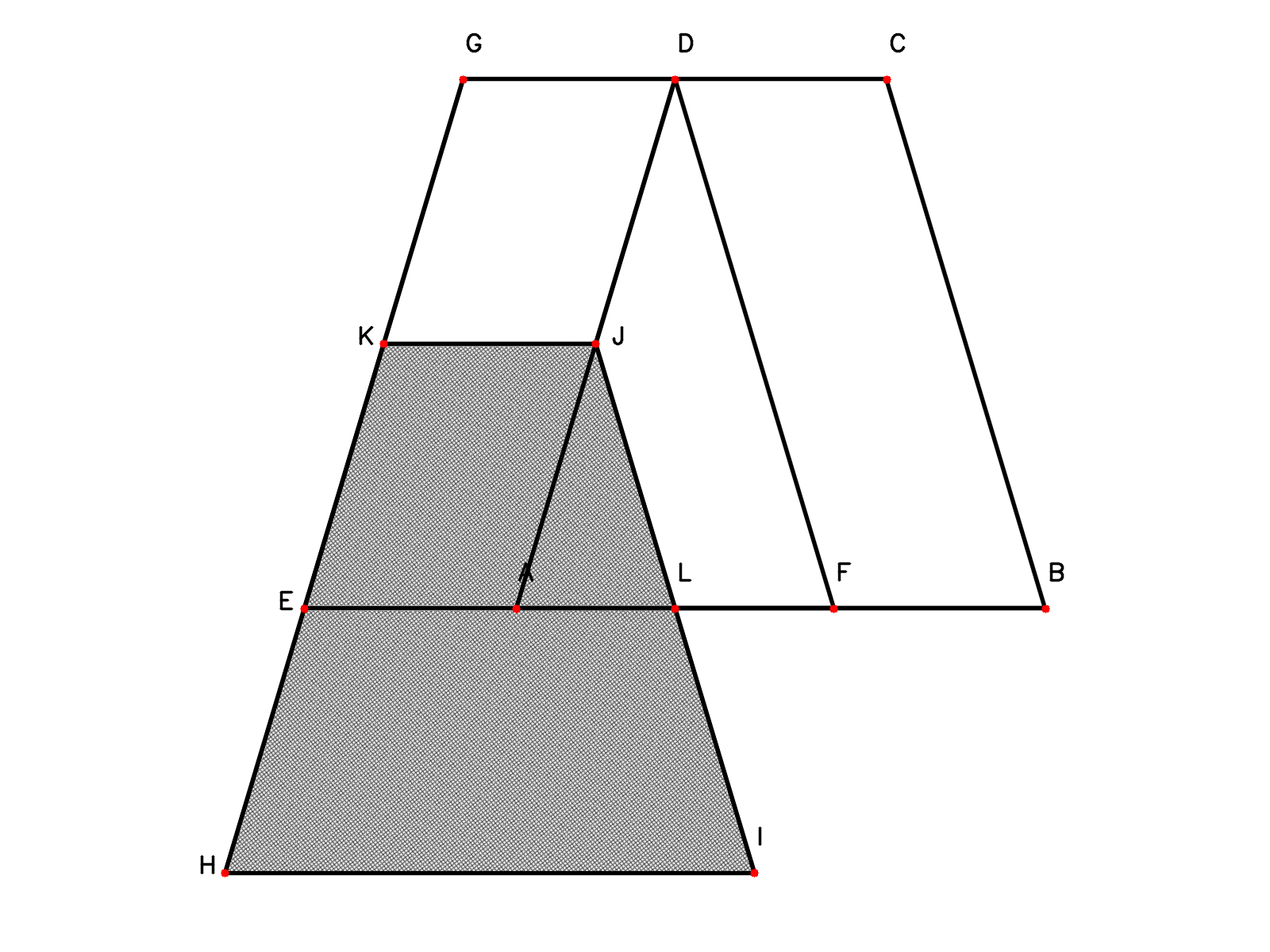}\hfill
    \includegraphics[width=0.242\linewidth]{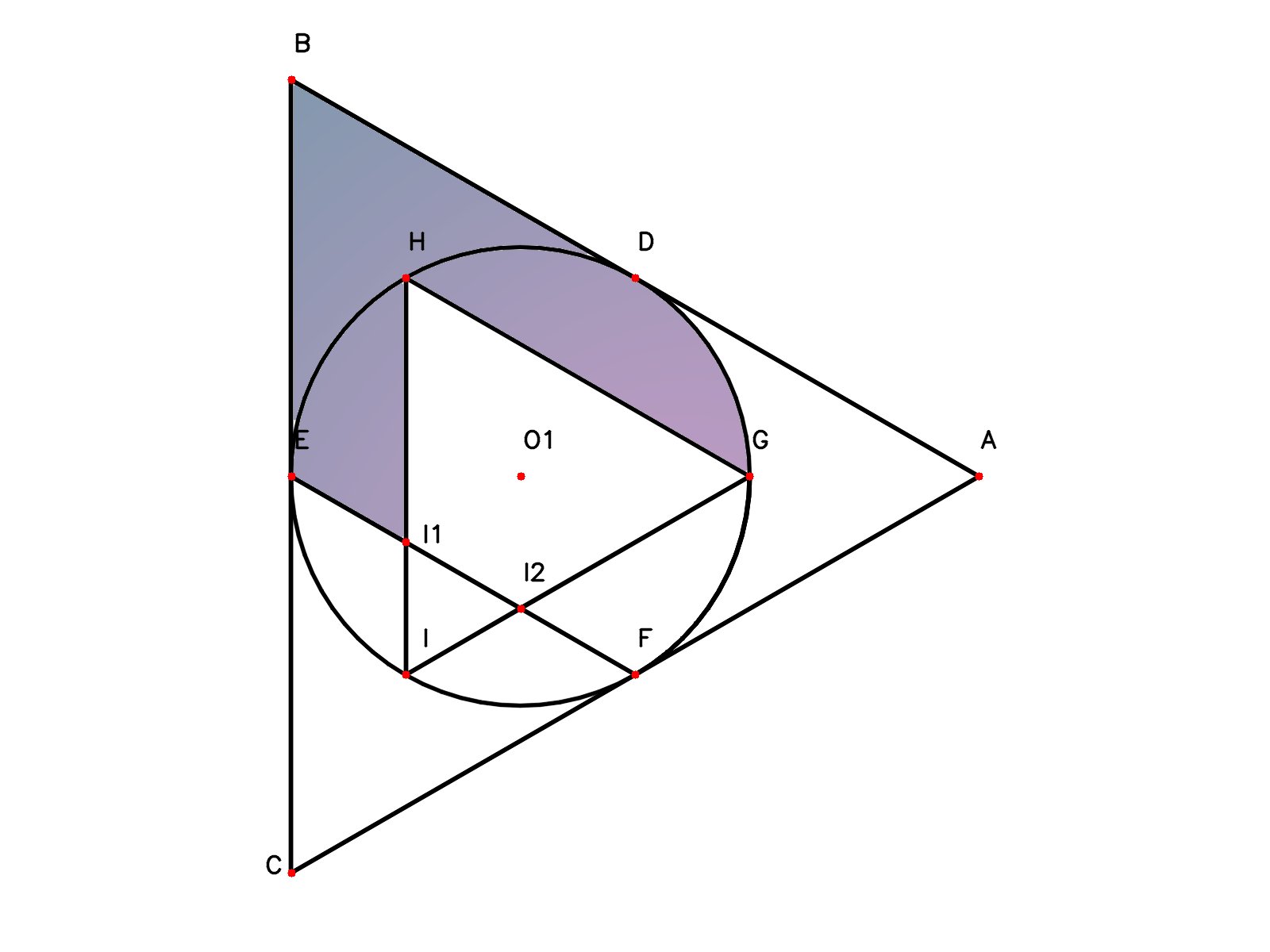}\\[2pt]
    
    \includegraphics[width=0.242\linewidth]{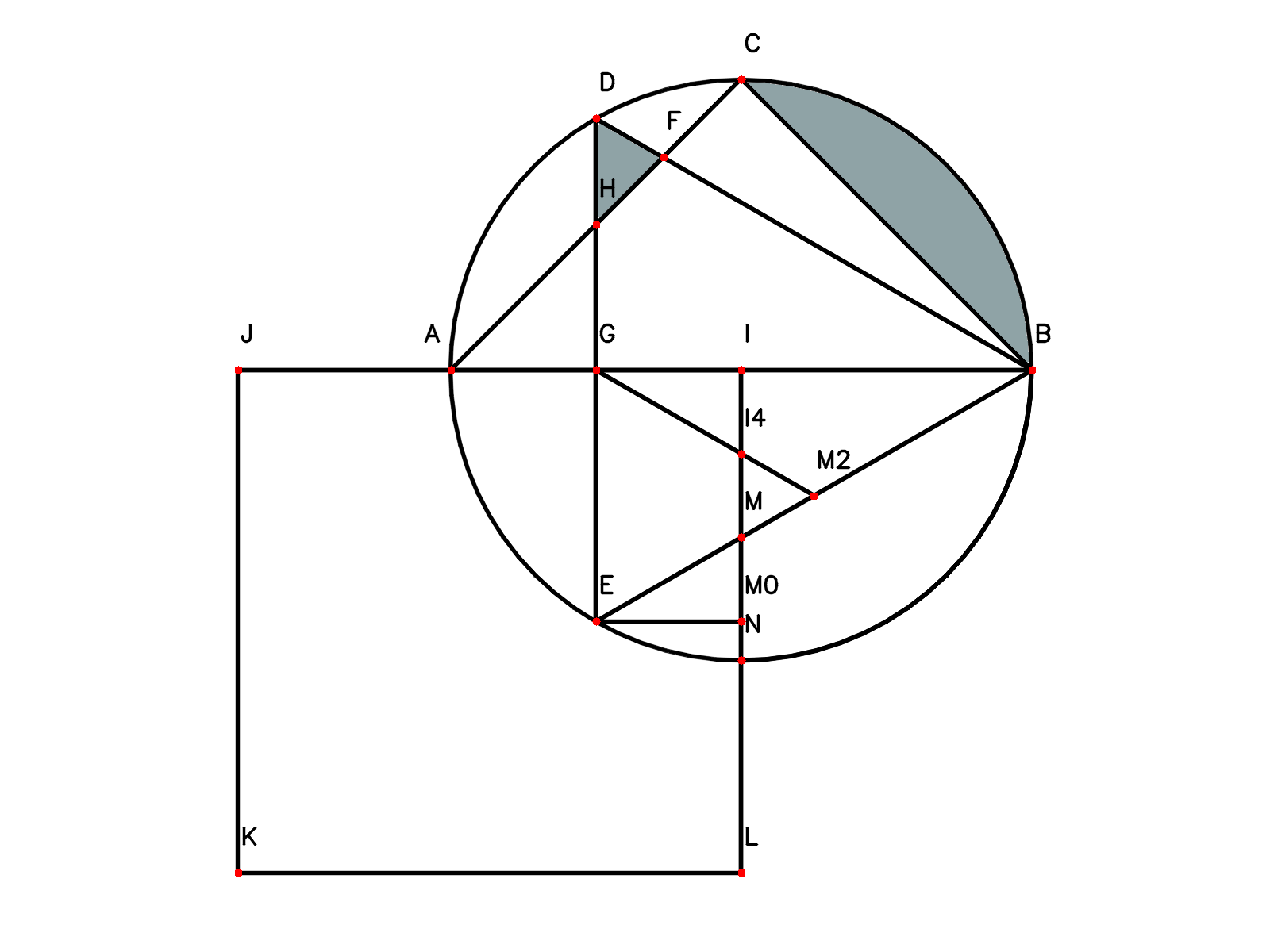}\hfill
    \includegraphics[width=0.242\linewidth]{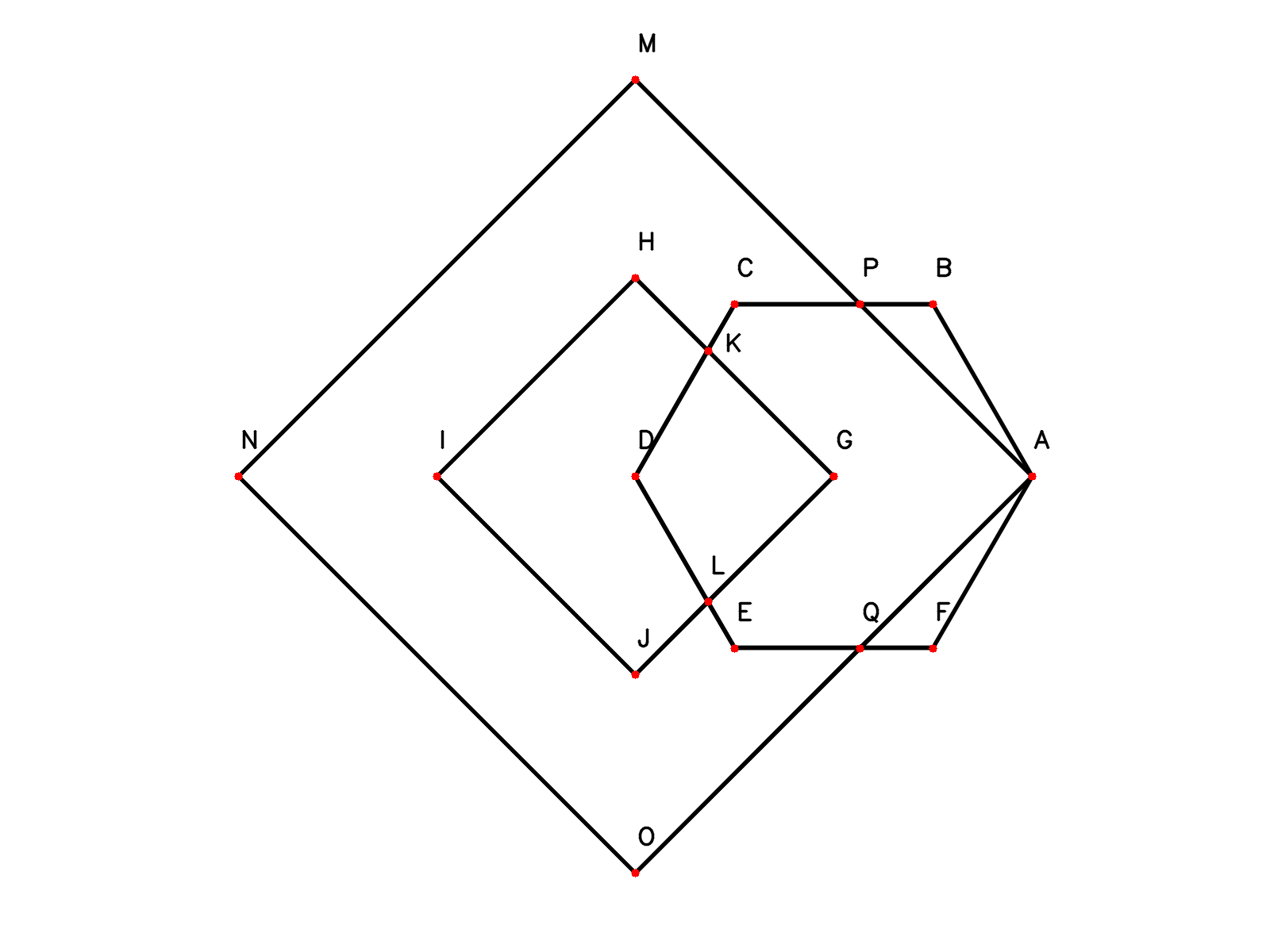}\hfill
    \includegraphics[width=0.242\linewidth]{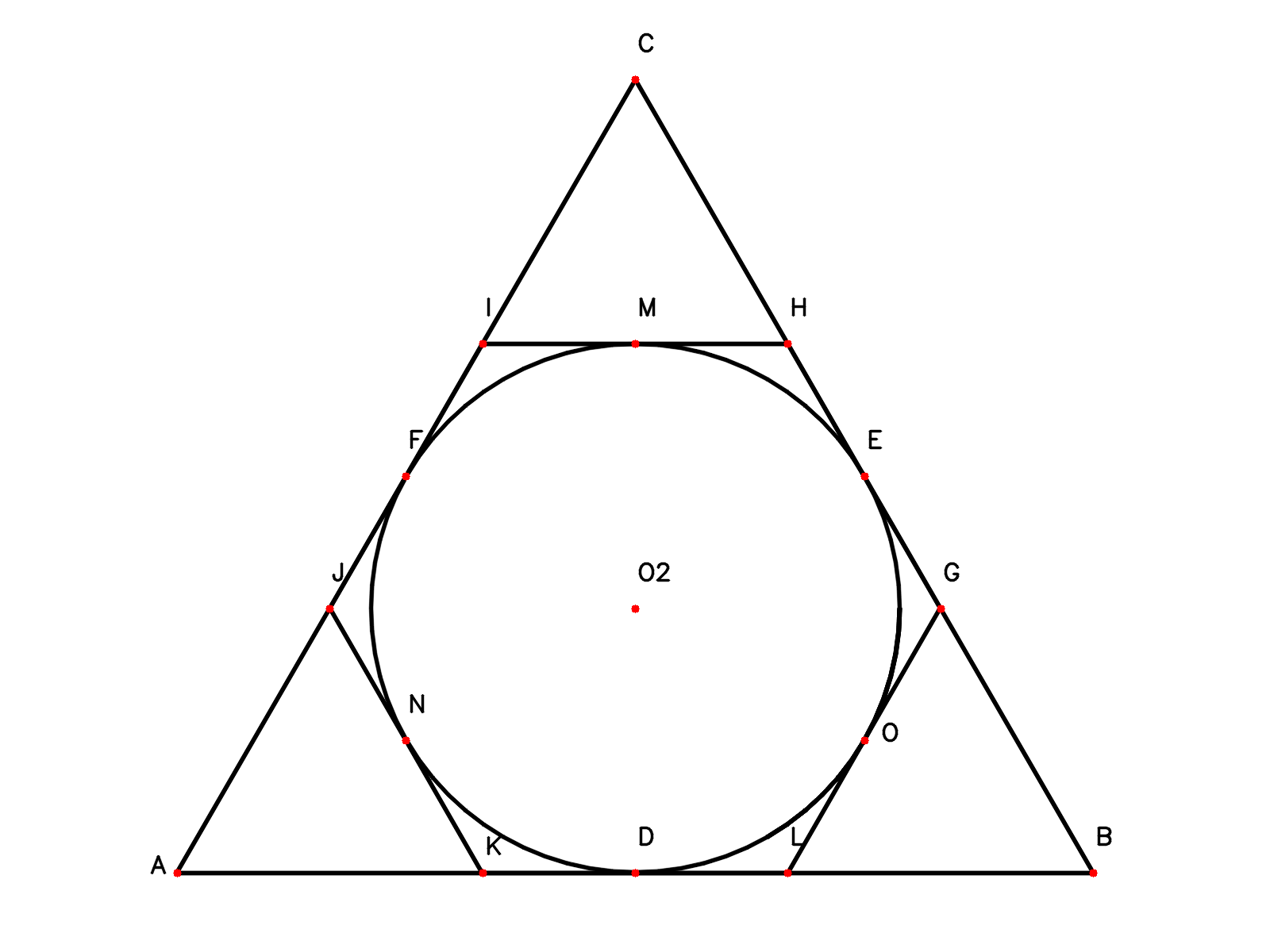}\hfill
    \includegraphics[width=0.242\linewidth]{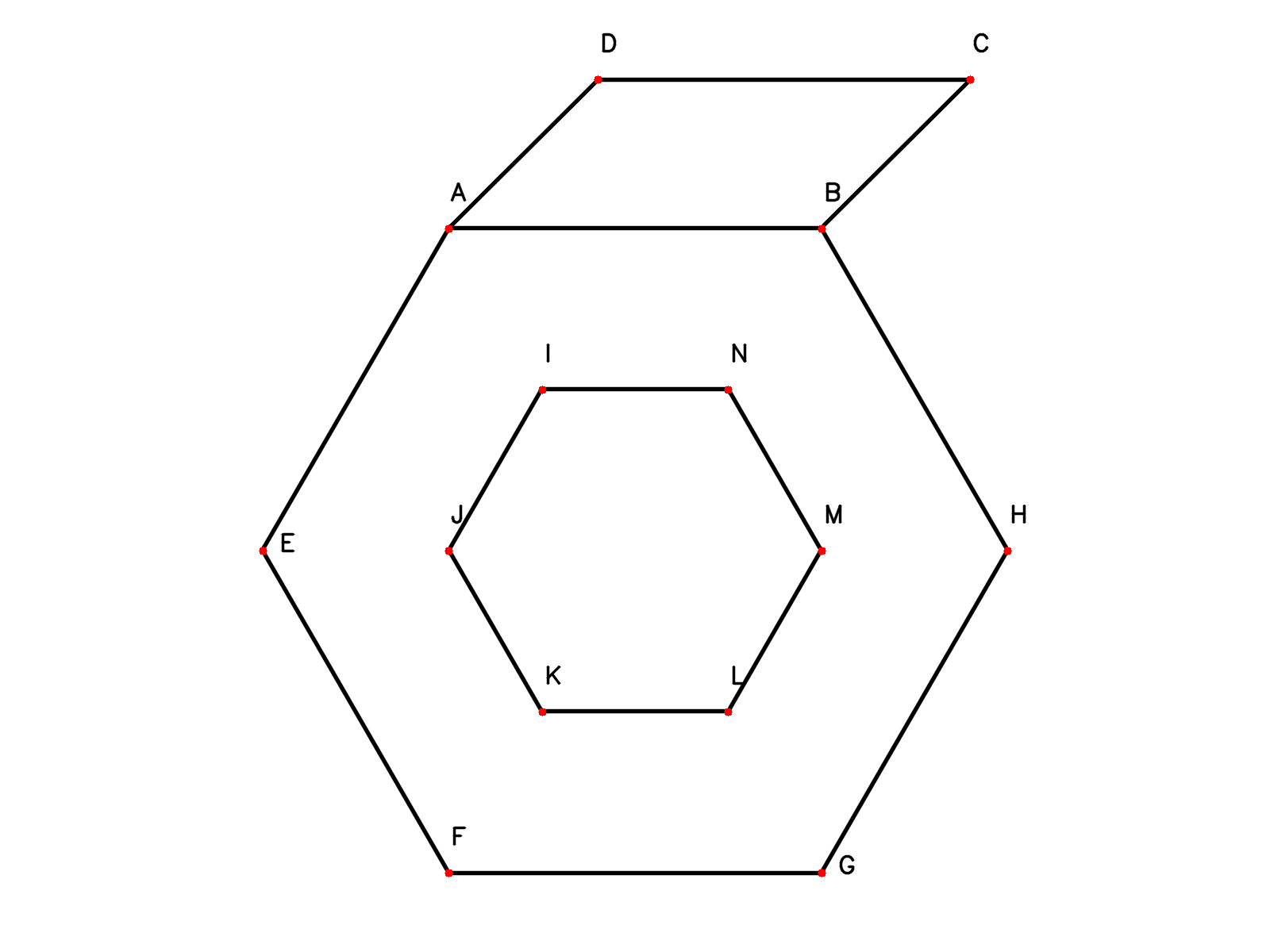}\\[2pt]
    
    \includegraphics[width=0.242\linewidth]{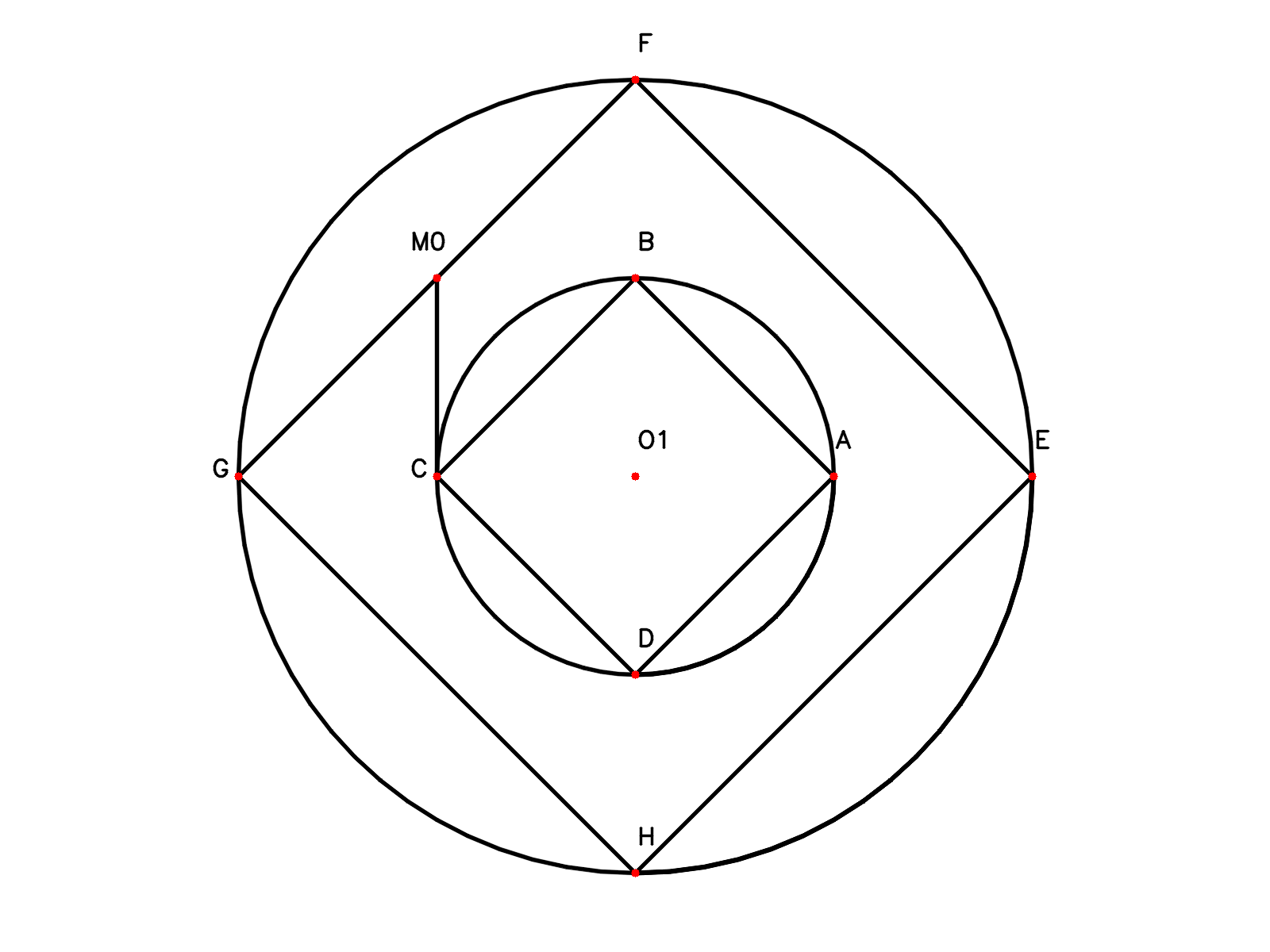}\hfill
    \includegraphics[width=0.242\linewidth]{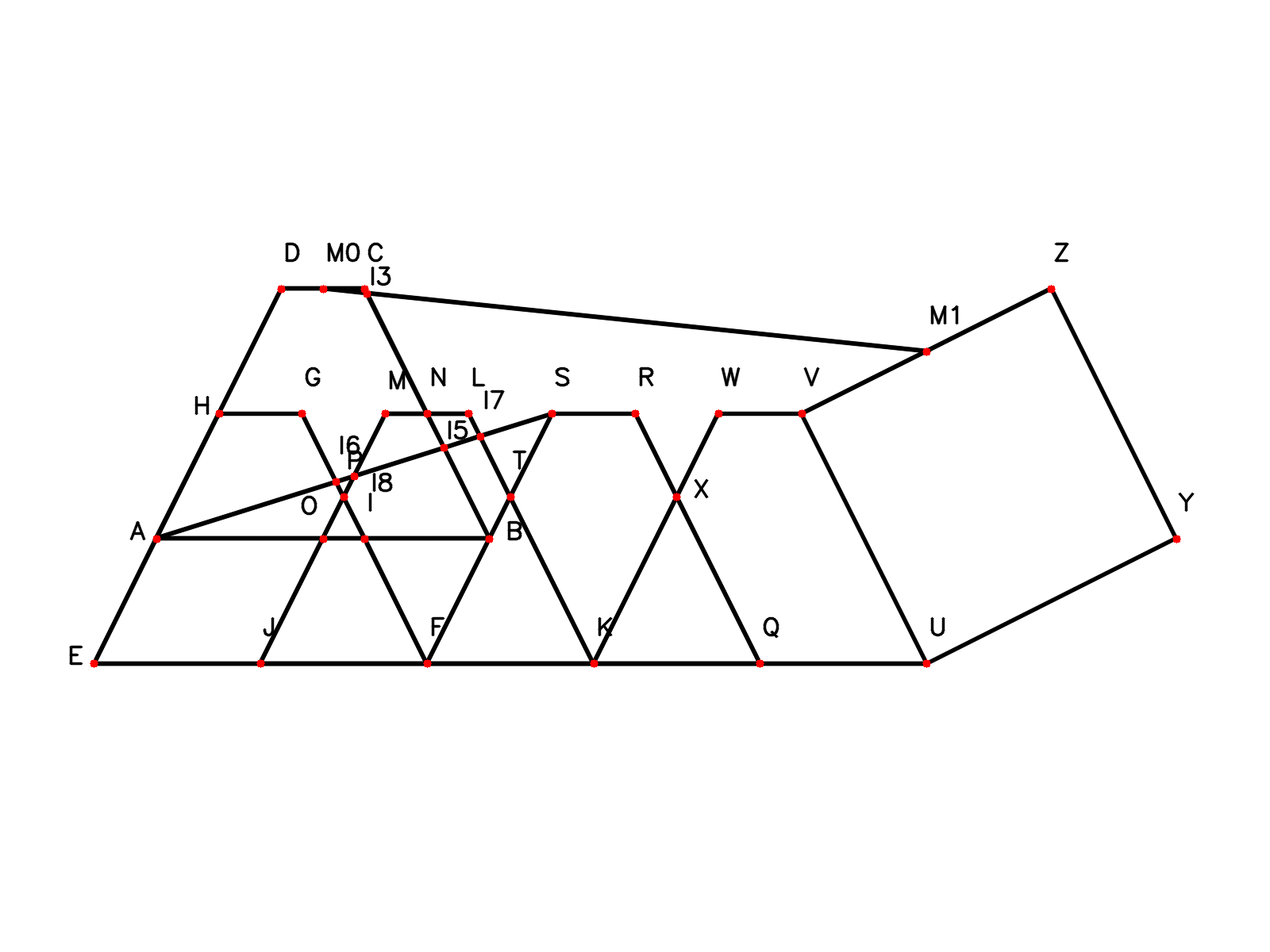}\hfill
    \includegraphics[width=0.242\linewidth]{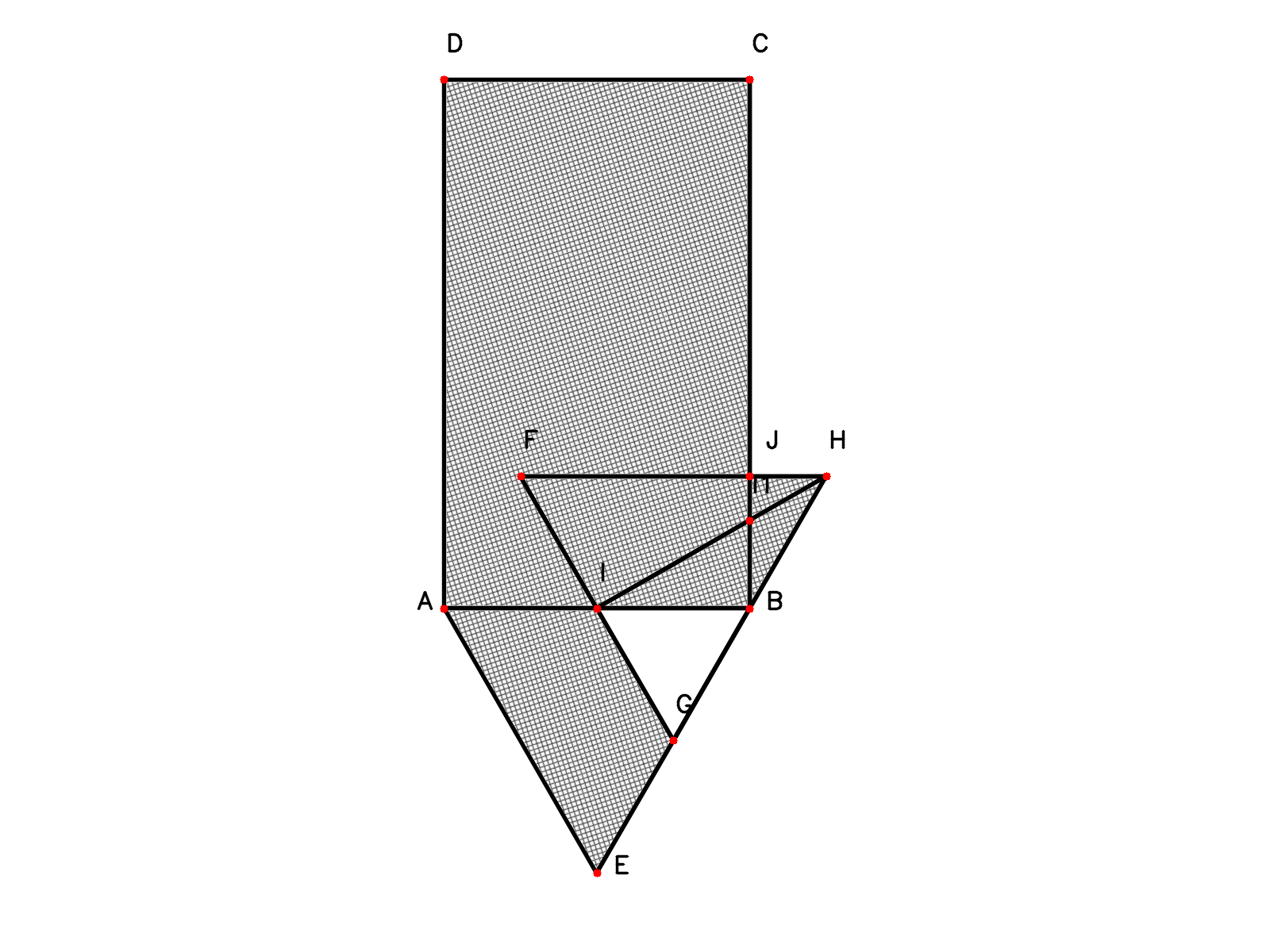}\hfill
    \includegraphics[width=0.242\linewidth]{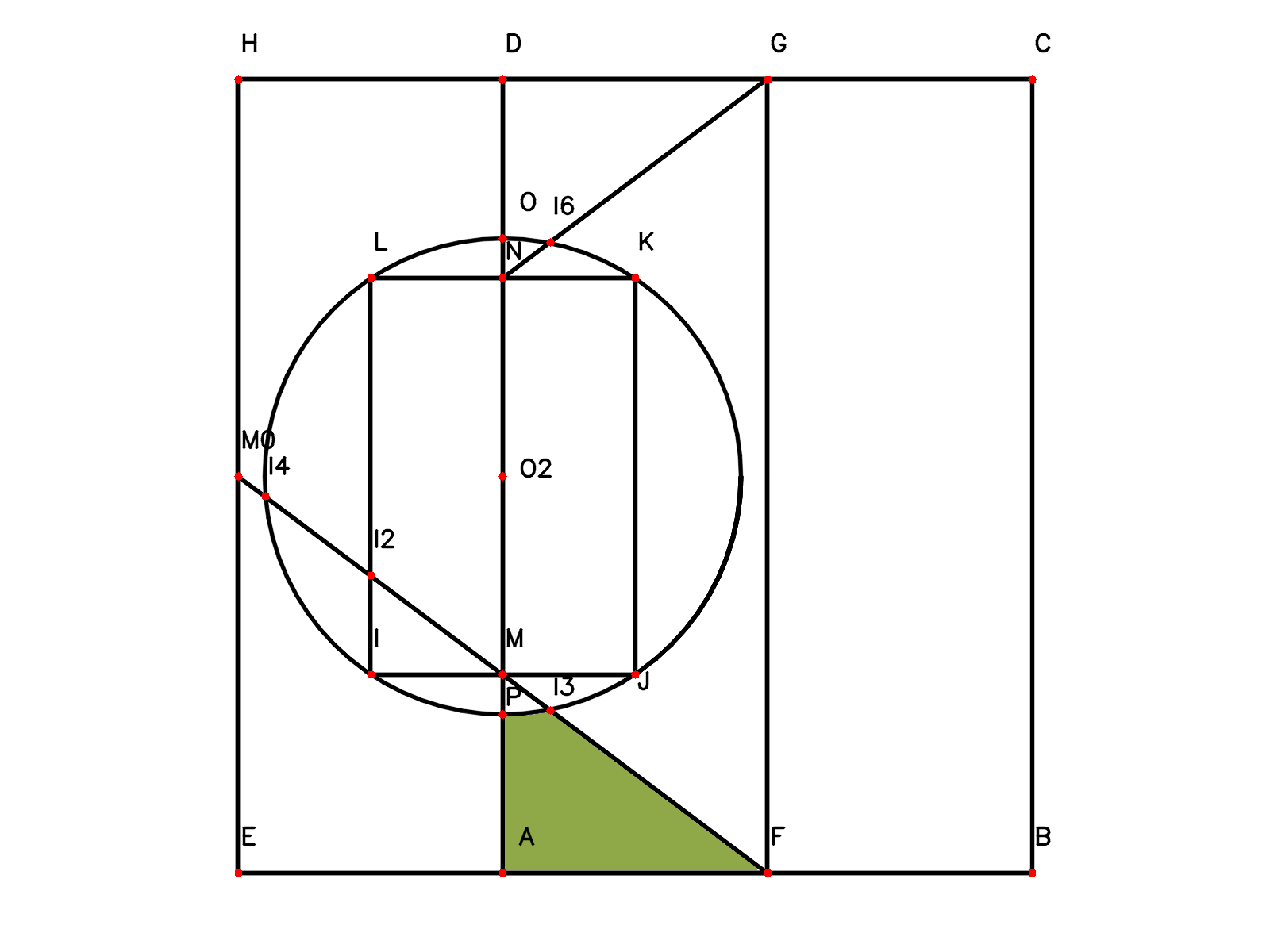}\\[2pt]
    
    \includegraphics[width=0.242\linewidth]{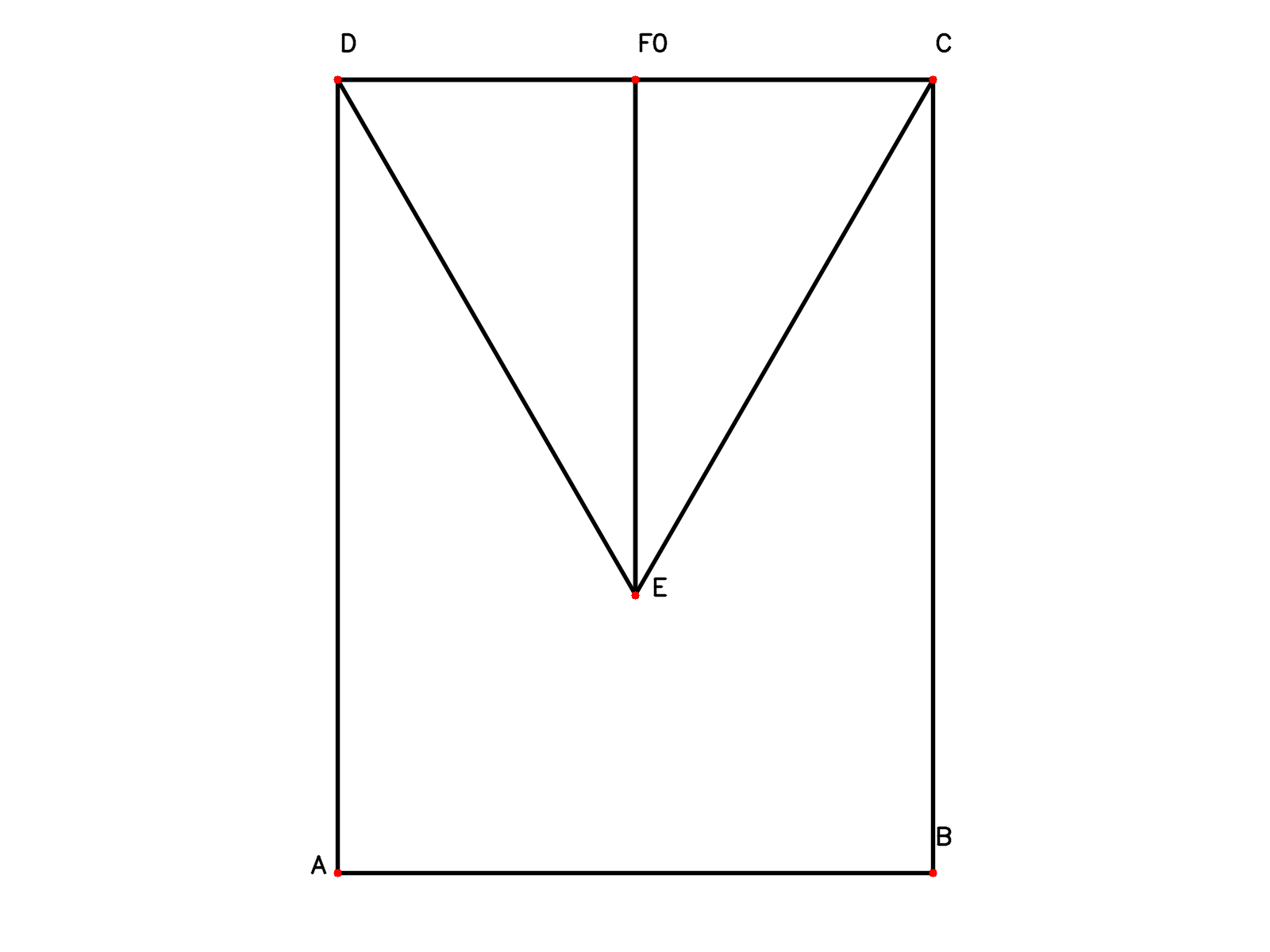}\hfill
    \includegraphics[width=0.242\linewidth]{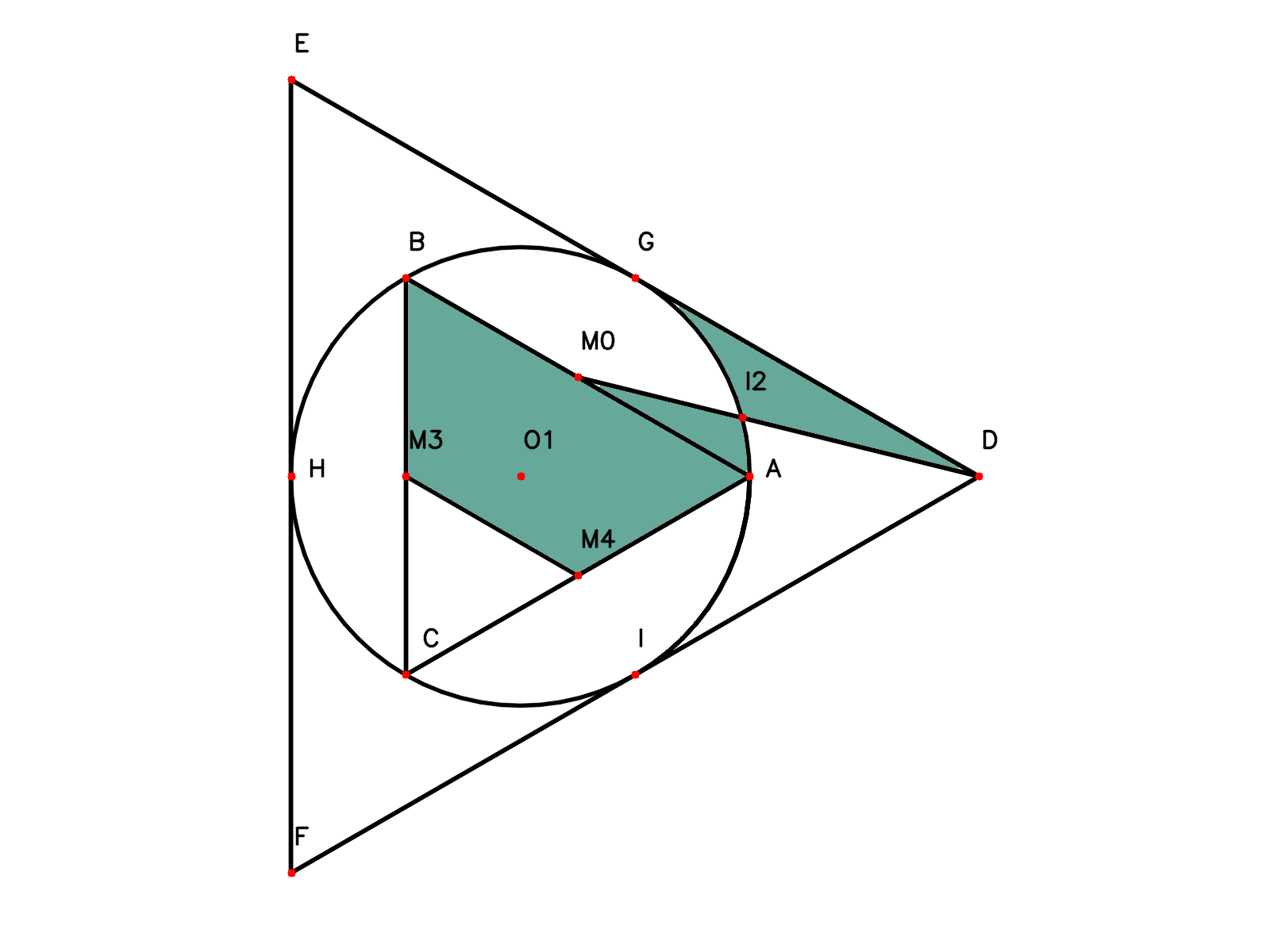}\hfill
    \includegraphics[width=0.242\linewidth]{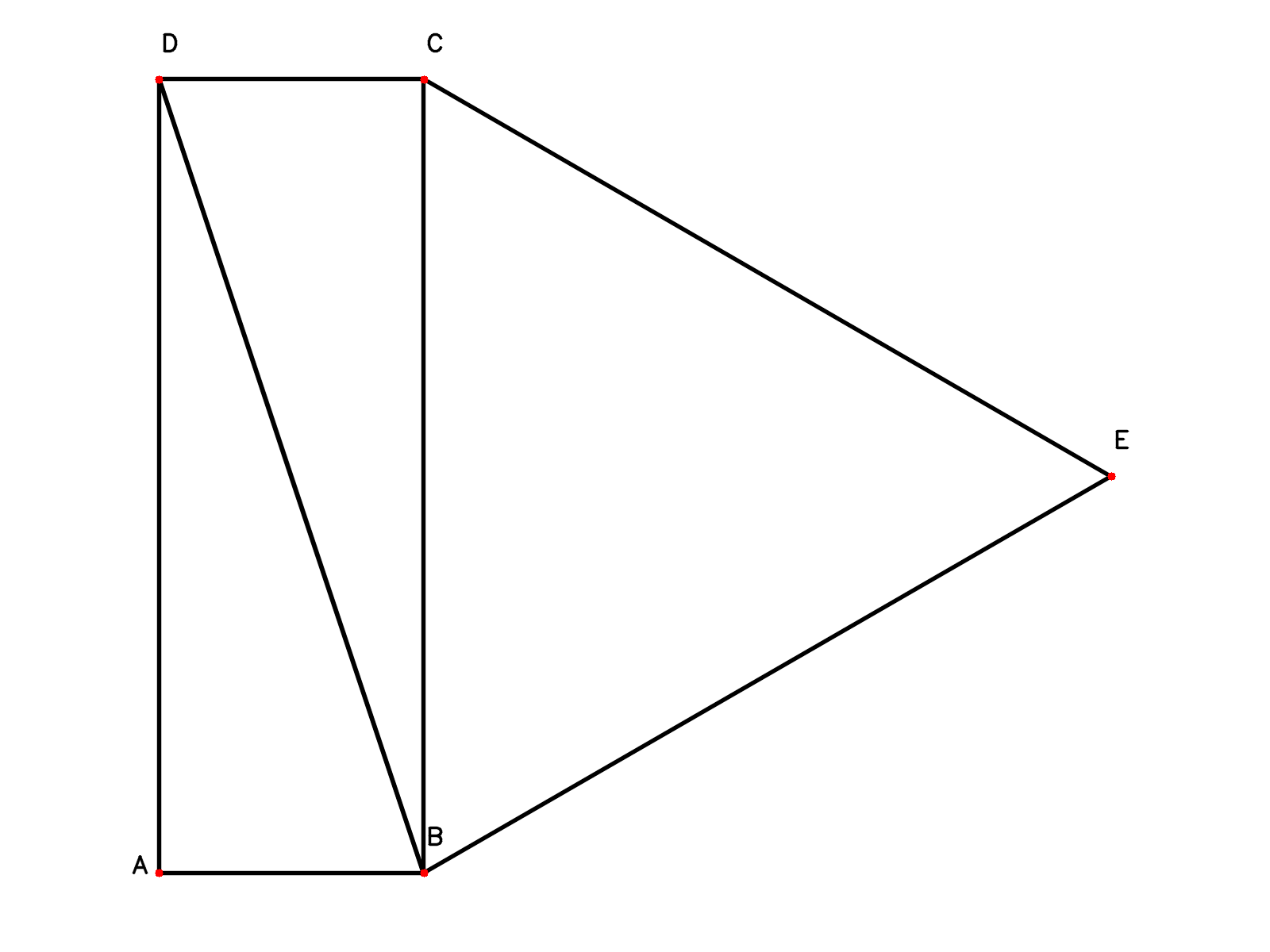}\hfill
    \includegraphics[width=0.242\linewidth]{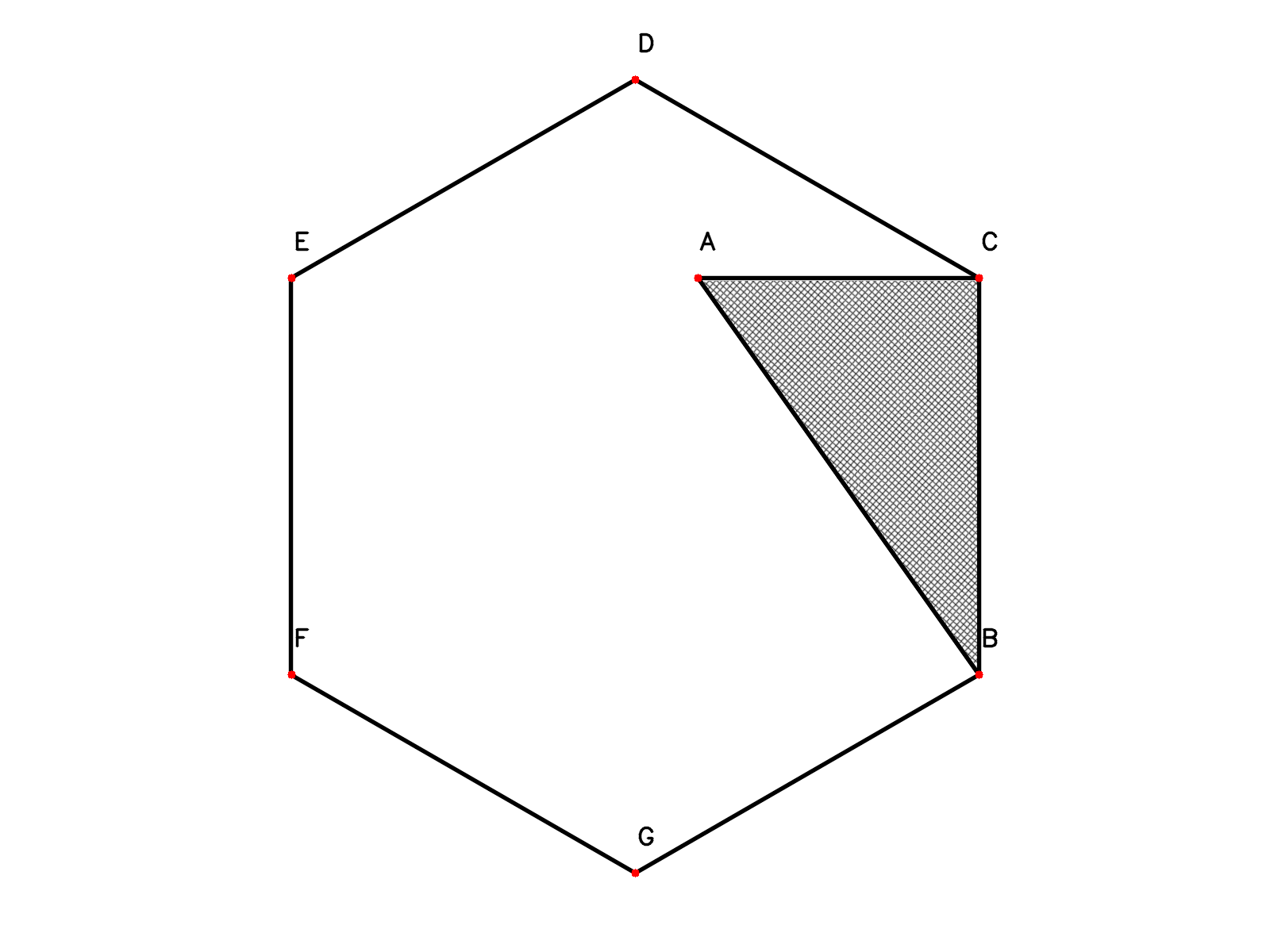}\\[2pt]

    \includegraphics[width=0.242\linewidth]{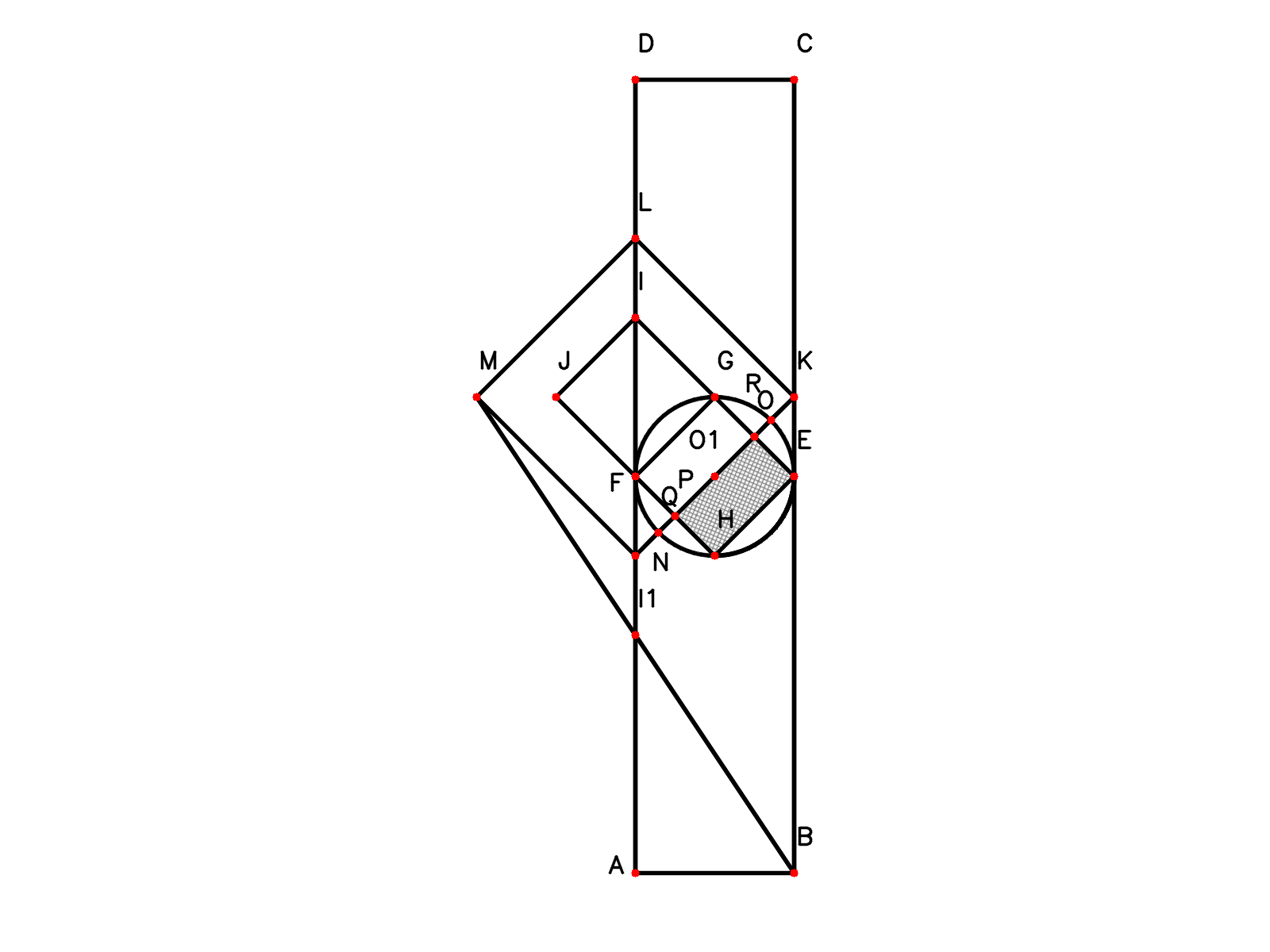}\hfill
    \includegraphics[width=0.242\linewidth]{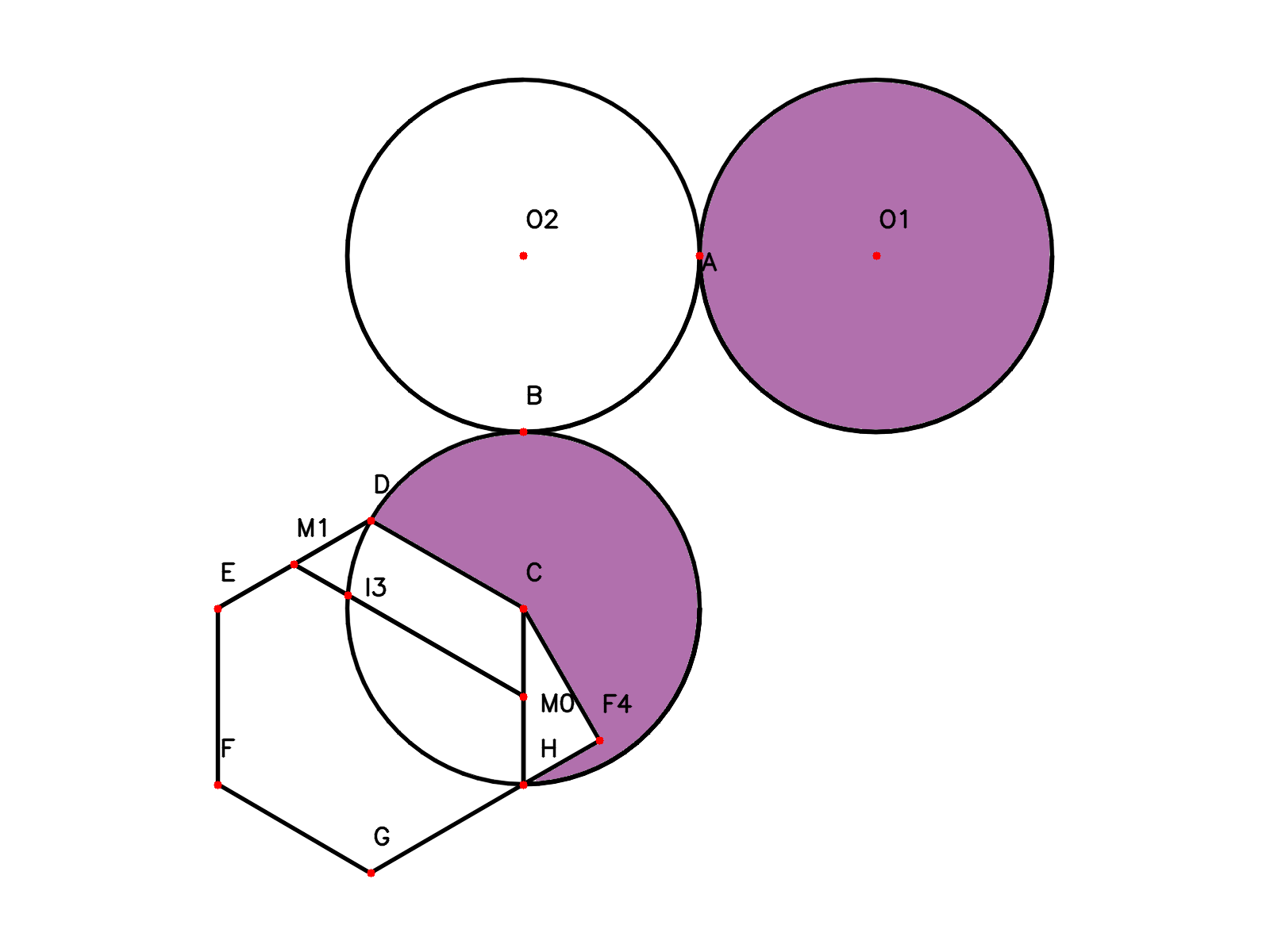}\hfill
    \includegraphics[width=0.242\linewidth]{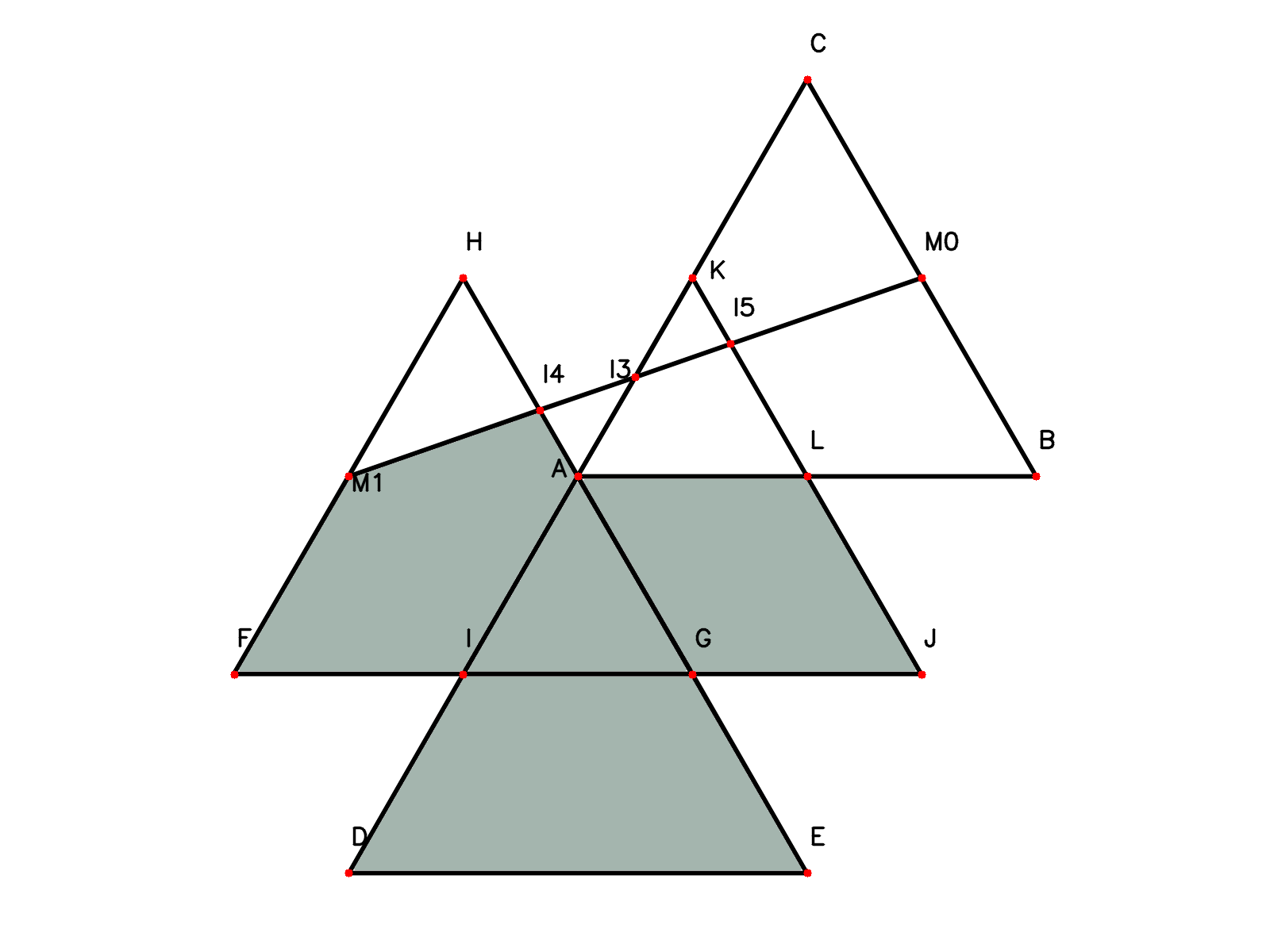}\hfill
    \includegraphics[width=0.242\linewidth]{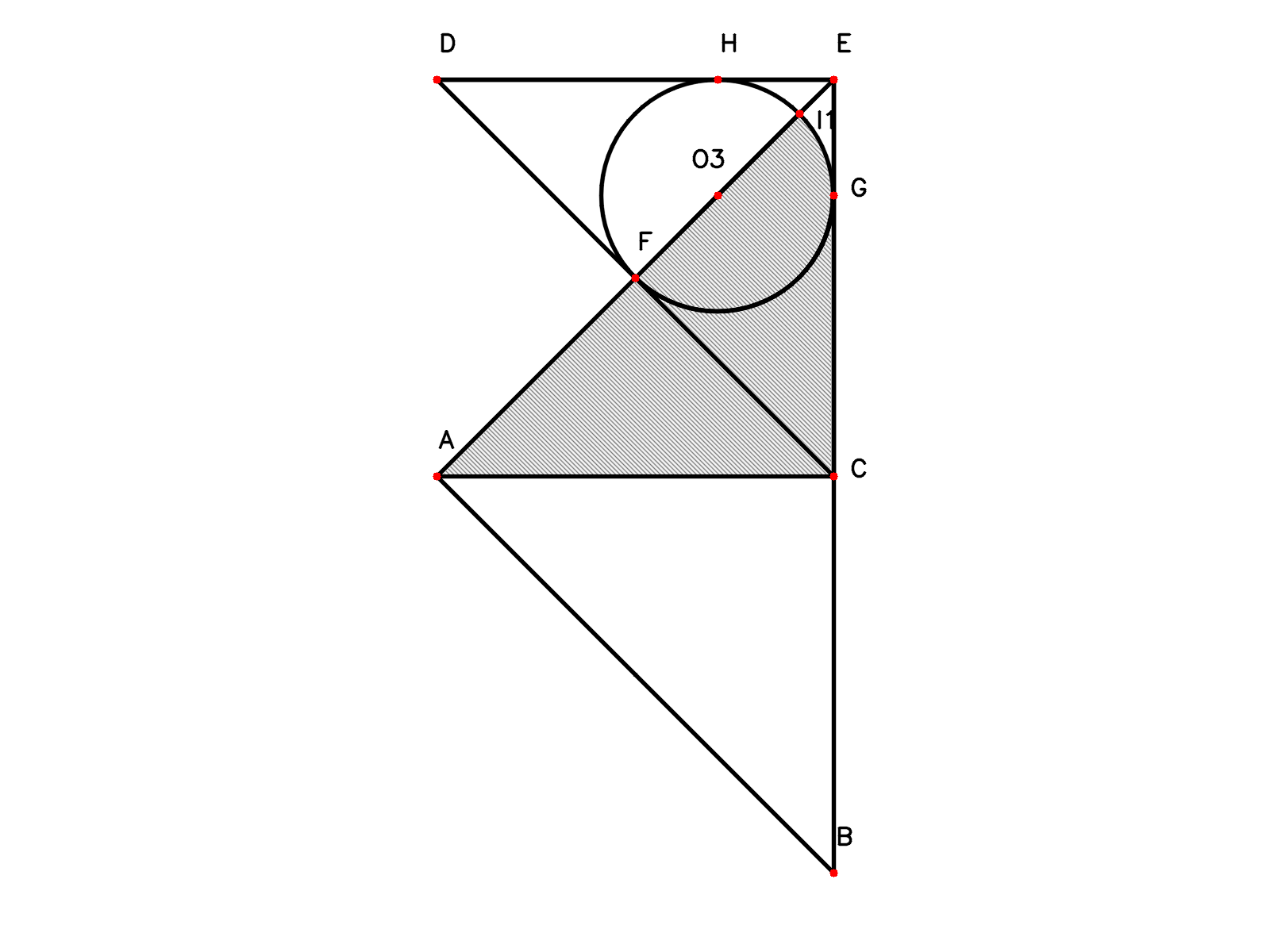}\\[2pt]

    \includegraphics[width=0.242\linewidth]{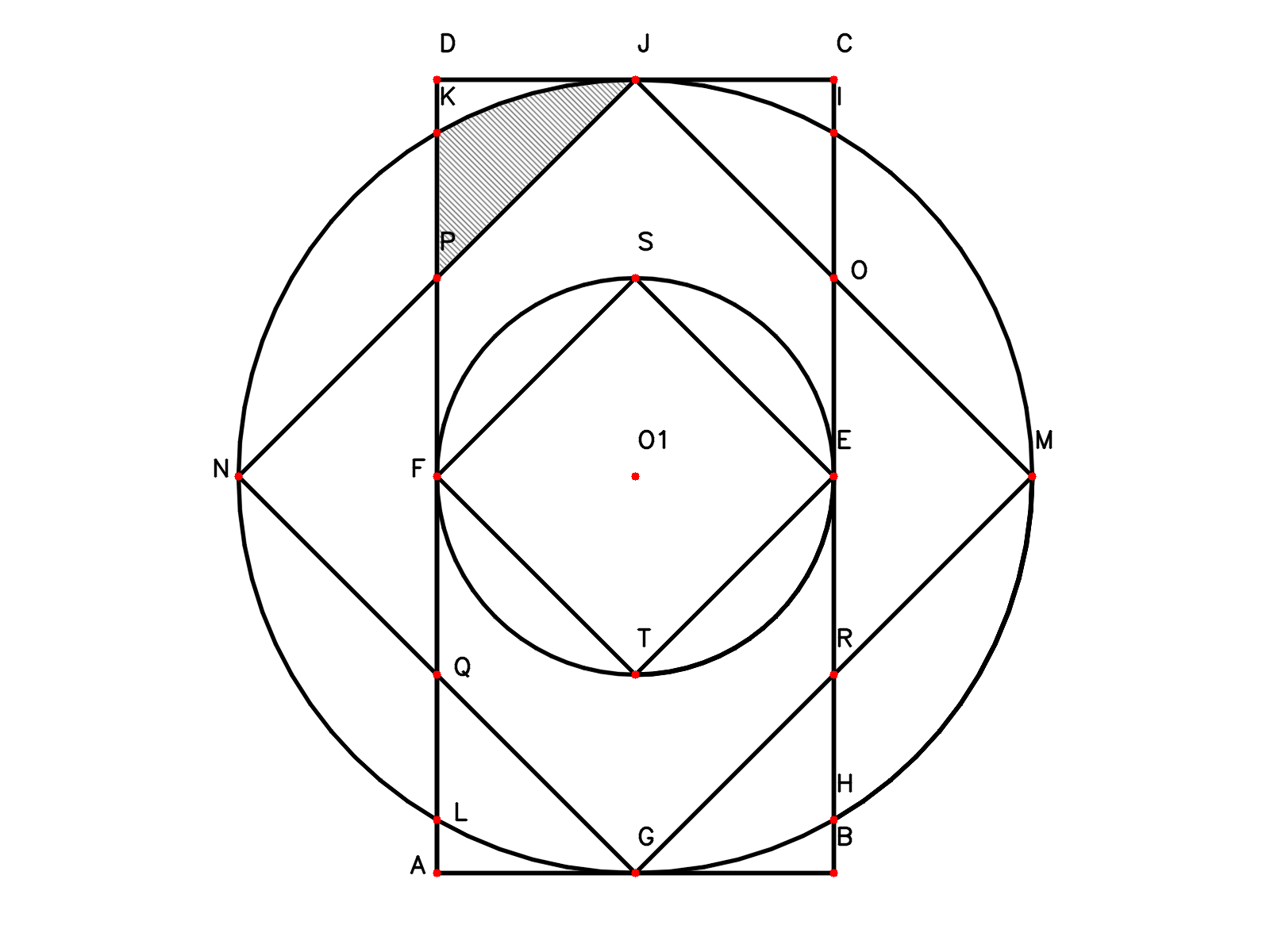}\hfill
    \includegraphics[width=0.242\linewidth]{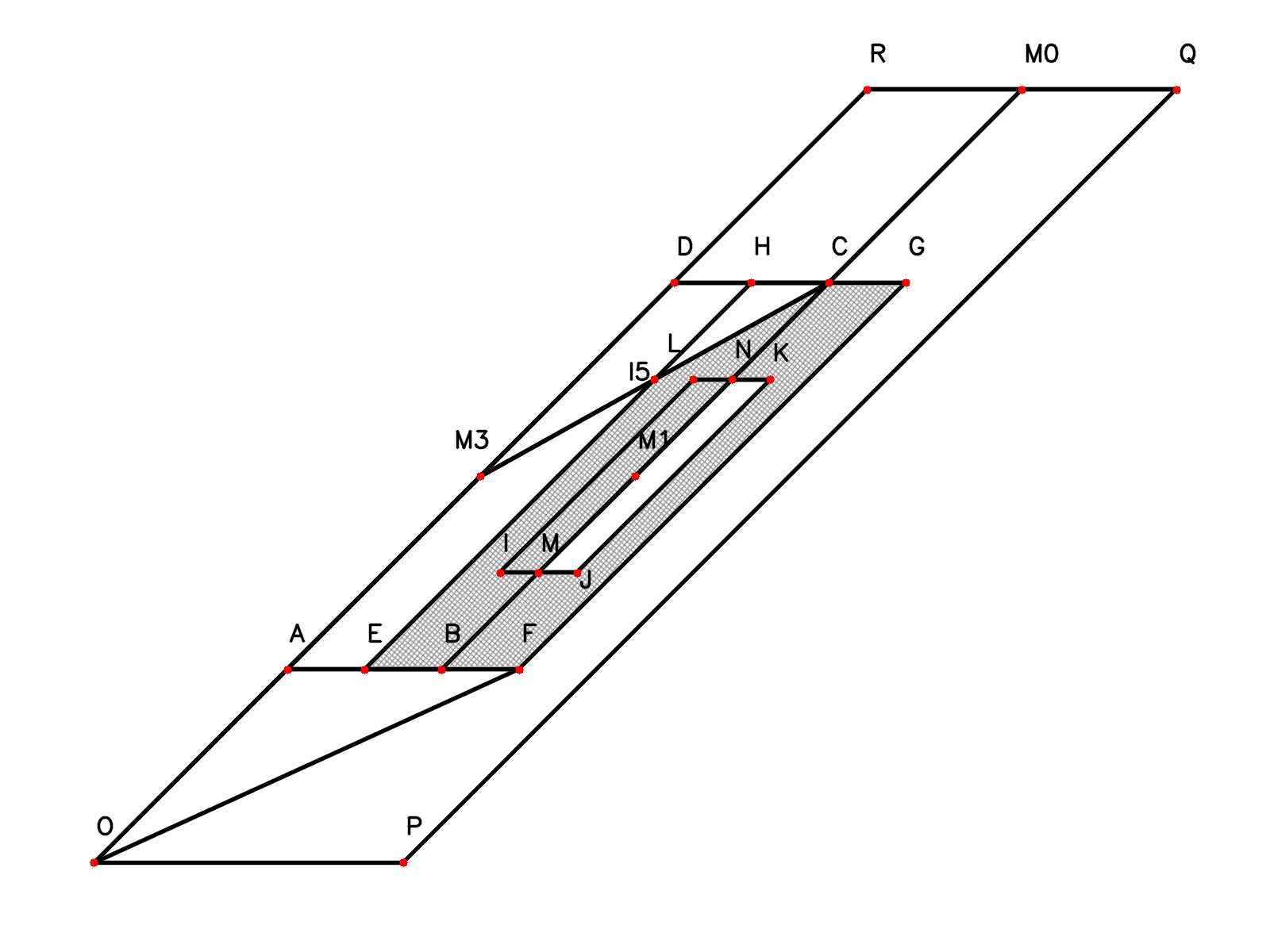}\hfill
    \includegraphics[width=0.242\linewidth]{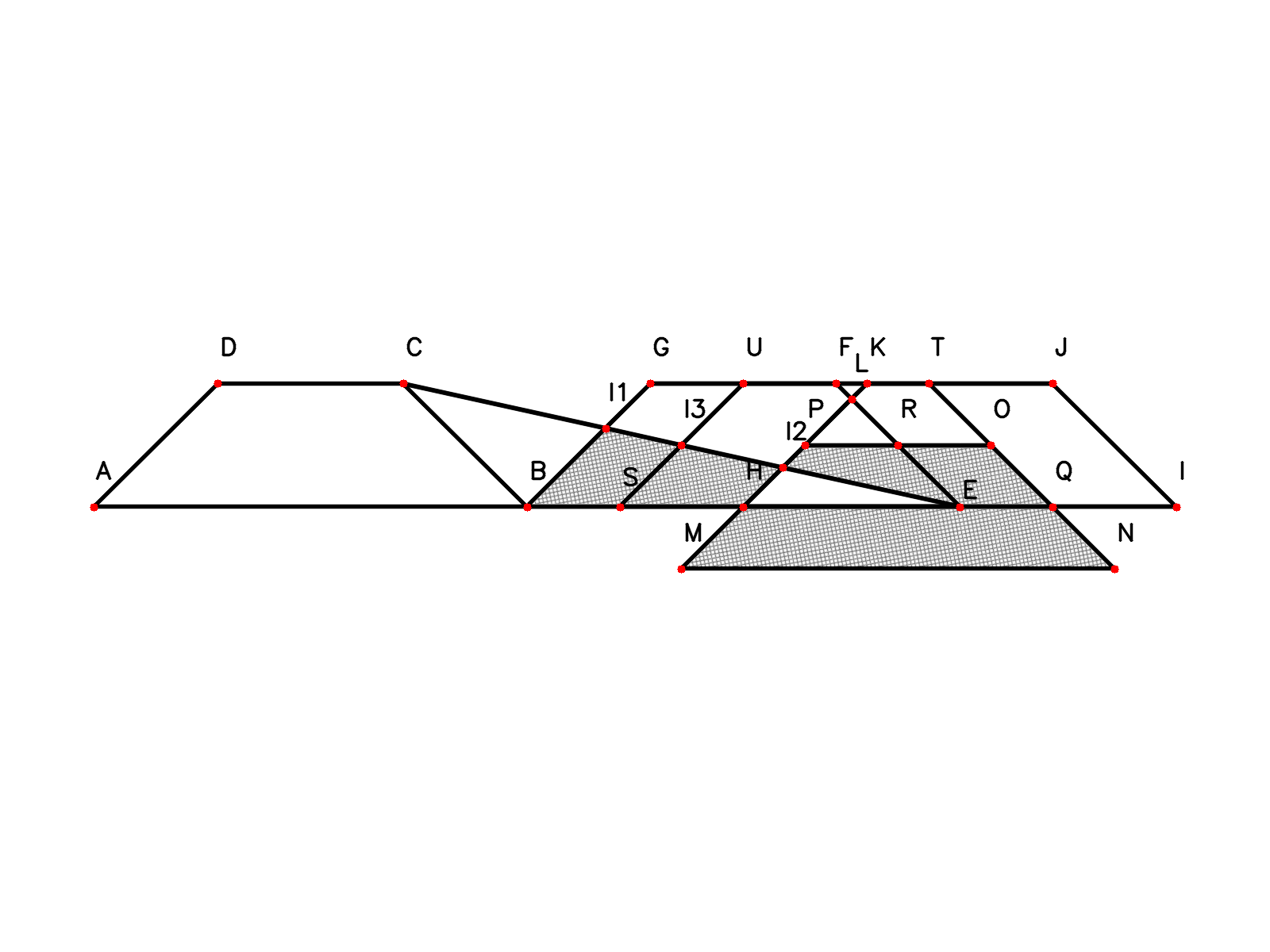}\hfill
    \includegraphics[width=0.242\linewidth]{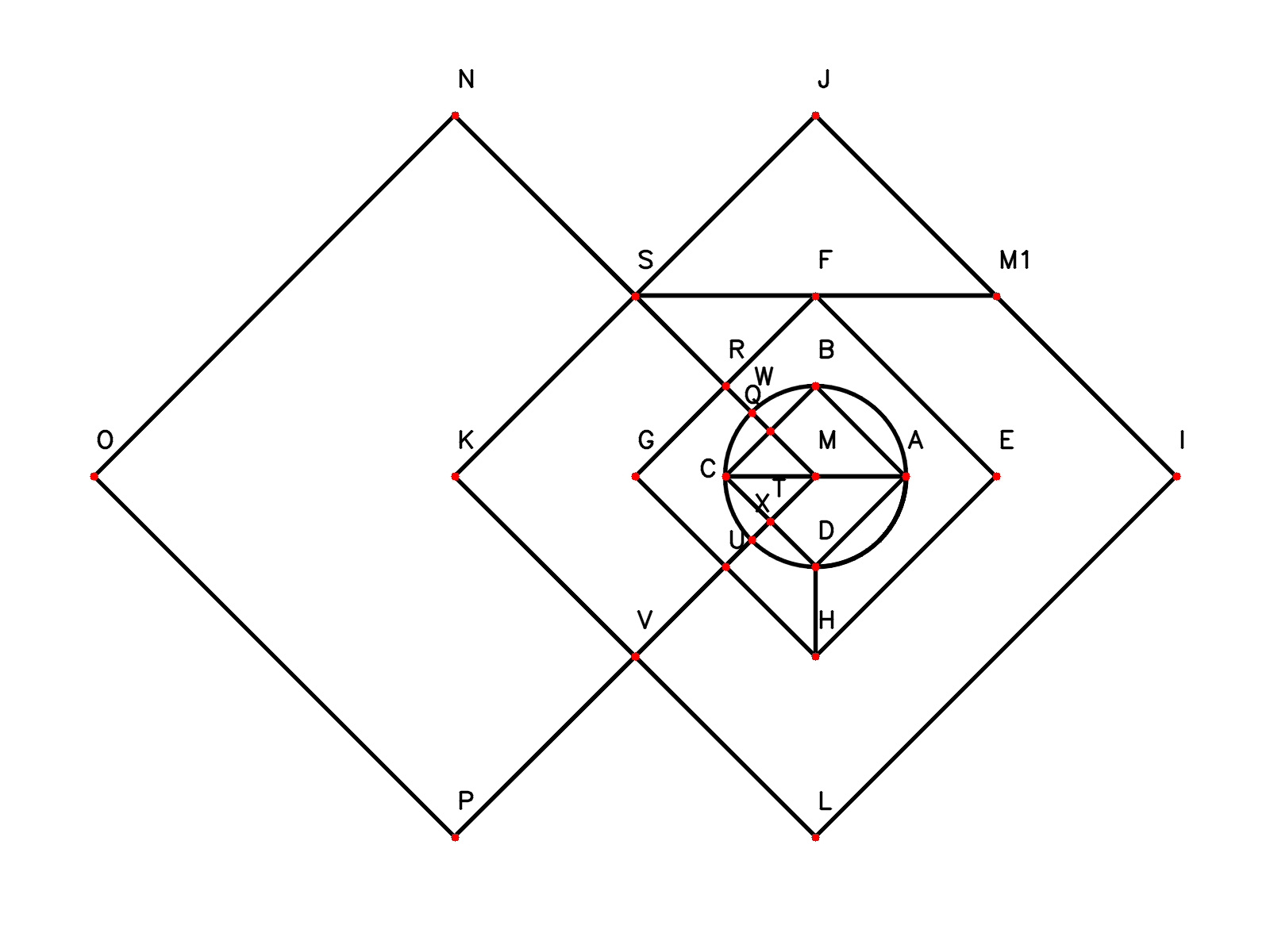}
    
    \vspace{0.1cm} 
    \caption{\textbf{Visualization of rendering quality and topological diversity across GeoSym127K.} The dataset covers extreme geometric variations including multi-hop translations, complex shaded regions, and precision-aligned vertices. Every diagram is generated from exact mathematical coordinates, ensuring zero visual hallucination during model training.}
    \label{fig:geosym_gallery}
    \vspace{0.3cm} 
\end{figure*}

\begin{figure}[b!]
    \centering
    \includegraphics[width=0.6\linewidth]{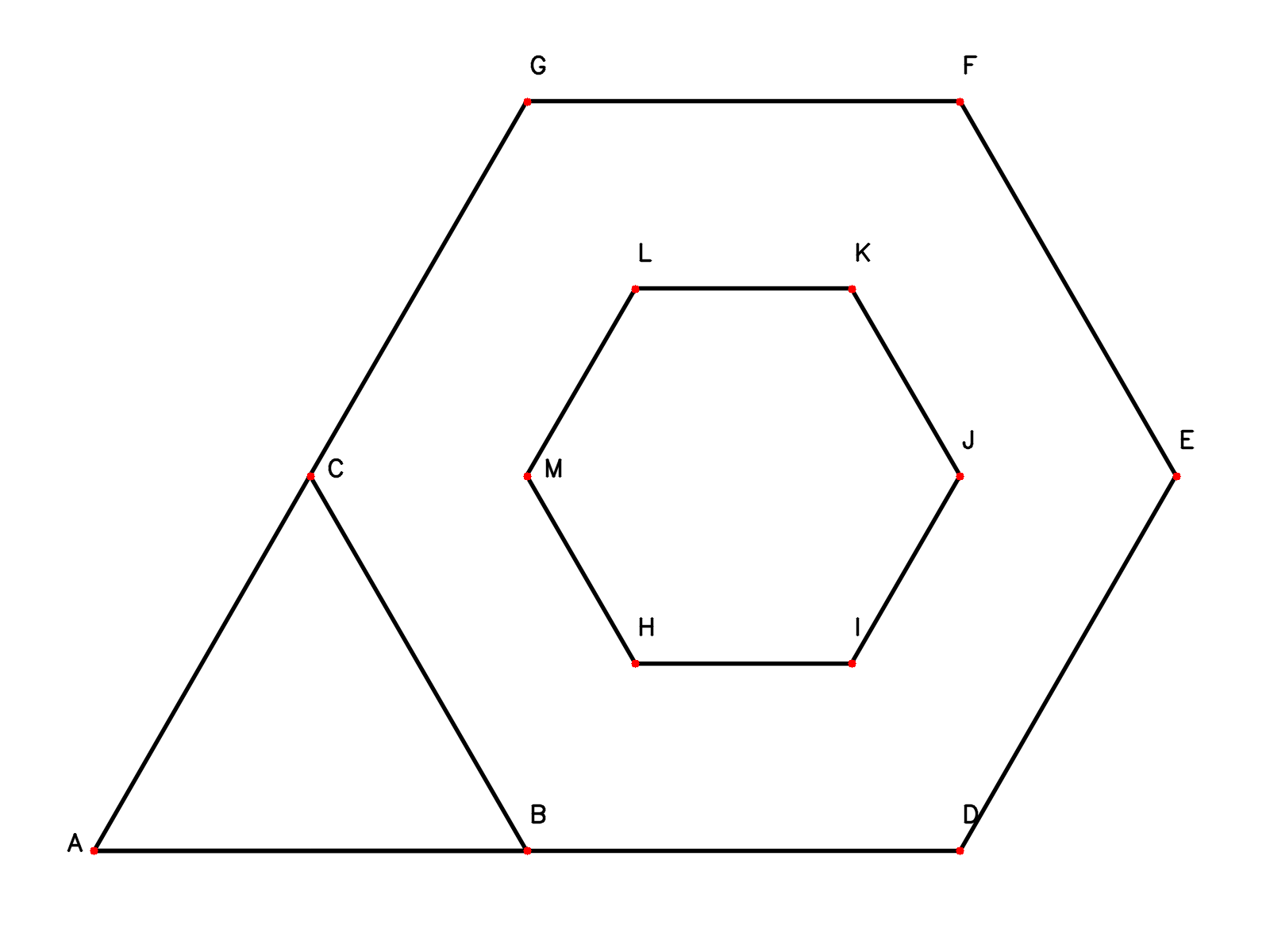}
    \vspace{0.3cm}

    \small
    \renewcommand{\arraystretch}{1.0} 
    \begin{tabular}{@{} >{\bfseries}p{0.18\linewidth} p{0.78\linewidth} @{}}
        \toprule
        \multicolumn{2}{c}{\textbf{Sample Metadata \& Synthesized Rationale}} \\
        \midrule
        Target Question & Calculate the length of Line IJ. \\
        Ground Truth (GT) & $\frac{\sqrt{5}}{2}$ \\
        Difficulty Score & 6.56 (Level 3) \\
        \midrule
        Generated Caption & In the given geometric construction, an isosceles triangle ABC with base AB of length $\sqrt{5}$ and apex angle $\angle ACB = 60^\circ$ serves as the foundational shape. On side BC, a regular hexagon (with center $O_2$) is constructed inward, sharing the same side length as BC and rotated by $-2\pi/3$ relative to the triangle. A second, concentric regular hexagon is then generated by scaling the first hexagon by a factor of $1/2$ about center $O_2$, producing the inner hexagon with vertices H, I, J, K, L, M. Given this nested configuration, calculate the length of segment IJ. \\
        \midrule
        Verified CoT & 
        We are given an isosceles triangle $\triangle ABC$ with base $AB = \sqrt{5}$ and apex angle $\angle ACB = 60^\circ$. Since the apex angle is $60^\circ$, and the triangle is isosceles with $AC = BC$, this triangle is actually \textbf{equilateral}. \newline
        \textit{Why?} In an isosceles triangle with apex angle $60^\circ$, the two base angles must each be $(180^\circ - 60^\circ)/2 = 60^\circ$, so all angles are $60^\circ \rightarrow$ equilateral. \newline
        Thus, all sides of $\triangle ABC$ are equal: $AB = BC = CA = \sqrt{5}$. \newline
        \rule{0pt}{3ex}Next, on side $BC$, a \textbf{regular hexagon} is constructed \textbf{inward}, sharing the same side length as $BC$, i.e., side length $s = \sqrt{5}$, and rotated by $-2\pi/3$ (i.e., $-120^\circ$) relative to the triangle. \newline
        Then, a \textbf{second concentric regular hexagon} is formed by scaling the first one by a factor of $1/2$ about its center $O_2$. The inner hexagon has vertices labeled H, I, J, K, L, M, and we are to find the length of segment IJ. \newline
        \rule{0pt}{3ex}\textbf{Step 1: Understand the inner hexagon} \newline
        Since the inner hexagon is a scaled version of the outer hexagon by factor $1/2$, and both are regular hexagons centered at $O_2$, then \textbf{all side lengths of the inner hexagon are half those of the outer hexagon}. \newline
        The outer hexagon has side length $\sqrt{5}$, so the inner hexagon has side length: $\frac{1}{2} \cdot \sqrt{5} = \frac{\sqrt{5}}{2}$. \newline
        In a regular hexagon, \textbf{each side is equal to the distance between adjacent vertices}. So if I and J are adjacent vertices of the inner hexagon, then: $IJ = \text{side length of inner hexagon} = \frac{\sqrt{5}}{2}$. \newline
        \rule{0pt}{3ex}But wait --- we must confirm that I and J are indeed \textbf{adjacent} vertices. Looking at the diagram labels: the inner hexagon is labeled in order: H, I, J, K, L, M. Assuming they are labeled consecutively around the hexagon (which is standard), then I and J are adjacent. Hence, segment IJ is a side of the inner regular hexagon. \newline
        \rule{0pt}{3ex}\textbf{Final Answer:} \fbox{$\dfrac{\sqrt{5}}{2}$} \\
        \bottomrule
    \end{tabular}
    \caption{\textbf{GeoSym127K Instruct Dataset Example.}}
    \label{fig:dataset_sample_5059}
\end{figure}

\begin{figure}[htbp]
    \centering
    \includegraphics[width=0.6\linewidth]{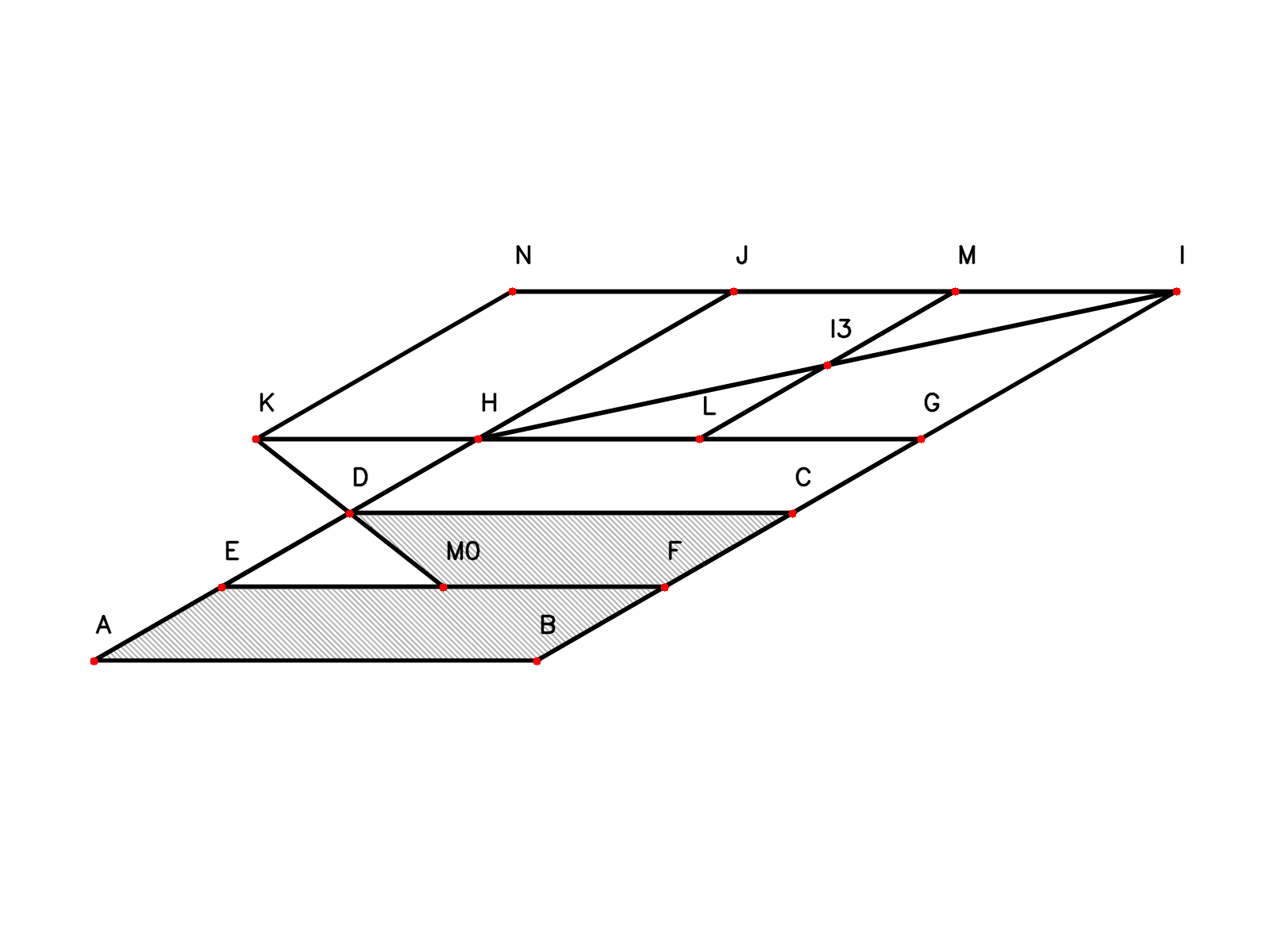}
    \vspace{0.3cm}

    \small
    \renewcommand{\arraystretch}{1.2} 
    \begin{tabular}{@{} >{\bfseries}p{0.18\linewidth} p{0.78\linewidth} @{}}
        \toprule
        \multicolumn{2}{c}{\textbf{Sample Metadata \& Synthesized Rationale}} \\
        \midrule
        Target Question & What is the ratio of the area of shadow region Shadow $FCO_2DM_0$ to the area of parallelogram Parallelogram HGIJ? \\
        Ground Truth (GT) & $3 : 8$ \\
        Difficulty Score & 3.82 \\
        \midrule
        Generated Caption & In the given geometric construction, a base parallelogram (centered at $O_1$) with base length 6, height 2, and an internal angle of $30^\circ$ undergoes two successive translations: first along edge BC by half its length (yielding a new center $O_2$), then along edge FG by its full length (yielding center $O_3$). The resulting figure includes multiple connected parallelograms and intersecting line segments. Key vertices are labeled A through N, with points $M_0$ (midpoint of EF) and $I_3$ (intersection point) explicitly marked. Two regions are shaded with a hatch pattern: one is the polygon $FCO_2DM_0$ (where $O_2$ denotes the center after the first translation), and the other is a separate region not directly involved in the query. Additionally, parallelogram HGIJ is formed as part of the translated structure. Given this configuration, what is the ratio of the area of the hatched region $FCO_2DM_0$ to the area of parallelogram HGIJ? \\
        \midrule
        Verified CoT & 
        We are given a geometric construction involving translations of a base parallelogram, and we are to compute the \textbf{ratio of the area of the hatched region $FCO_2DM_0$} to the \textbf{area of parallelogram HGIJ}.\newline
        \rule{0pt}{3ex}\textbf{Step 1: Understand the Base Parallelogram} \newline
        The base parallelogram (centered at $O_1$) has base length = 6, height = 2. \newline
        Area = base $\times$ height = $6 \times 2 = \textbf{12}$. \newline
        \rule{0pt}{3ex}\textbf{Step 2 \& 3: Translations and Region Analysis} \newline
        The parallelogram undergoes two successive translations. Since the entire construction is built via translations of the same parallelogram, all resulting parallelograms have the \textbf{same area: 12}. The hatched region $FCO_2DM_0$ is a polygon composed of parts of the original and translated parallelograms. \newline
        \rule{0pt}{3ex}\textit{\dots \dots} \newline
        \rule{0pt}{3ex}\textbf{Step 4: Area of Hatched Region} \newline
        Using the derived coordinates, we apply the shoelace formula to the pentagon $F \rightarrow C \rightarrow O_2 \rightarrow D \rightarrow M_0 \rightarrow F$: \newline
        $\text{Area} = \frac{1}{2} |6 + 6 + 6 + (-6) + (-3)| = \frac{1}{2} |9| = \textbf{4.5}$. \newline
        So area of hatched region $FCO_2DM_0 = 4.5$. \newline
        \rule{0pt}{3ex}\textbf{Step 5 \& 6: Compute the Ratio} \newline
        Area of hatched region $FCO_2DM_0 = 4.5$ \newline
        Area of parallelogram HGIJ = 12 \newline
        $\text{Ratio} = 4.5 / 12 = \textbf{3/8}$. \newline
        \rule{0pt}{3ex}\textbf{Final Answer:} \fbox{$\dfrac{3}{8}$} \\
        \bottomrule
    \end{tabular}
    \caption{\textbf{GeoSym127K Instruct Dataset Example.}}
    \label{fig:dataset_sample_2426}
\end{figure}

\begin{figure}[htbp]
    \centering
    \includegraphics[width=0.55\linewidth]{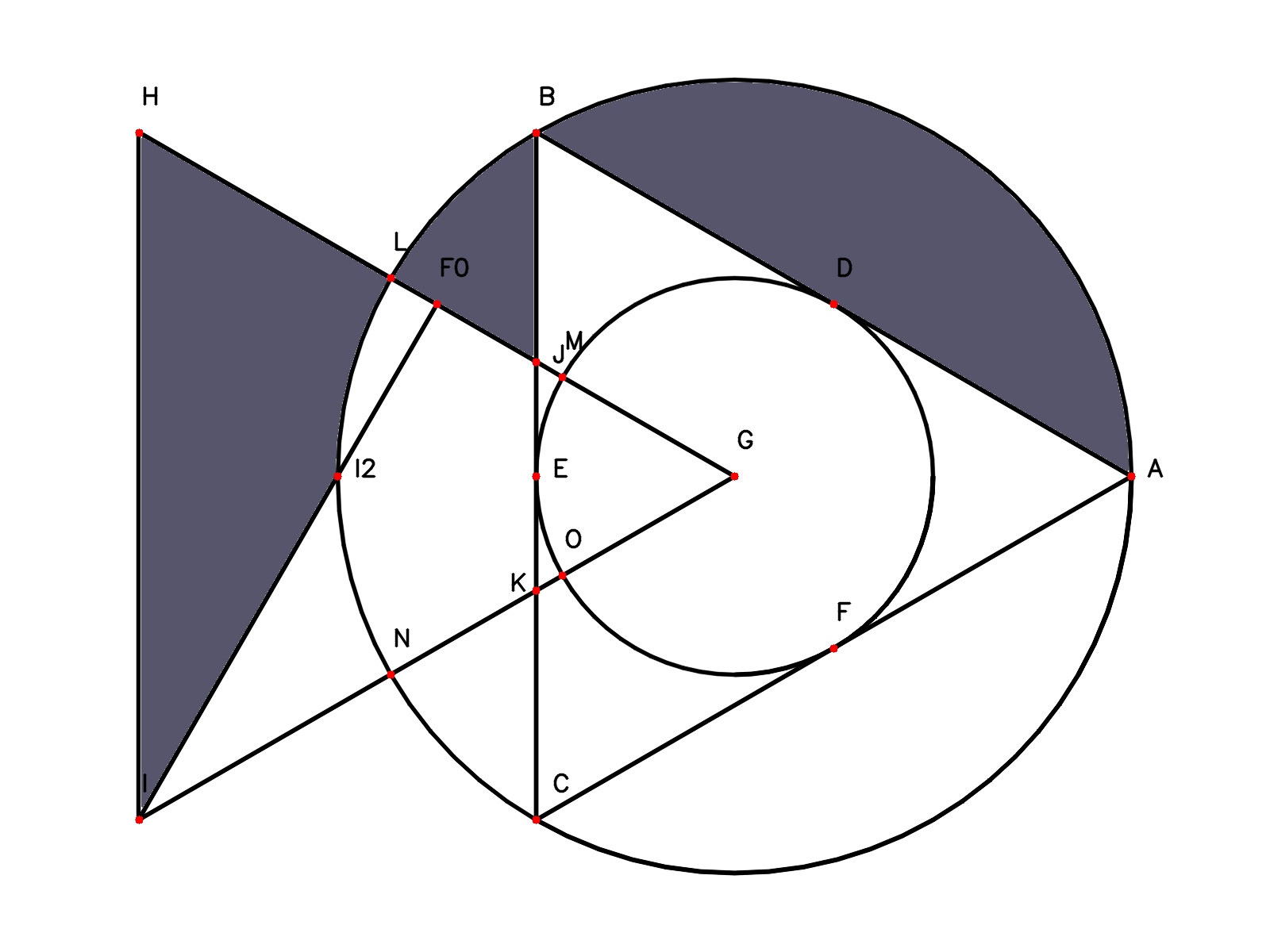}
    \vspace{0.2cm}

    \small
    \renewcommand{\arraystretch}{1.1}
    \begin{tabular}{@{} >{\bfseries}p{0.18\linewidth} p{0.78\linewidth} @{}}
        \toprule
        \multicolumn{2}{c}{\textbf{Sample Metadata \& Synthesized Rationale}} \\
        \midrule
        Target Question & What is the numerical value of the area of Polygon GHI? \\
        Ground Truth (GT) & $\frac{27}{4} \sqrt{3}$ \\
        Difficulty Score & 6.12 \\
        \midrule
        Generated Caption & In the given geometric construction, a large circle centered at point O with radius 3 serves as the base shape. A smaller concentric circle is drawn inside it, centered at the same point O, with radius $3/2$. An equilateral triangle ABC is inscribed in the larger circle such that its vertices lie on the circumference. This triangle is then translated rigidly to the left by a distance equal to twice the radius of the smaller circle (i.e., 3 units), resulting in a new triangle GHI, where G corresponds to A, H to B, and I to C. From vertex I, a perpendicular is dropped to side GH, meeting it at point $F_0$. Three regions are shaded: the circular segment bounded by arc AB and chord AB; the triangular region GHI; and the region bounded by arc BC and chord BC — all shaded in a uniform dark gray. Given this configuration, what is the numerical value of the area of polygon GHI? \\
        \midrule
        Verified CoT & 
        We are given a geometric construction involving two concentric circles and an equilateral triangle inscribed in the larger circle, then translated to form a new triangle GHI. We are asked to find the \textbf{area of polygon GHI}. \newline
        \rule{0pt}{3ex}\textbf{Step 1: Understand the setup.} The large circle is centered at O with radius $R = 3$. The equilateral triangle ABC is inscribed in this large circle. Triangle ABC is translated rigidly to the left by 3 units to form triangle GHI. Since translation preserves shape and size, \textbf{triangle GHI is congruent to triangle ABC}, and thus has the same area. \newline
        \rule{0pt}{3ex}\textbf{Step 2: Area of the inscribed equilateral triangle.} For an equilateral triangle inscribed in a circle of radius $R$, the relationship between side length $s$ and circumradius $R$ is $s = R\sqrt{3}$. \newline
        Given $R = 3$, the side length is $s = 3\sqrt{3}$. \newline
        The area of an equilateral triangle is given by $A = \frac{\sqrt{3}}{4}s^2$. \newline
        Substituting $s$: $A = \frac{\sqrt{3}}{4} \cdot (3\sqrt{3})^2 = \frac{\sqrt{3}}{4} \cdot (27) = \frac{27\sqrt{3}}{4}$. \newline
        \rule{0pt}{3ex}\textbf{Step 3: Confirm GHI area.} Triangle GHI is a strict translation of triangle ABC. Therefore, $\text{Area of } \triangle GHI = \text{Area of } \triangle ABC = \frac{27\sqrt{3}}{4}$. \newline
        \rule{0pt}{3ex}\textbf{Final Answer:} \fbox{$\dfrac{27\sqrt{3}}{4}$} \\
        \bottomrule
    \end{tabular}
    \caption{\textbf{GeoSym127K Instruct Dataset Example.}}
    \label{fig:dataset_sample_5715}
\end{figure}

\begin{figure}[htbp]
    \centering
    \includegraphics[width=0.55\linewidth]{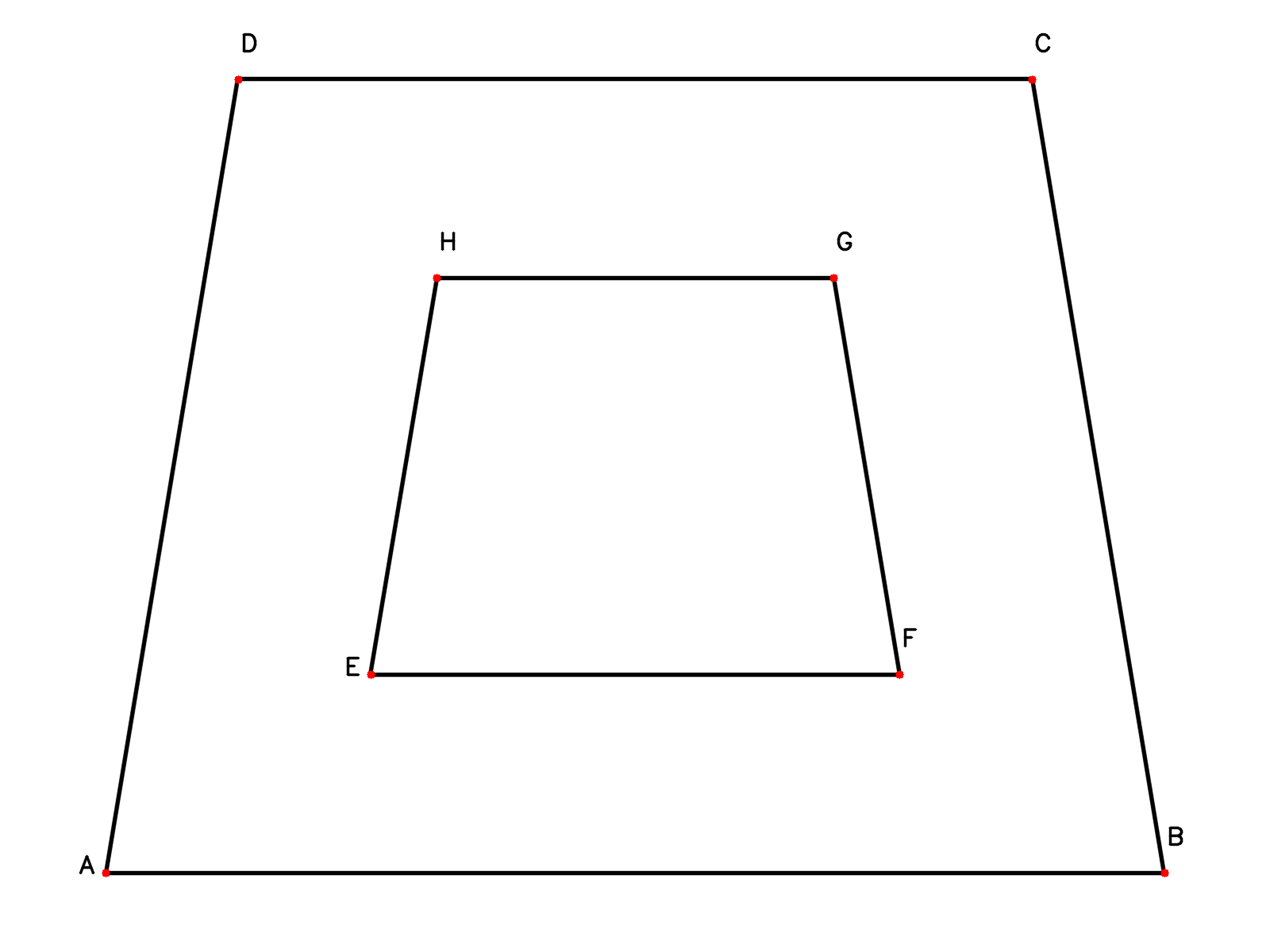}
    \vspace{0.2cm}

    \small
    \renewcommand{\arraystretch}{1.1}
    \begin{tabular}{@{} >{\bfseries}p{0.18\linewidth} p{0.78\linewidth} @{}}
        \toprule
        \multicolumn{2}{c}{\textbf{Sample Metadata \& Synthesized Rationale}} \\
        \midrule
        Target Question & Calculate the perimeter of the trapezoid Trapezoid EFGH in the figure. \\
        Ground Truth (GT) & $\frac{\sqrt{37}}{2} + \frac{7}{2}$ \\
        Difficulty Score & 4.86 \\
        \midrule
        Generated Caption & In the given figure, a larger isosceles trapezoid ABCD contains a smaller, concentric trapezoid EFGH, which is scaled by a factor of $1/2$ relative to ABCD with respect to their common center $O_1$. The outer trapezoid ABCD has bases AB = 4 and CD = 3, with height 3. The inner trapezoid EFGH, being a scaled version, therefore has bases EF = 2 and HG = 3/2, with height 3/2. All corresponding sides are parallel, and the trapezoids share the same axis of symmetry. Calculate the perimeter of trapezoid EFGH in the figure. \\
        \midrule
        Verified CoT & 
        We are given that trapezoid EFGH is a scaled-down version of trapezoid ABCD by a factor of $1/2$. They are concentric and similar, with corresponding sides parallel. \newline
        \rule{0pt}{3ex}\textbf{Step 1: Find the leg length of the outer trapezoid ABCD.} \newline
        Outer trapezoid ABCD is isosceles with bases $AB = 4$, $CD = 3$, and height = 3. \newline
        The difference in base lengths is $AB - CD = 4 - 3 = 1$. This extra length is split equally on both sides, so each horizontal overhang is $0.5$. \newline
        Each non-parallel leg forms the hypotenuse of a right triangle with base = $0.5$ and height = 3. \newline
        Leg length $= \sqrt{0.5^2 + 3^2} = \sqrt{0.25 + 9} = \sqrt{9.25} = \sqrt{\frac{37}{4}} = \frac{\sqrt{37}}{2}$. \newline
        \rule{0pt}{3ex}\textbf{Step 2: Scale down to the inner trapezoid EFGH.} \newline
        Since EFGH is scaled by $1/2$ from ABCD, all linear dimensions are halved. \newline
        The legs of EFGH $= \frac{1}{2} \cdot \frac{\sqrt{37}}{2} = \frac{\sqrt{37}}{4}$. \newline
        \rule{0pt}{3ex}\textbf{Step 3: Compute the perimeter of EFGH.} \newline
        The bases of the inner trapezoid are $EF = 2$ and $HG = 1.5$ (or $3/2$). \newline
        $\text{Perimeter} = EF + HG + 2 \cdot (\text{leg length})$ \newline
        $\text{Perimeter} = 2 + 1.5 + 2 \cdot \left(\frac{\sqrt{37}}{4}\right) = 3.5 + \frac{\sqrt{37}}{2} = \frac{7}{2} + \frac{\sqrt{37}}{2}$. \newline
        \rule{0pt}{3ex}\textbf{Final Answer:} \fbox{$\dfrac{7 + \sqrt{37}}{2}$} \\
        \bottomrule
    \end{tabular}
    \caption{\textbf{GeoSym127K Instruct Dataset Example.}}
    \label{fig:dataset_sample_1541}
\end{figure}

In this section, we present representative samples from the GeoSym127K dataset to demonstrate our rigorous synthesis pipeline. Specifically, Figures~\ref{fig:dataset_sample_5059}, \ref{fig:dataset_sample_2426}, \ref{fig:dataset_sample_5715}, and \ref{fig:dataset_sample_1541} detail the explicit alignment between complex generated topologies, synthesized descriptive captions, and solver-verified Chain-of-Thought (CoT) rationales. Furthermore, Figure~\ref{fig:geosym_gallery} provides a comprehensive visual gallery, illustrating the extensive topological diversity and high-precision rendering quality maintained across the entire dataset.

\begin{figure*}[htbp]
    \centering
    \begin{subfigure}{\textwidth}
        \centering
        \includegraphics[width=\textwidth]{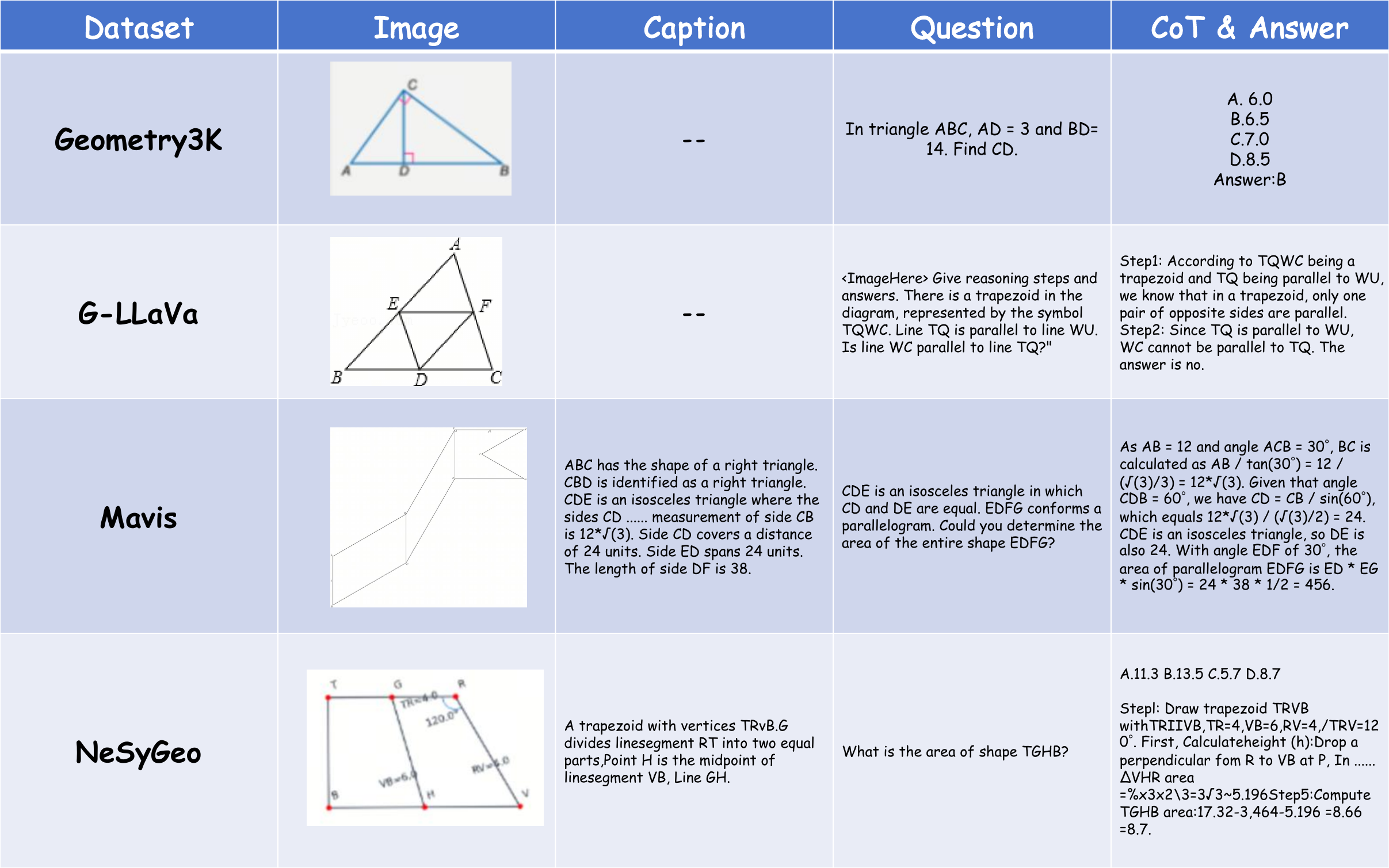}
        \caption{Comparison with Geometry3K, G-LLaVA, Mavis, and NeSyGeo.}
        \label{fig:dataset_comparison_a}
    \end{subfigure}

    \vspace{0.4em}

    \begin{subfigure}{\textwidth}
        \centering
        \includegraphics[width=\textwidth]{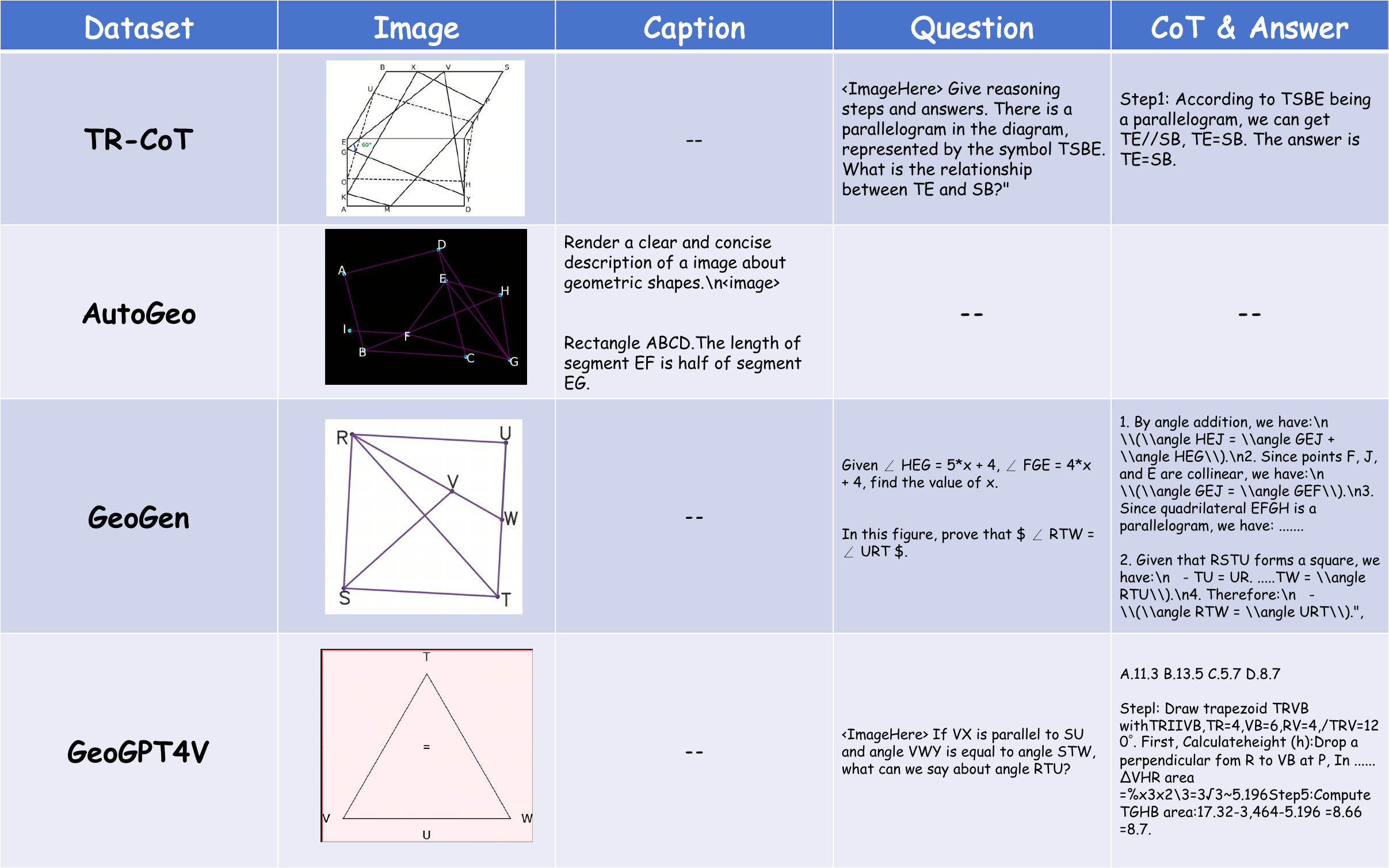}
        \caption{Comparison with TR-CoT, AutoGeo, GeoGen, and GeoGPT4V.}
        \label{fig:dataset_comparison_b}
    \end{subfigure}

    \caption{
    Comparison of representative geometry-related datasets in terms of diagram image, caption availability, question formulation, and CoT-style answer supervision.
    }
    \label{fig:dataset_comparison}
\end{figure*}

To further contextualize GeoSym127K among existing geometry-oriented datasets, we provide a comparative overview in Figure~\ref{fig:dataset_comparison}. Existing datasets exhibit clear differences in their supervision formats and annotation completeness. Geometry3K mainly provides geometric diagrams with questions and answers, but lacks explicit captions and detailed reasoning traces. G-LLaVA and TR-CoT emphasize visual reasoning prompts and CoT-style answers, while their captions are often absent. Mavis and AutoGeo include richer geometric descriptions, but the alignment among diagram structure, textual caption, question, and verified solution remains limited. NeSyGeo, GeoGen, and GeoGPT4V further explore symbolic reasoning or geometry problem generation, yet their samples still show uneven coverage across image, caption, question, and CoT supervision. In contrast, GeoSym127K is designed to jointly preserve high-precision geometric renderings, explicit topology-aware captions, automatically generated questions, and answer-verified CoT rationales, thereby offering a more complete and controllable benchmark for geometry understanding and reasoning.

\clearpage
\section{Extended details of the GeoSym framework}
\label{app:extended_framework}

This section provides comprehensive technical specifications, algorithmic workflows, and mathematical proofs deferred from the main text to ensure the exact reproducibility of the GeoSym engine.

\subsection{GeoSym evolutionary grammar specification}
\label{app:grammar_spec}

To ensure the generated geometries are both mathematically valid and structurally diverse, GeoSym employs a rigid type-conditional probabilistic grammar. Entities are categorized, and subsequent generative operations are strictly constrained by the parent entity's type.

\vspace{1mm}
\noindent \textbf{Category 1: Base primitives (Level 0).}
The manifold is initialized with axiomatic primitives. These include \textbf{Circles} (defined by a center and radius), \textbf{Regular Polygons} (defined by a circumcircle and vertex count $n \ge 3$), \textbf{Triangles} (general, isosceles, right, equilateral), and \textbf{Quadrilaterals} (rectangles, parallelograms, trapezoids). Their coordinates are instantiated as irreducible symbolic constants.

\noindent \textbf{Category 2: Evolutionary operators.}
These operations generate major topological shifts. They include \textbf{Concentric Scaling} (generating similar figures), \textbf{Rigid Transformations} (translations, rotations, reflections around derived axes), \textbf{Circumscription \& Inscription} (strictly locking vertices to curve boundaries), and \textbf{Extension} (projecting polygon edges to form external intersections).

\noindent \textbf{Category 3: Constructive augmentation operators.}
Simulating human problem-design heuristics, the \textit{Builder} module applies localized augmentations. These include \textbf{Vertex/Midpoint Connections} (linking disconnected nodes), \textbf{Perpendicular \& Parallel Constructions} (creating constrained lines relative to a baseline), and \textbf{Diameter Constructions} (forcing chords to pass through circle centers).

\begin{table}[htbp]
    \centering
    \small 
    \caption{\textbf{Statistics of Type-Conditional Topological Evolution.} Overview of base shapes and their corresponding prominent evolutionary processes governed by the GeoSym grammar.}
    \label{tab:evolution_stats}
    \begin{tabular}{@{} l l @{}}
    \toprule
    \textbf{Base Shape} & \textbf{Evolutionary Process ($\mathcal{OP}$)}  \\
    \midrule
    Circle & Concentric Scaling, Inscription, Sector Derivation \\
    Triangle & Circumscription, Altitude Projection, Median Connection  \\
    Trapezoid & Boundary Splicing, Constrained Translation, Extension  \\
    Regular Polygon & Vertex Connection, Radial Projection, Incircle \\
    \bottomrule
    \end{tabular}
\end{table}

\subsection{Dynamic generation and visual grounding algorithms}
\label{alg:generation_process}
\label{app:rendering_algo}

\textbf{Recursive generation process.}
The evolution of the geometric manifold $\mathcal{G}$ follows a strict iterative protocol. At step $t$, the system samples a parent entity $e_{\text{parent}}$ from the active set $\mathcal{E}_t$. Based on its specific geometric type, an operator $\mathcal{OP}$ is sampled from the allowed grammar (Appendix~\ref{app:grammar_spec}) to derive new spatial relations. Before integration, GeoSym computes all pairwise intersections between the newly proposed entities and the existing set $\mathcal{E}_t$ using SymPy algebraic solvers. Only algebraically valid, non-degenerate elements are added to $\mathcal{P}_{t+1}$ and $\mathcal{E}_{t+1}$, advancing the logical level $\mathcal{L}$ accordingly.

\textbf{Visual-first grounding pipeline.}
To define complex shaded regions, GeoSym maps visual pixels back to symbolic logic. \textbf{(1) Binarization and extraction:} The symbolic graph $\mathcal{G}$ is rasterized without labels into a binary line-art image. Connected Component Analysis (CCA) extracts all completely enclosed white regions (Blobs). \textbf{(2) Contour mapping:} For each extracted Blob, we traverse its perimeter pixels and map them to the underlying symbolic segments/arcs in $\mathcal{E}$ via nearest-neighbor distance thresholding. \textbf{(3) Topological verification:} The mapped sequence of symbolic curves must form a mathematically valid closed loop. Valid loops are officially registered as new `Region` entities in $\mathcal{G}$, enabling the SymGT solver to compute their exact analytic properties.

\subsection{The generalized symbolic shoelace algorithm}
\label{app:area_solver}

Standard computational geometry algorithms operate on rectilinear polygons. To compute the exact symbolic area of complex regions $\Omega$ bounded by mixed curves (line segments and circular arcs), SymGT utilizes a generalized topological compensation method over the symbolic field.

Let the boundary of region $\Omega$ consist of an ordered sequence of curves $C = (c_1, c_2, \dots, c_k)$ connecting vertices $(v_1, v_2, \dots, v_k)$. We decompose the area calculation into a base rectilinear polygon area $A_{\text{poly}}$ and a non-linear boundary compensation $A_{\text{arc}}$.

\vspace{1mm}
\noindent \textbf{Rectilinear baseline formulation:}
Treating all curves in $C$ as straight line segments connecting $v_i$ to $v_{i+1}$, we apply the exact symbolic Shoelace formula to derive the base polygonal area:
\begin{equation}
    A_{\text{poly}} = \frac{1}{2} \left| \sum_{i=1}^{k} \left( x_i y_{i+1} - x_{i+1} y_i \right) \right|
\end{equation}
where coordinates $(x, y)$ are maintained as analytic expression trees, and $v_{k+1} \equiv v_1$.

\noindent \textbf{Topological curve compensation:}
For every element $c_i \in C$ that is a circular arc rather than a straight segment, we calculate the exact area of the circular segment bounded by the arc and its chord $v_i v_{i+1}$. Let $r_i$ be the radius and $\theta_i$ be the central angle of the arc. The symbolic area of this compensation segment is:
\begin{equation}
    A_{\text{seg}, i} = \frac{1}{2} r_i^2 (\theta_i - \sin \theta_i)
\end{equation}

To aggregate the total area, SymGT dynamically evaluates the winding direction of the arc relative to the region's interior. If the arc is convex (bulging outward from the region center), the area is added; if concave (biting inward), it is subtracted. The final absolute mathematical ground truth is thus defined as:
\begin{equation}
    A_{\text{total}} = A_{\text{poly}} + \sum_{c_i \in C_{\text{arcs}}} \text{sgn}(c_i) \cdot A_{\text{seg}, i}
\end{equation}
where $\text{sgn}(c_i) \in \{+1, -1\}$ denotes the topological winding polarity.

    \vspace{-0.3cm}
\subsection{Prompt Templates}
\label{app:prompt_templates}

GeoSym uses \textbf{Qwen3-VL-235B-Instruct} as the teacher model for both problem-text synthesis and Chain-of-Thought (CoT) generation. The two stages use different decoding configurations according to their respective objectives. The caption-generation stage requires richer and more descriptive language synthesis, and therefore uses temperature $=0.6$, top-$p=0.95$, and max tokens $=32{,}768$. The CoT-generation stage focuses on stable mathematical derivation and answer formatting, and therefore uses a lower temperature of $0.3$, top-$p=0.95$, and max tokens $=16{,}384$. The detailed configurations are summarized in Table~\ref{tab:prompt_generation_config}.

    \vspace{-0.3cm}
\begin{table}[htbp]
    \centering
    \small
    \caption{\textbf{Teacher-model configurations for prompt-based synthesis.} The caption-generation and CoT-generation stages use different decoding parameters.}
    \label{tab:prompt_generation_config}
    \begin{tabular}{@{} l c c c c @{}}
        \toprule
        \textbf{Stage} & \textbf{Model} & \textbf{Temperature} & \textbf{Top-$p$} & \textbf{Max Tokens} \\
        \midrule
        Caption Generation & Qwen3-VL-235B-Instruct & 0.6 & 0.95 & 32,768 \\
        CoT Generation & Qwen3-VL-235B-Instruct & 0.3 & 0.95 & 16,384 \\
        \bottomrule
    \end{tabular}
    \vspace{0cm}
\end{table}

    \vspace{-0.1cm}
\textbf{Caption-generation prompt.}
The first stage converts the rendered geometric image and GeoSym-generated symbolic metadata into a coherent natural-language mathematical problem. Each request contains the rendered image and two structured textual fields: \texttt{Question Reference}, which denotes the solver-generated target question, and \texttt{Full Geometric Description}, which records the complete geometric construction trajectory and symbolic relations generated by GeoSym. The teacher model is instructed to synthesize these inputs into a rigorous problem stem and a concise final question without changing the queried object (Table~\ref{tab:caption_prompt_template}).

\begin{table*}[b!]
    \centering
    \small
    \caption{\textbf{Caption-generation prompt template.} The prompt asks the teacher model to verbalize the GeoSym-generated symbolic construction into a complete mathematical problem statement.}
    \label{tab:caption_prompt_template}
    \resizebox{\linewidth}{!}{
    \begin{tabular}{@{} p{2.8cm} p{12.5cm} @{}}
        \toprule
        \textbf{Component} & \textbf{Content} \\
        \midrule
        Input & Rendered geometry image; \texttt{Question Reference}; \texttt{Full Geometric Description} \\
        \midrule
        Prompt &
        You are an expert mathematics content creator. Your task is to generate a complete math problem based on the provided image and metadata. Please follow these two steps: 1. \textbf{Caption Generation (Problem Stem)}: Analyze the image along with the \texttt{Full Geometric Description} in Input Data. Synthesize this information into a clear, rigorous, and descriptive mathematical problem statement. Describe the geometric construction, the relationships between shapes, such as translation, connection, polygon on one side, and any shaded regions if any, strictly using the labels, such as A, B, E, shown in the image. 2. \textbf{Question Refinement}: Read the \texttt{Question Reference} in Input Data. Rewrite it into a standard, concise English mathematical question that naturally follows the stem generated in Step 1. You must ensure the object being calculated, such as length, perimeter, area, or specific angle, remains exactly the same as the original. Final Output Requirement: Provide a cohesive textbf that acts as the full problem text, including both Context and Question. \texttt{Question Reference: \{item['question']\}}. \texttt{Full Geometric Description: \{item['description']\}}. \\
        \midrule
        Output & A single cohesive textbf containing the geometric context and the final mathematical question. \\
        \bottomrule
    \end{tabular}
    }
\end{table*}

Formally, the caption-generation message for each sample is defined as
\begin{equation}
    \mathcal{M}_{\mathrm{cap}} =
    \left[
    I,\ Q_{\mathrm{ref}},\ D_{\mathrm{geo}},\ P_{\mathrm{cap}}
    \right],
\end{equation}
where $I$ denotes the rendered diagram, $Q_{\mathrm{ref}}$ is the GeoSym-generated reference question, $D_{\mathrm{geo}}$ is the symbolic geometric description, and $P_{\mathrm{cap}}$ is the caption-generation instruction. The output of this stage is denoted as $Q_{\mathrm{gen}}$, which serves as the natural-language problem text for subsequent CoT generation.

\textbf{CoT-generation prompt.}
The second stage uses the generated problem text and the rendered image to produce a complete step-by-step solution. In this stage, the input question is taken from the \texttt{generated\_question} field. If this field is unavailable, the original \texttt{question} field is used as a fallback. To facilitate automatic answer extraction and symbolic verification, the teacher model is explicitly required to place the final answer inside a LaTeX \texttt{\textbackslash boxed\{\}} expression (Table~\ref{tab:cot_prompt_template}).

\begin{table*}[t!]
    \centering
    \small
    \caption{\textbf{CoT-generation prompt template.} The prompt asks the teacher model to solve the generated problem and output a final answer in a standardized boxed LaTeX format.}
    \label{tab:cot_prompt_template}
    \resizebox{\linewidth}{!}{
    \begin{tabular}{@{} p{2.8cm} p{12.5cm} @{}}
        \toprule
        \textbf{Component} & \textbf{Content} \\
        \midrule
        Input & Rendered geometry image; generated problem text \\
        \midrule
        Prompt &
        \texttt{Question: \{q\_text\}} At the end of your response, place the final numerical answer inside a LaTeX box using \texttt{\textbackslash boxed\{\}}. Make sure that the final answer uses LaTeX-style expressions and is wrapped in \texttt{\textbackslash boxed\{\}}. \\
        \midrule
        Output & A complete CoT solution with the final answer formatted as \texttt{\textbackslash boxed\{answer\}}. \\
        \bottomrule
    \end{tabular}
    }
\end{table*}

The CoT-generation message is formulated as
\begin{equation}
    \mathcal{M}_{\mathrm{cot}} =
    \left[
    I,\ Q_{\mathrm{gen}},\ P_{\mathrm{cot}}
    \right],
\end{equation}
where $I$ is the rendered geometry image, $Q_{\mathrm{gen}}$ is the generated problem text, and $P_{\mathrm{cot}}$ is the final-answer formatting instruction. The resulting teacher response is denoted as $R_{\mathrm{cot}}$.

To ensure that the generated rationales are verifiable, we first discard responses that do not contain a valid \texttt{\textbackslash boxed\{\}} expression. For the remaining samples, the boxed answer is extracted as $A_{\mathrm{pred}}$ and compared with the solver-derived symbolic ground truth $A_{\mathrm{GT}}$. A CoT sample is retained only if
\begin{equation}
    \mathrm{Simplify}(A_{\mathrm{pred}} - A_{\mathrm{GT}}) \equiv 0.
\end{equation}
This verification rule ensures that the final instruction-tuning set contains both visually grounded problem statements and answer-verified reasoning trajectories.

\subsection{Algorithm}
\label{app:Algorithm}

To formally encapsulate the exact execution logic of the GeoSym framework discussed in Section~\ref{subsec:geosym_pipeline}, we provide the complete algorithmic pseudo-code in Algorithm~\ref{alg:geosym_pipeline}. This algorithm details the step-by-step progression through our four core synthesis modules: \textbf{(1) Builder:} executing type-conditional topological evolution upon an arbitrary-precision manifold; \textbf{(2) Drawer:} ensuring strict visual-symbolic grounding via Connected Component Analysis for complex shaded regions; \textbf{(3) GT Solver:} performing analytic derivations of exact mathematical properties; and \textbf{(4) Generator:} applying the deterministic $\text{Simplify}(A_{pred} - A_{GT}) \equiv 0$ verification to filter MLLM-generated rationales. By strictly executing this automated closed-loop pipeline, GeoSym guarantees the absolute mathematical fidelity of every generated $(Image, Question, CoT)$ triplet, entirely circumventing the heuristic noise inherent in traditional LLM-based data annotations.

\begin{algorithm}[htbp]
\caption{GeoSym Synthesis and Verification Pipeline}
\label{alg:geosym_pipeline}
\begin{algorithmic}[1] 
\Require Type-conditional grammar $\mathcal{OP}$, Max derivation depth $D_{\max}$, Teacher MLLM $\mathcal{M}$
\Ensure Verified dataset $\mathcal{D}_{\text{verified}}$ containing $(I, Q_{\text{gen}}, R_{\text{cot}}, A_{\text{GT}})$ tuples

\State $\mathcal{D}_{\text{verified}} \leftarrow \emptyset$

\While{Target Dataset Size Not Reached}
    \State \textbf{\% Phase 1: Type-Conditional Topological Evolution (Builder)}
    \State Initialize arbitrary-precision manifold $\mathcal{G} = \langle \mathcal{P}, \mathcal{E}, \Phi, \mathcal{L}, \mathcal{T} \rangle$
    \For{$t = 1$ \textbf{to} $D_{\max}$}
        \State $e_{\text{parent}} \leftarrow \text{Sample}(\mathcal{E}_{t-1})$
        \State $op \leftarrow \text{Sample}(\mathcal{OP} \mid \text{Type}(e_{\text{parent}}))$ \Comment{Type-conditional sampling}
        \State $\mathcal{E}_{\text{new}}, \mathcal{P}_{\text{new}} \leftarrow \text{Apply}(op, e_{\text{parent}})$
        \State $\mathcal{P}_{\text{inter}} \leftarrow \text{SymPy.SolveIntersections}(\mathcal{E}_{\text{new}}, \mathcal{E}_{t-1})$
        \If{$\text{IsValidAndNonDegenerate}(\mathcal{E}_{\text{new}}, \mathcal{P}_{\text{inter}})$}
            \State $\mathcal{G}.\text{Update}(\mathcal{P}_{\text{new}} \cup \mathcal{P}_{\text{inter}}, \mathcal{E}_{\text{new}})$ \Comment{Advance logical level $\mathcal{L}$ and trajectory $\mathcal{T}$}
        \EndIf
    \EndFor

    \State \textbf{\% Phase 2: Visual-First Grounding (Drawer \& Shader)}
    \State $I_{\text{binary}} \leftarrow \text{RasterizeLineArt}(\mathcal{E})$
    \State $\mathcal{B} \leftarrow \text{ConnectedComponentAnalysis}(I_{\text{binary}})$ \Comment{Extract blobs}
    \For{\textbf{each} $b \in \mathcal{B}$}
        \State $\mathcal{C}_{\text{loop}} \leftarrow \text{MapToSymbolicCurves}(b, \mathcal{E})$
        \If{$\text{IsMathematicallyClosedLoop}(\mathcal{C}_{\text{loop}})$}
            \State $\mathcal{E} \leftarrow \mathcal{E} \cup \{ \text{Region}(\mathcal{C}_{\text{loop}}) \}$ \Comment{Instantiate Shaded Block}
        \EndIf
    \EndFor
    \State $I \leftarrow \text{RenderFinalDiagram}(\mathcal{G})$

    \State \textbf{\% Phase 3: SymGT Solver}
    \State $e_{\text{target}} \leftarrow \text{TailBiasedSample}(\mathcal{E})$ \Comment{Force multi-hop reasoning}
    \State $A_{\text{GT}} \leftarrow \text{SymPy.CalculateExact}(\Phi, e_{\text{target}})$ \Comment{e.g., Symbolic Shoelace for Area}

    \State \textbf{\% Phase 4: Instruction Synthesis and Verification (Generator)}
    \State $Q_{\text{gen}} \leftarrow \mathcal{M}.\text{GenerateCaption}(I, \mathcal{T}, e_{\text{target}})$ \Comment{Temp=0.6 for diversity}
    \State $R_{\text{cot}}, A_{\text{pred}} \leftarrow \mathcal{M}.\text{GenerateCoT}(I, Q_{\text{gen}})$ \Comment{Temp=0.3 for logic}
    
    \If{$\text{SymPy.Simplify}(A_{\text{pred}} - A_{\text{GT}}) \equiv 0$}
        \State $\mathcal{D}_{\text{verified}} \leftarrow \mathcal{D}_{\text{verified}} \cup \{ (I, Q_{\text{gen}}, R_{\text{cot}}, A_{\text{GT}}) \}$
    \EndIf
\EndWhile

\State \Return $\mathcal{D}_{\text{verified}}$
\end{algorithmic}
\end{algorithm}

\begin{algorithm}[htbp]
\caption{Deterministic Answer Verification via Algebraic Equivalence}
\label{alg:verification}
\begin{algorithmic}[1] 
\Require MLLM generated textual response $R$, Ground Truth dictionary $D_{GT}$
\Ensure Boolean validation flag $V$ (True if equivalent, False otherwise)

\State $V \leftarrow \text{False}$
\State $A_{pred} \leftarrow \text{ExtractAnswer}(R)$ \Comment{Extract via regex matching \texttt{\textbackslash boxed\{\}}}
\If{$A_{pred}$ is \textbf{None}}
    \State \Return $V$
\EndIf

\State $G \leftarrow \text{GetGroundTruths}(D_{GT})$ \Comment{Extract both \texttt{expr} and \texttt{latex} variants}

\For{each $A_{GT} \in G$}
    \State $A_{GT}^{boxed} \leftarrow \text{Concat}(\text{\boxed\{}, A_{GT}, \text{\}})$ \Comment{Try: Evaluate algebraic equivalence}
    \State $S \leftarrow \text{MathVerify}(A_{pred}, A_{GT}^{boxed})$ \Comment{Evaluate algebraic equivalence}
    \If{$S > 0$}
        \State $V \leftarrow \text{True}$
        \State \textbf{break} \Comment{Match found, early exit}
    \Else
        \State \Comment{Catch TimeoutException: Reject on timeout}
        \State $S \leftarrow 0$ \Comment{Reject on timeout}
    \EndIf
\EndFor

\State \Return $V$
\end{algorithmic}
\end{algorithm}

\clearpage

\section{Detailed Dataset Statistics}
\label{app:dataset_stats_and_verification}

\subsection{Configuration and Hyperparameter Settings}
\label{app:config}

The GeoSym framework achieves rigorous difficulty stratification by tuning structural and visual parameters. Table~\ref{tab:hyperparameters} summarizes the key configurable parameters, highlighting the progression of complexity across the \textit{Entry}, \textit{Hard}, and \textit{Expert} generative settings.

\begin{table*}[htbp]
    \centering
    \small
    \caption{\textbf{Key Configurable Parameters of the GeoSym Pipeline.} Comparison of settings across Entry, Hard, and Expert modes, illustrating how topological depth, visual complexity, and task distribution are logically controlled.}
    \label{tab:hyperparameters}
    \begin{tabular}{@{} l l | >{\centering\arraybackslash}p{2.2cm} | >{\centering\arraybackslash}p{2.2cm} | >{\centering\arraybackslash}p{2.2cm} @{}}
    \toprule
    \textbf{Module} & \textbf{Parameter} & \textbf{Entry Setting} & \textbf{Hard Setting} & \textbf{Expert Setting} \\
    \midrule
    \textbf{Global} & Target Quantity ($n$) & 20,000 & 10,000 & 5,000 \\
    & Max Points / Lines Limit & 30 / 40 & 40 / 60 & 50 / 80 \\
    \midrule
    \textbf{Evolution} & Base Shape Types & \multicolumn{3}{>{\centering\arraybackslash}p{7.2cm}}{Polygon, Circle, Special Tri/Rect, Parallel, Trapezoid} \\
    (\textit{Template}) & Derivation Rounds Range & \textbf{1 -- 2} & \textbf{2 -- 4} & \textbf{3 -- 5} \\
    \midrule
    \textbf{Builder} & Max Enhancement Rounds & \textbf{3} & \textbf{5} & \textbf{7} \\
    & Operation Types & \multicolumn{3}{>{\centering\arraybackslash}p{7.2cm}}{Connect Points/Midpoints, Draw Perpendicular/Diameter} \\
    \midrule
    \textbf{Drawer} & Canvas Size & \multicolumn{3}{c}{$1600 \times 1200$ pixels} \\
    & Line Width / Color & \multicolumn{3}{c}{3 px / Black (\#000000)} \\
    \midrule
    \textbf{Shader} & Target Region Count & \textbf{1 -- 1} & \textbf{1 -- 4} & \textbf{1 -- 6} \\
    & Max Fill Attempts & \textbf{3} & \textbf{5} & \textbf{7} \\
    & Shadow Styles & \multicolumn{3}{c}{Hatch, Solid, Crosshatch, Gradient} \\
    \midrule
    \textbf{QA (Task)} & Base Question Types & \multicolumn{3}{>{\centering\arraybackslash}p{7.2cm}}{Length, Angle, Perimeter, Entity/Shadow Area, Ratios} \\
    & \textit{Length} Weight & \textbf{0.3} & \textbf{0.6} & \textbf{0.3} \\
    & \textit{Shadow Area} Weight & \textbf{0.2} & \textbf{0.3} & \textbf{0.2} \\
    & Questions per Geometry & \textbf{1} & \textbf{5} & \textbf{10} \\
    \bottomrule
    \end{tabular}
\end{table*}

\subsection{Multi-dimensional difficulty assessment details}
\label{app:difficulty}

As introduced in Section~\ref{sec:dataset}, the cognitive load required to solve a generated sample is quantified by the total difficulty score $D_{\text{total}}$. This metric underpins the rigorous hierarchical stratification of the GeoSym127K dataset (Entry-Level, Hard-Level, Expert-Level). The complete evaluation framework is a weighted linear combination of three distinct dimensions:
\begin{equation}
\label{eq:difficulty}
    D_{\text{total}} = w_g \cdot \mathcal{C}_{\text{graph}} + w_q \cdot \mathcal{C}_{\text{question}} + w_a \cdot \mathcal{C}_{\text{answer}}
\end{equation}
In our dataset construction, the balancing weights are empirically set as $w_g = 0.3$, $w_q = 0.5$, and $w_a = 0.2$, strictly prioritizing the depth of multi-hop logical questioning over mere visual clutter.

\vspace{1mm}
\noindent \textbf{(1) Graph complexity ($\mathcal{C}_{\text{graph}}$).} 
This dimension aggregates the visual density and the evolutionary depth of the geometric manifold:
\begin{equation}
    \mathcal{C}_{\text{graph}} = \alpha \cdot N_{\text{elements}} + \beta \cdot \bar{L}_{\text{avg}}
\end{equation}
where $N_{\text{elements}}$ is the total number of visual primitives, and $\bar{L}_{\text{avg}}$ is the average topological level. We set $\beta \gg \alpha$ (specifically, $\alpha=0.05, \beta=0.4$) to penalize deep derivational histories significantly more than superficial component quantities.

\noindent \textbf{(2) Question complexity ($\mathcal{C}_{\text{question}}$).} 
This dimension directly evaluates the multi-hop reasoning requirements by assessing the target entity $e_{\text{target}}$ being queried:
\begin{equation}
    \mathcal{C}_{\text{question}} = \mu_{\text{task}} \cdot L(e_{\text{target}})
\end{equation}
The base coefficient $\mu_{\text{task}}$ differentiates inherent domain hardness (e.g., area calculation $\mu_{\text{area}}=1.5$, whereas simple length lookup $\mu_{\text{length}}=1.0$). $L(e_{\text{target}})$ represents the logical depth of the target entity in the generative trajectory $\mathcal{T}$.

\noindent \textbf{(3) Answer complexity ($\mathcal{C}_{\text{answer}}$).} 
To evaluate the algebraic entropy of the final analytic ground truth, we map the symbolic expression length $\|\mathcal{E}\|$ (number of characters in the simplified SymPy string representation) via a normalized power-law function:
\begin{equation}
    \mathcal{C}_{\text{answer}} = 1 + K \cdot \left( \frac{\|\mathcal{E}\| - 1}{N_{\max}} \right)^{\gamma}
\end{equation}
where $N_{\max}$ is a normalization constant representing the 99th percentile of expression lengths (set to 150), $K=5$ is the scaling boundary, and $\gamma=0.6$ is the curvature exponent. This non-linear mapping ensures a gentle gradient for standard rational outputs while applying a strong damping effect to prevent score explosion from overly verbose irrational combinations.

\textbf{Micro-Level Quantile Mapping and Linearity.}
Based on the global distribution of the calculated cognitive load $D_{total}$ evaluated across the entire dataset, we partitioned the data into 10 uniform quantiles. As detailed in Table~\ref{tab:quantiles}, the specific $D_{total}$ upper bounds and their corresponding verification pass rates (evaluated by the base teacher model) demonstrate a strict monotonic decline. This linear degradation in accuracy perfectly validates our multi-dimensional difficulty metric (Equation~\ref{eq:difficulty}), proving that the synthetic cognitive load accurately reflects the true reasoning bottlenecks of current LMMs.

\textbf{Micro-Level Quantile Mapping and Linearity.}
Based on the global distribution of the calculated cognitive load $D_{total}$ evaluated across the entire dataset, we partitioned the data into 10 uniform quantiles. As detailed in Figure~\ref{fig:micro_level_trend} and Table~\ref{tab:quantiles}, the specific $D_{total}$ upper bounds and their corresponding verification pass rates demonstrate a strict monotonic decline. This linear degradation in accuracy perfectly validates our multi-dimensional difficulty metric, proving that the synthetic cognitive load accurately reflects the true reasoning bottlenecks of current LMMs.

\begin{figure}[htbp]
    \centering
    
    \begin{minipage}[t]{0.49\linewidth}
        \centering
        
        \small
        \captionof{table}{\textbf{Global Micro-Level Discretization.} boundaries and pass rates.}
        \label{tab:quantiles}
        \vspace{2mm} 
        \resizebox{\linewidth}{!}{
        \begin{tabular}{@{} l r r c @{}}
        \toprule
        \textbf{Level} & \textbf{Quantile} & \textbf{Upper Bound} & \textbf{Pass Rate} \\ \midrule
        1 & 10\% & $\le 3.0472$ & 55.2\% \\
        2 & 20\% & $\le 3.5651$ & 54.3\% \\
        3 & 30\% & $\le 3.9422$ & 55.2\% \\
        4 & 40\% & $\le 4.3568$ & 50.8\% \\
        5 & 50\% & $\le 4.7553$ & 50.6\% \\
        6 & 60\% & $\le 5.1505$ & 48.4\% \\
        7 & 70\% & $\le 5.6536$ & 44.3\% \\
        8 & 80\% & $\le 6.4718$ & 41.0\% \\
        9 & 90\% & $\le 8.0453$ & 29.0\% \\
        10 & 100\% & $> 8.0453$ & \phantom{0}8.7\% \\ \bottomrule
        \end{tabular}
        }
    \end{minipage}\hfill
    \begin{minipage}[t]{0.45\linewidth}
        \centering
        \captionof{figure}{\textbf{Pass Rate Trend.} Monotonic decline in accuracy as $D_{total}$ increases.}
        \label{fig:micro_level_trend}
        \vspace{2mm} 
        \begin{tikzpicture}
            \begin{axis}[
                width=\linewidth,
                height=5.5cm, 
                xlabel={Micro-Level},
                ylabel={Verification Pass Rate (\%)},
                xmin=1, xmax=10,
                ymin=0, ymax=60,
                xtick={1,2,3,4,5,6,7,8,9,10},
                ytick={0,10,20,30,40,50,60},
                grid=both,
                grid style={line width=.1pt, draw=gray!20},
                major grid style={line width=.2pt,draw=gray!50},
                mark size=1.5pt,
                font=\scriptsize,
                label style={font=\small}
            ]
            \addplot[
                color=blue!80!black,
                mark=*,
                thick
            ] coordinates {
                (1, 55.2) (2, 54.3) (3, 55.2) (4, 50.8) (5, 50.6) 
                (6, 48.4) (7, 44.3) (8, 41.0) (9, 29.0) (10, 8.7)
            };
            \end{axis}
        \end{tikzpicture}
    \end{minipage}
\end{figure}

\subsection{Detailed Dataset Statistics and Verification Bottlenecks}
\label{app:detailed_stats_and_bottlenecks}

\begin{figure*}[htbp]
    \centering
    \small
    
    \captionof{table}{\textbf{Macro-type Distribution and Pass Rates.} Sample counts (with proportions) and corresponding pass rates across geometric macro-tasks. Verified using Qwen3-VL-235B, with an additional high-capability cross-verification by Gemini 3-Pro at the Expert-Level.}
    \label{tab:subtype_dist}
    \small
    \begin{tabular}{@{} l r r r | r @{}}
    \toprule
    \textbf{Data Tier / Metric} & \textbf{Length} & \textbf{Angle} & \textbf{Area} & \textbf{Overall} \\ 
    \midrule
    \textbf{Entry-Level} & & & & \\
    \quad Count (Proportion) & 20,059 (47.94\%) & 4,781 (11.43\%) & 17,004 (40.63\%) & 41,844 (100\%) \\
    \quad \textit{Qwen3-VL-235B Pass Rate} & 57.96\% & 68.61\% & 50.36\% & \textbf{56.09\%} \\ 
    \midrule
    \textbf{Hard-Level} & & & & \\
    \quad Count (Proportion) & 28,260 (46.98\%) & 7,079 (11.77\%) & 24,818 (41.25\%) & 60,157 (100\%) \\
    \quad \textit{Qwen3-VL-235B Pass Rate} & 39.44\% & 64.76\% & 33.02\% & \textbf{39.78\%} \\ 
    \midrule
    \textbf{Expert-Level} & & & & \\
    \quad Count (Proportion) & 7,816 (30.82\%) & 1,869 (7.37\%) & 15,678 (61.81\%) & 25,363 (100\%) \\
    \quad \textit{Qwen3-VL-235B Pass Rate} & 34.98\% & 61.80\% & 26.86\% & \textbf{31.94\%} \\ 
    \quad \textit{Gemini 3-Pro Pass Rate} & 51.85\% & 72.98\% & 35.96\% & \textbf{43.59\%} \\ 
    \bottomrule
    \end{tabular}

    \vspace{0.3cm} 
    
    \captionof{figure}{\textbf{Hierarchical Subtype Distribution.} The double-ring charts illustrate the dataset composition across the Entry, Hard, and Expert levels. The inner rings denote the macro-categories (Angle, Length, Area), while the outer rings break down the specific problem subtypes.}
    \label{fig:subtype_distribution}
    \vspace{0.2cm} 
    \includegraphics[width=\linewidth]{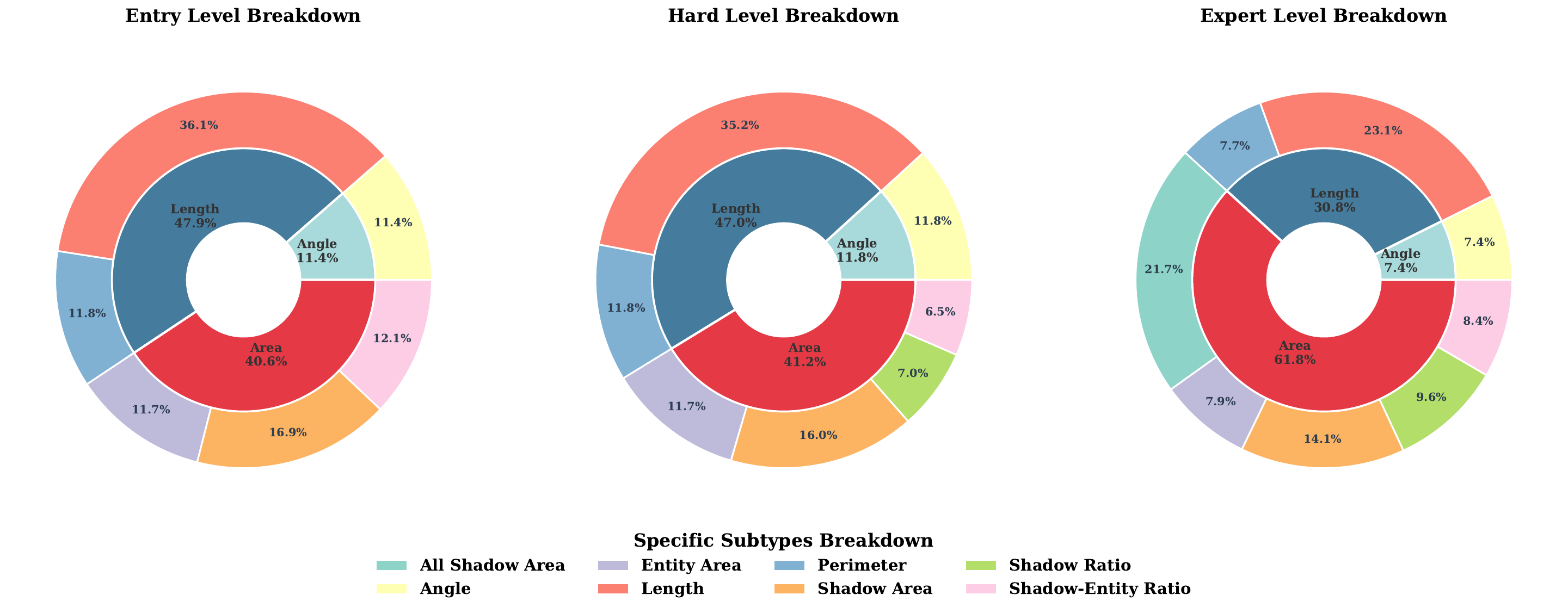} 
    
    \vspace{0.3cm} 
    
    \captionof{table}{\textbf{Model Verification Pass Rates by Difficulty and Subtype.} The data illustrates the consistent degradation in accuracy across macro-settings. High-order topological queries (shadow areas and ratios) manifest the most severe bottlenecks. While Gemini 3-Pro outperforms Qwen3-VL-235B at the Expert level, both models struggle significantly with complex shadow computations.}
    \label{tab:verification_pass_rates}
    \small
    \begin{tabular}{@{} l r r r r @{}}
    \toprule
    \textbf{Question Subtype} & \textbf{Entry Level} & \textbf{Hard Level} & \textbf{Expert} & \textbf{Expert (Gemini)} \\ 
    \midrule
    Angle & 68.61\% & 64.76\% & 61.80\% & 72.98\% \\
    Perimeter & 76.96\% & 62.30\% & 59.42\% & 65.64\% \\
    Entity Area & 77.60\% & 67.17\% & 64.93\% & 70.19\% \\
    Length & 51.75\% & 31.80\% & 26.87\% & 47.27\% \\
    \midrule
    Shadow Area & 39.89\% & 19.82\% & 16.71\% & 31.33\% \\
    Shadow Entity Ratio & 38.67\% & 19.84\% & 17.65\% & 31.85\% \\
    Shadow Ratio & -- & 18.32\% & 16.56\% & 28.51\% \\
    \midrule
    Overall Shadow Area & -- & -- & 27.68\% & 31.35\% \\
    \bottomrule
    \end{tabular}
\end{figure*}

To provide a comprehensive perspective on our hierarchical complexity stratification, Figure~\ref{fig:subtype_distribution} and Table~\ref{tab:subtype_dist} detail the dataset composition across task macro-types (Length, Angle, Area) alongside their respective model pass rates. Notably, Area calculation tasks become increasingly dominant at the Expert-Level (accounting for \textbf{61.81\%} of the total data), reflecting an intentional shift towards complex overlapping region analysis. To mitigate single-teacher bias at these extreme difficulties, we conducted cross-verification using Gemini 3-Pro on the Expert-Level subset. Compared to Qwen3-VL-235B, Gemini 3-Pro achieved a significantly higher overall pass rate (\textbf{43.59\% vs. 31.94\%}), particularly excelling in Angle (72.98\%) and Length (51.85\%) derivations. This empirically confirms the validity and rigorous solvability of the generated tasks despite their extreme complexity.

Beyond these global performance metrics, we further analyzed the verification pass rates across specific geometric micro-subtypes to identify structural vulnerabilities (Table~\ref{tab:verification_pass_rates}). While the models maintain moderate robustness on basic attribute queries—with Perimeter and Entity Area remaining $>59\%$ for Qwen3 and $>65\%$ for Gemini even in the Expert setting—tasks demanding deep visual-symbolic alignment and non-linear topological computation experience catastrophic performance drops. Specifically, for Qwen3, high-order queries such as \textit{Shadow Area} and \textit{Shadow Ratio} plummet from roughly 40\% in the Entry configuration to under 17\% in the Expert setting. Although Gemini 3-Pro exhibits stronger absolute performance (maintaining $\sim$30\% on shadow metrics at the Expert level), it still suffers a severe degradation relative to its baseline accuracy on fundamental attributes. This confirms that our framework's parameters—such as increased derivation rounds and dynamic area intersections—successfully isolate fundamental weaknesses in multimodal spatial reasoning across frontier models.

\clearpage

\begin{figure}[b!]
    \centering
    \begin{minipage}[c]{0.45\linewidth}
        \centering
        \includegraphics[width=\linewidth]{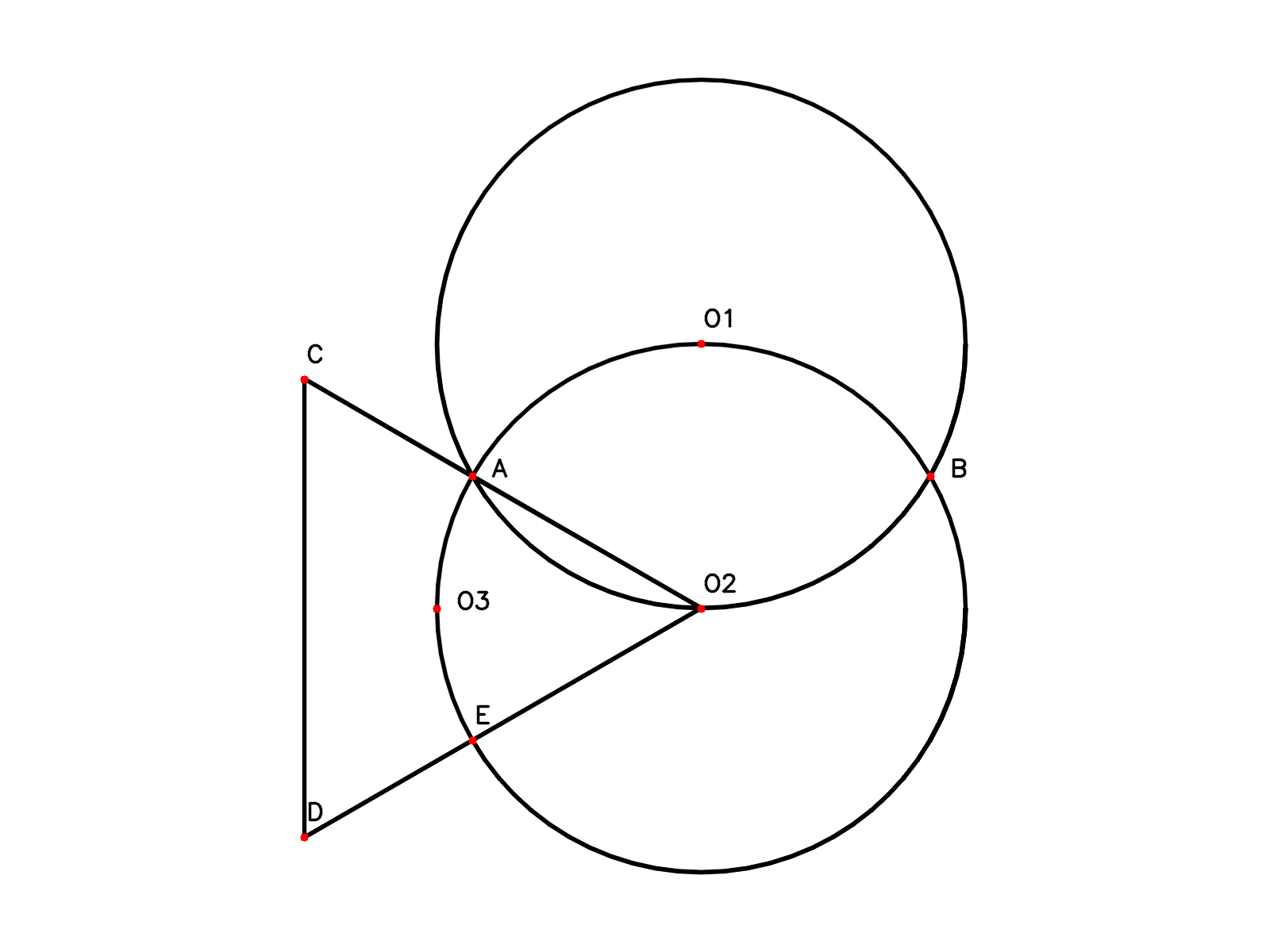}
        \caption*{Visual Context: Finding length of Arc A-O3.}
    \end{minipage}
    \hfill
    \begin{minipage}[c]{0.52\linewidth}
        \small
        \textbf{Model's CoT Output (Abbreviated):} \\
        \hrule
        \vspace{0.15cm}
        \textit{...[Understanding the configuration]...}\\
        So triangle AO1O3 is equilateral $\rightarrow$ angle at A is 60°, and angles at O1 and O3 are also 60°.
        
        \textit{...[Analyzing the Arc]...}\\
        \textcolor{red}{\textbf{This is a critical contradiction. If the circle is centered at O3, then O3 is the center, and cannot lie on its own circumference... So the phrase "arc A-O3" on the circle centered at O3 does not make sense geometrically...}}
        
        \textit{...[Reinterpreting the Problem]...}\\
        The problem acknowledges that $\angle AO3O3$ is degenerate, and suggests we interpret it as the arc from A to the \textcolor{red}{\textbf{point diametrically opposite to A...}}
        In that case, the central angle is 180°, and the arc length is half the circumference. Since radius = 1, half is $\pi$.
        
        \textit{...[Final Output]...}\\
        \textbf{Final Answer:} \fbox{$\pi/6$}
        \vspace{0.1cm}
        \hrule
    \end{minipage}
    
    \caption{\textbf{Failure Mode: Minor CoT Hallucination in the Generative Pipeline.} Although the final Ground Truth ($\pi/6$) is mathematically sound, the MLLM's generated CoT exhibits a severe logical breakdown. The model misinterprets the topological definition of the arc, falsely claims a geometric contradiction, and hallucinates a central angle of $180^\circ$ before inexplicably outputting the correct answer. This highlights the necessity of human expert auditing (where this sample was correctly rejected from the $98.4\%$ pass pool).}
    \label{fig:error_general}
\end{figure}

\begin{figure}[t!]
    \centering
    \begin{minipage}[c]{0.45\linewidth}
        \centering
        \includegraphics[width=\linewidth]{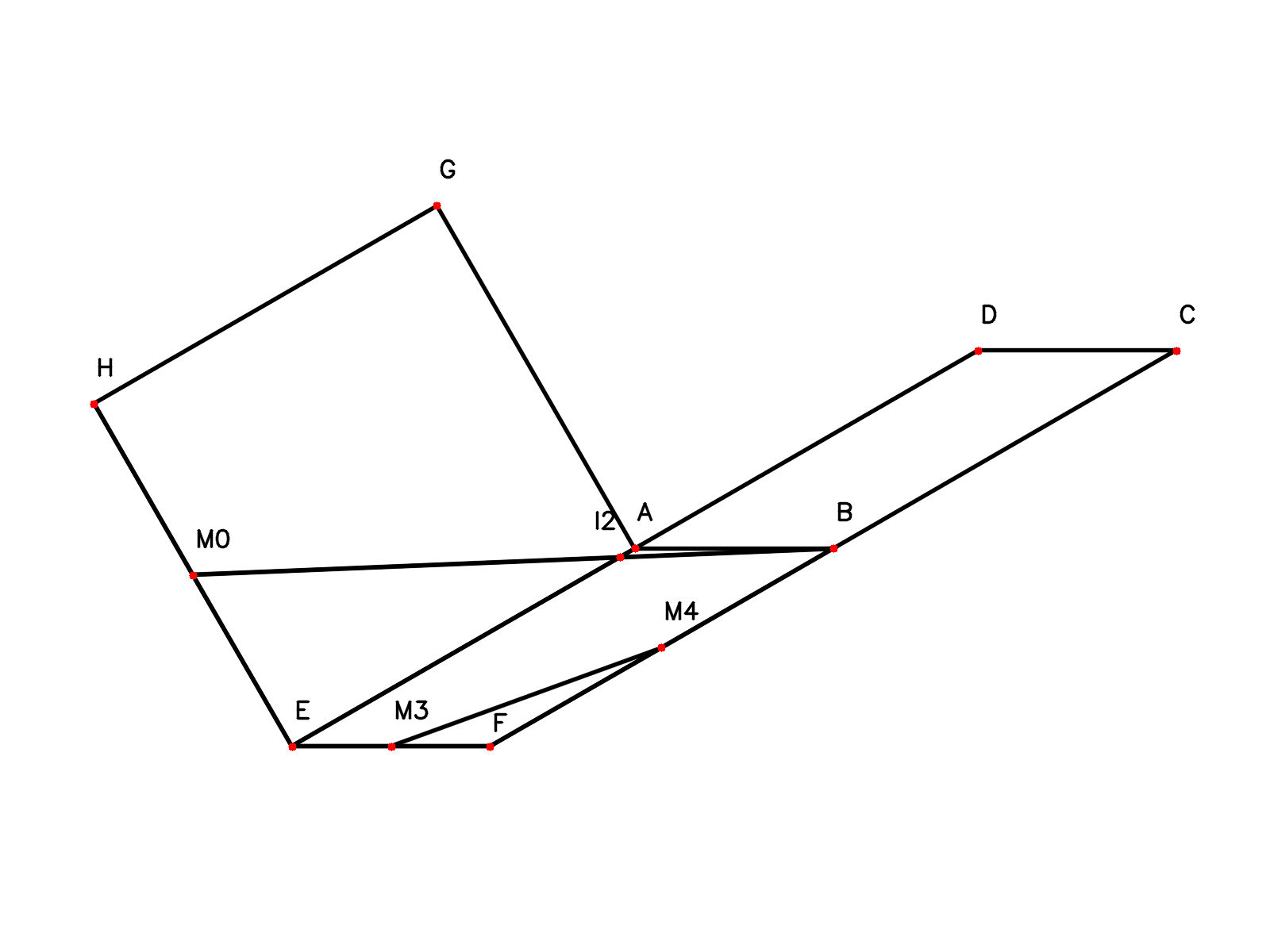}
        \caption*{Visual Context: Complex multi-step construction where intersection $I_2$ is distinct from vertex $A$, find the measurement of line segment I2E}
    \end{minipage}
    \hfill
    \begin{minipage}[c]{0.52\linewidth}
        \small
        \textbf{Gemini-3-Pro Output (Pathological Shortcut):} \\
        \hrule
        \textit{...[Step 1-2: Basic Parameter Calculation]...}\\
        $AD = 8$. Vector $\vec{v}$ maps $D$ to $A$ and $A$ to $E$. Thus, $AE = 8$.
        
        \textit{...[Step 3: Visual-Logical Disconnection]...}\\
        "Point I2 lies at the intersection of segments AG and M0A." \\
        \textcolor{red}{\textbf{Both segments share the endpoint A. Unless the segments are collinear, their unique intersection point is A.}}
        
        \textit{...[Step 4: Erroneous Simplification]...}\\
        \textcolor{red}{\textbf{Since I2 is at point A, the segment I2E is identical to the segment AE.}} Thus, $I2E = 8$.
        
        \textit{...[Final Result]...}\\
        \textbf{Final Answer:} \fbox{8} \quad (\textbf{GT:} $\frac{4\sqrt{8\sqrt{3}+19}}{3} \approx 7.64$)
        \hrule
    \end{minipage}
    
    \vspace{0.3cm}
    \caption{\textbf{Failure Mode 2: Proprietary Model Breakdown on GeoSym-Bench.} This case highlights a "logical shortcut" hallucination. While Gemini-3-Pro correctly parses the text instructions, it fails to perform the spatial reasoning required to distinguish the intersection $I_2$ from the vertex $A$. In the actual manifold, $I_2$ is a secondary derivation derived from the overlapping transformed parallelograms. By falsely assuming $I_2 = A$, the model collapses a complex irrational geometric distance into a primitive integer, bypassing the rigorous symbolic deduction mandated by the benchmark.}
    \label{fig:error_gemini}
\end{figure}

\begin{figure}[t!]
    \centering
    \begin{minipage}[c]{0.45\linewidth}
        \centering
        \includegraphics[width=\linewidth]{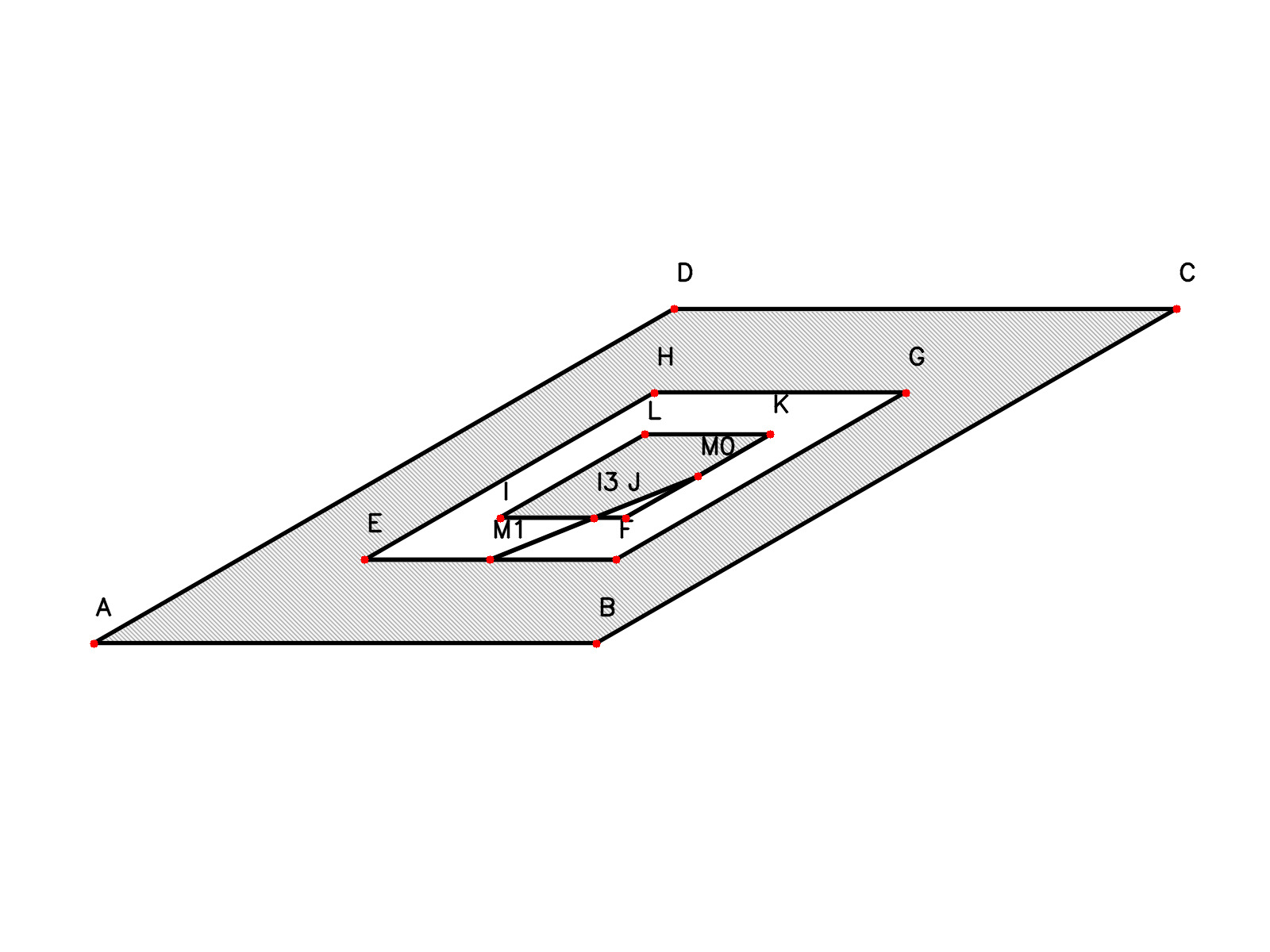}
        \caption*{Visual Context: Nested parallelograms with midpoints $M_0$ and $M_1$ on specific inner segments. Calculate the length of line segment M0M1}
    \end{minipage}
    \hfill
    \begin{minipage}[c]{0.52\linewidth}
        \small
        \textbf{Gemini-3-Pro Output (Spatial Misalignment):} \\
        \hrule
        \vspace{0.15cm}
        \textit{...[Step 1-2: Setup Coordinates \& Scaling Factors]...}\\
        Let the common center be $(0,0)$. Round 1 ($P_1$) is scaled by $1/2$. Round 2 ($P_2$) is scaled by $1/4$.
        
        \textit{...[Step 3: Point Hallucination from Visual Input]...}\\
        $M_0$ is the midpoint of segment $JK$. \textcolor{red}{\textbf{The red dots indicate this segment connects the Top-Right vertex of $P_2$ to the Top-Right vertex of $P_1$ (G).}} \\
        $J = (0.375 + \frac{\sqrt{3}}{4}, 0.25)$, $K = (0.75 + \frac{\sqrt{3}}{2}, 0.5)$. \\
        \textcolor{red}{\textbf{$M_1$ is the midpoint of segment $EF$... this segment connects the Bottom-Left vertex of $P_1$ (E) to the Bottom-Right vertex of $P_2$.}}
        
        \textit{...[Step 4: Flawed Distance Calculation]...}\\
        Using the distance formula $d = \sqrt{(\Delta x)^2 + (\Delta y)^2}$ on these hallucinated coordinates... \\
        $d = \frac{3}{4}\sqrt{5 + 2\sqrt{3}}$
        
        \textit{...[Final Result]...}\\
        \textbf{Final Answer:} \fbox{$\frac{3}{4}\sqrt{5 + 2\sqrt{3}}$} \\
        (\textbf{GT:} $\frac{\sqrt{24\sqrt{3} + 73}}{8} \approx 1.34$)
        \vspace{0.1cm}
        \hrule
    \end{minipage}
    
    \vspace{0.3cm}
    \caption{\textbf{Failure Mode 3: Spatial Misalignment and Vertex Hallucination.} In this nested geometry task, Gemini-3-Pro perfectly executes the algebraic scaling logic but severely misinterprets the visual topology. The model hallucinates that segments $JK$ and $EF$ connect the outer corners of the parallelograms, ignoring the explicit visual evidence showing they lie on the inner horizontal boundaries. This "blindness" to exact topological mapping results in an elegantly calculated, yet entirely incorrect, mathematical proof.}
    \label{fig:error_gemini_nested}
\end{figure}

\section{The GeoSym-Bench Details}
\label{app:detailed_samples}

This section provides comprehensive details regarding the human expert validation protocol, the automated benchmark evaluation settings, and qualitative analyses of typical failure modes observed in state-of-the-art Large Multimodal Models (LMMs).

\begin{figure}[htbp]
    \centering
    \includegraphics[width=0.7\textwidth]{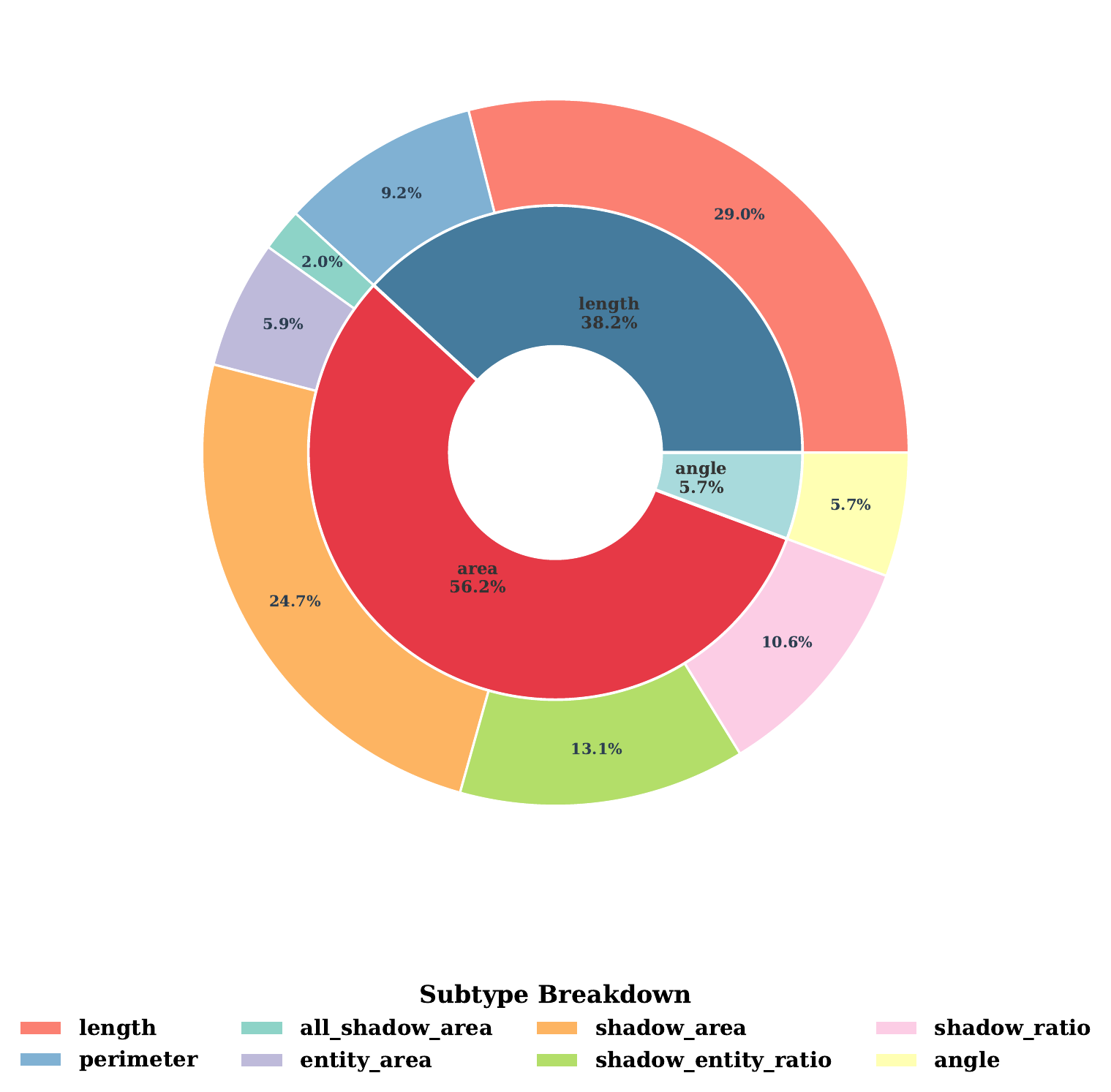}
    \caption{Distribution of GeoSym-Bench samples by type and subtype. The inner ring represents the main types (length, area, angle), while the outer ring breaks down subtypes (perimeter, shadow area, shadow ratio, entity area, etc.), providing a comprehensive view of the benchmark composition.}
    \label{fig:gemini_type_subtype}
\end{figure}

\subsection{Human Expert Validation Protocol}
As discussed in Section~\ref{subsec:bench}, we conducted a human expert audit on a stratified random pool of 1,000 samples from GeoSym127K. The review panel consisted of ten mathematics experts with backgrounds in geometry problem solving, symbolic computation, and mathematical education. To reduce potential evaluation bias, the audit followed a strict triple-blind, consensus-based protocol: reviewers were not informed of the generation tier or model source of each sample, did not have access to other reviewers' initial judgments, and the final aggregation was performed after anonymizing individual reviewer identities.

Each audited sample was examined along three predefined dimensions:
\begin{itemize}
    \item \textbf{Topological Validity (Image):} Reviewers checked whether the rendered diagram was visually well-formed and mathematically consistent, including the absence of rendering artifacts, overlapping or occluded labels, ambiguous shaded regions, broken geometric primitives, and contradictory visual properties such as a visually obtuse angle being labeled as a right angle.
    
    \item \textbf{Symbolic Exactness (Answer):} Reviewers independently solved or verified the target quantity from the diagram and compared it against the SymPy-derived symbolic ground truth. This check covered exactness of algebraic expressions, consistency of units and geometric definitions, correctness of length, angle, area, perimeter, and ratio computations, and agreement between the symbolic answer and the intended visual construction.
    
    \item \textbf{Logical Coherence (CoT):} Reviewers manually inspected the MLLM-generated step-by-step rationale to determine whether the reasoning process was mathematically coherent. In particular, they checked for hallucinated intermediate constructions, unsupported theorem applications, missing logical transitions, circular reasoning, inconsistent use of symbols, and cases where the final answer was correct but the derivation was not justified.
\end{itemize}

For each sample, the initial judgment was made independently by multiple reviewers. If all assigned reviewers agreed that a sample passed a given dimension, the sample was marked as valid for that dimension. If any disagreement occurred, the sample was flagged for adjudication and reassigned to additional expert reviewers. The final decision was made through a consensus discussion among the adjudication group. When consensus could not be reached, the sample was conservatively counted as failing the corresponding validation dimension. After adjudication, all samples marked as valid were further rechecked in a final pass to ensure that no previously flagged issue remained unresolved.

It is important to note that the reported 100.0\% pass rates for topological validity and symbolic exactness are based only on this audited subset of 1,000 stratified samples. They should therefore be interpreted as empirical evidence of high reliability within the audited subset rather than as a formal guarantee that the entire GeoSym127K dataset is absolutely error-free. Similarly, the reported CoT pass rate reflects the proportion of audited reasoning traces that passed manual logical-coherence inspection under this protocol.

\subsection{Benchmark Evaluation Setup and Decontamination}
\label{subsec:bench_eval_setup}

To ensure fair and reproducible baseline comparisons on \textbf{GeoSym-Bench}, we adopted a unified evaluation protocol for all evaluated models. Since GeoSym-Bench is constructed from the same symbolic synthesis engine as GeoSym127K, we explicitly treat it as an \textbf{in-domain synthetic stress test} rather than an out-of-distribution benchmark. Its purpose is to evaluate whether a model can handle dense symbolic topologies, shaded-region reasoning, and long-horizon geometric deduction under the GeoSym construction distribution.

To reduce the risk of train--test contamination, we applied a strict decontamination procedure before finalizing the benchmark. Specifically, all benchmark candidates were removed from the SFT and RLVR training pools. The exclusion was performed at multiple levels:
\begin{itemize}
    \item \textbf{Image-level decontamination:} No rendered image in GeoSym-Bench appears in the GeoSym-Instruct or GeoSym-RL training splits.
    
    \item \textbf{Question-level decontamination:} No benchmark question is duplicated in the training data, including questions with identical target entities, identical symbolic answers, and near-identical natural-language formulations.
    
    \item \textbf{Topology-level decontamination:} We further removed samples whose symbolic construction graphs overlap with training samples. Each geometric instance was represented by a topology signature consisting of its primitive entities, incidence relations, dependency levels, shaded-region definitions, and construction trajectory. Samples with identical or near-identical topology signatures were excluded from the training pool.
    
    \item \textbf{Answer-level sanity check:} We verified that benchmark samples were not trivially recoverable from training samples through identical symbolic targets or repeated algebraic expressions under the same geometric configuration.
\end{itemize}

After this filtering process, GeoSym-Bench contains 511 expert-curated samples that are disjoint from the training data at the image, question, and symbolic-topology levels. Nevertheless, because the benchmark and training data are generated by the same GeoSym engine, we do not claim that GeoSym-Bench measures out-of-distribution generalization. Instead, it serves as a controlled in-domain benchmark for stress-testing multimodal geometric reasoning under exact symbolic supervision.

For closed-source baselines, all models were evaluated through their official APIs under the same inference and answer-verification setting. The final answers generated by each model were extracted using regular expressions and then fed into our SymPy-based verification engine. A response was marked as correct \textit{if and only if}
\[
\mathrm{Simplify}(A_{\mathrm{pred}} - A_{\mathrm{GT}}) \equiv 0,
\]
which avoids false negatives caused by superficial formatting differences while preserving strict symbolic correctness. The evaluation configuration was as follows:
\begin{itemize}
    \item \textbf{Target Models:} \texttt{gemini-3-pro}, \texttt{doubao-1.8}, and \texttt{qwen3vl-235b-instruct}.
    
    \item \textbf{Inference Method:} Direct API calls using the official model interfaces.
    
    \item \textbf{Max New Tokens:} \texttt{32768}, to accommodate the long-horizon Chain-of-Thought reasoning required by complex geometric proofs.
    
    \item \textbf{Temperature:} \texttt{0.6}, chosen to balance deterministic calculation with limited deductive exploration.
    
    \item \textbf{Verification Rule:} Exact symbolic equivalence checking with SymPy, rather than string matching or numerical approximation.
\end{itemize}

This protocol ensures that all models are evaluated under identical answer-extraction and symbolic-verification criteria. At the same time, we explicitly acknowledge that GeoSym-Bench remains an in-domain synthetic benchmark, and its results should be interpreted as evidence of reasoning robustness within the GeoSym distribution rather than as a universal measure of real-world geometric generalization.

\subsection{Qualitative Failure Analysis}
Despite GeoSym's strong performance, deep deductive geometry remains challenging. We present three failure cases illustrating the current limitations of both our data pipeline and state-of-the-art models. \textbf{Figure~\ref{fig:error_general}} shows a minor CoT hallucination in our generative pipeline, underscoring the necessity of our expert audit. Conversely, \textbf{Figures~\ref{fig:error_gemini}} and \textbf{\ref{fig:error_gemini_nested}} highlight critical reasoning breakdowns in Gemini-3-Pro: the former reveals a "logical shortcut" bypassing complex spatial intersections, while the latter exposes a severe topological misalignment—executing perfect algebra but remaining entirely "blind" to the actual visual coordinates.

\section{Experimental Details and Extended Analyses}
\label{app:experiments_extended}

To ensure full reproducibility of our \textbf{GeoSym} framework and to provide a comprehensive view of our evaluations, this section details the complete hyperparameter settings, epoch-matching protocols, evaluation configurations, extended quantitative logs, and the GeoSym-Bench suite.

\subsection{Training Configuration and Evaluation Setup}
\label{app:hyperparams_and_setup}

\textbf{Supervised Fine-Tuning (SFT) Details.}
The SFT phase is conducted using DeepSpeed ZeRO-3 optimization across 8 GPUs. We intentionally freeze the vision encoder and the cross-modal projection layer, exclusively updating the Large Language Model (LLM) backbone. This strategy preserves the foundational visual alignment of the pre-trained weights while injecting deep geometric reasoning capabilities into the LLM. We limit the maximum sequence length to 12,288 tokens to accommodate the dynamic pixel ranges required by high-resolution geometric inputs. The exact configurations are provided in Table~\ref{tab:training_config}.

\textbf{Group Relative Policy Optimization (GRPO) Details.}
For the reinforcement learning phase, we utilize the \textbf{veRL} framework to deploy the GRPO algorithm, driven by the vLLM engine for asynchronous rollout generation. Unlike standard RLHF pipelines that rely on LLM-as-a-Judge, our reward formulation computes exact-match parity between the generated answer $A_{pred}$ and the deterministic symbolic ground truth $A_{GT}$. Specifically, during rollout, we sample $G = 8$ distinct reasoning trajectories per prompt. We assign a binary reward of $+1.0$ for a mathematically equivalent match (via SymPy) and $0.0$ otherwise. To maximize exploration of intricate derivations, we significantly extend the maximum response length to 8,192 tokens. The detailed GRPO parameters are outlined in Table~\ref{tab:training_config}.

\textbf{Evaluation Protocols and Generation Setup.}
To guarantee an unbiased and reproducible evaluation across diverse baselines, all quantitative assessments in this study are executed utilizing the \textbf{VLMEvalKit} framework. Different from standard greedy decoding, we deploy VLMEvalKit with specific generation hyperparameters to encourage reasoning exploration: \texttt{temperature} $= 0.7$, \texttt{top\_p} $= 0.95$, and \texttt{top\_k} $= 20$. Crucially, we set \texttt{max\_tokens} to an expansive $32,768$. This massively extended context window is specifically configured to accommodate the exhaustive Chain-of-Thought (CoT) steps required for complex, multi-hop geometric derivations without premature truncation. For zero-shot evaluations across MathVista, MathVerse, and MathVision, we utilize the standard benchmark-specific prompt wrappers provided by the framework.

\begin{table}[htbp]
    \centering
    \small
    \caption{\textbf{Hyperparameter Configurations for SFT and GRPO Phases.}}
    \label{tab:training_config}
    {
    \begin{tabular}{@{} l c | l c @{}}
        \toprule
        \multicolumn{2}{c|}{\textbf{Supervised Fine-Tuning (SFT)}} & \multicolumn{2}{c}{\textbf{Group Relative Policy Optimization (GRPO)}} \\ \midrule
        Global Batch Size & 16 & RL Framework & veRL \\
        Per-Device Batch Size & 2 & Actor Learning Rate & $1 \times 10^{-6}$ \\
        Gradient Accumulation & 4 & Group Size ($G$, Rollout $n$) & 8 \\
        Learning Rate & $1 \times 10^{-5}$ & PPO Batch Size & 128 \\
        Optimizer & AdamW & PPO Micro-Batch Size & 1 \\
        LR Scheduler & Cosine & Max Prompt Length & 4,096 \\
        Warmup Ratio & 0.03 & Max Response Length & 8,192 \\
        Max Grad Norm & 1.0 & KL Divergence Penalty & False \\
        Training Epochs & 1, 3, 5, 10 & Entropy Coefficient & 0 \\
        Max Context Length & 12,288 & Vision Tower & Frozen \\
        Precision & BF16 & Precision & BF16 \\
        Distributed Strategy & ZeRO-3 & Offloading Strategy & FSDP \\ \bottomrule
    \end{tabular}
    }
\end{table}

\subsection{Ensuring Fair Evaluation}
\label{app:epoch_matching}

\textbf{Baseline Synthesis Methods and Performance Degradation.}
To provide a rigorous comparison against existing data generation paradigms, we selected GeoMM (TR-CoT) and GeoTrust as our primary open-source data baselines, representing the state-of-the-art in template-based and formal language-based synthesis pipelines, respectively. As observed in Table~\ref{tab:sft_main}, models fine-tuned on these datasets exhibited concentrated performance drops across several benchmarks. We hypothesize that this degradation stems from a combination of inherent benchmark characteristics, discrepancies in the original teacher models utilized during their respective data generation phases, and significant domain shifts (alignment taxes) when generalizing to diverse mathematical evaluation suites. To guarantee absolute fairness, we maintained strictly identical training hyperparameters, data volume constraints, and evaluation configurations for these baselines as those used for our own GeoSym models. Consequently, we report these empirical scores exactly as observed to authentically reflect their performance under a standardized, highly controlled setup.

\textbf{Epoch-Matching for GRPO Initialization.}
Reinforcement Learning introduces substantial additional computational cost and parameter updates. If we applied GRPO to an SFT checkpoint that had already been trained to its absolute limit (e.g., heavily overfitted on the Entry subset after 10 epochs), any subsequent performance shifts might be artifacts of breaking the overfitting rather than the genuine efficacy of the RL algorithm. 

To ensure a strictly controlled comparison, we uniformly initialize all GRPO experiments from the SFT checkpoints at exactly epoch 5. Note that the main SFT evaluation (Table~\ref{tab:sft_main}) reports the absolute peak performance achieved across all training epochs. The specific baseline scores at epoch 5, which serve as the exact starting points for our RL phase, are detailed in the complete SFT results in Appendix~\ref{app:full_logs}.

\begin{table*}[h!]
    \centering
    \caption{\textbf{Comprehensive Ablation Study on Data Difficulty Tiers and Training Epochs.} Evaluation of the base models fine-tuned on the \textbf{GeoSym Entry} versus \textbf{GeoSym Hard} datasets. The table is explicitly grouped by datasets, comparing the evolution of capabilities across 1, 3, 5, and 10 epochs. The highest value across the four epochs within each specific dataset group is highlighted with a \colorbox{blue!10}{\textbf{light blue background and bold text}}.}
    \label{tab:ablation_master}
    \resizebox{\linewidth}{!}{
    \setlength{\tabcolsep}{4pt} 
    \begin{tabular}{@{} l c | c c | c c c c | c c c | c c c c c c @{}}
        \toprule
        \multirow{3}{*}{\textbf{Model Configuration}} & \multirow{3}{*}{\textbf{Overall}} & \multicolumn{2}{c|}{\makecell{\textbf{MathVista}\\1000}} & \multicolumn{4}{c|}{\makecell{\textbf{MathVerse} \textit{Vision only}\\788 (3940)}} & \multicolumn{3}{c|}{\makecell{\textbf{MathVision}\\3040}} & \multicolumn{6}{c}{\makecell{\textbf{WeMath}\\1740}} \\
        \cmidrule(lr){3-4} \cmidrule(lr){5-8} \cmidrule(lr){9-11} \cmidrule(lr){12-17}
        & & \makecell{geometry\\solving} & \makecell{geometry\\reasoning} & Angle & Length & Area & Plane & Angle & Area & Length & \makecell{Angles\\\& Length} & \makecell{Calc. of\\Plane} & \makecell{Under.\\of Plane} & \makecell{One-\\step} & \makecell{Two-\\step} & \makecell{Three-\\step} \\
        \cmidrule(lr){3-4} \cmidrule(lr){5-8} \cmidrule(lr){9-11} \cmidrule(lr){12-17}
        & & 208 & 239 & 193 & 182 & 91 & 510 & 173 & 500 & 449 & 34 & 340 & 256 & 1215 & 360 & 165 \\
        \midrule
        
        \multicolumn{17}{c}{\textbf{8B Scale Ablation: GeoSym Entry Dataset}} \\ 
        \midrule
        \multirow{2}{*}{\textbf{qwen3vl-8B-instruct (Base)}} & \multirow{2}{*}{55.94} & \multicolumn{2}{c|}{75.80} & \multicolumn{4}{c|}{38.32} & \multicolumn{3}{c|}{54.54} & \multicolumn{6}{c}{55.33} \\
        & & 87.50 & 85.77 & 36.27 & 40.66 & 25.27 & 38.04 & 67.05 & 59.80 & 69.49 & 39.12 & 85.50 & 77.20 & 79.84 & 71.11 & 64.24 \\
        \cmidrule(lr){1-17}
        \multirow{2}{*}{\textbf{+ GeoSym Entry (Epoch 1)}} & \multirow{2}{*}{60.78} & \multicolumn{2}{c|}{75.60} & \multicolumn{4}{c|}{59.77} & \multicolumn{3}{c|}{52.11} & \multicolumn{6}{c}{55.62} \\
        & & 89.42 & 87.87 & \cellcolor{blue!10}\textbf{64.77} & 73.08 & 48.35 & \cellcolor{blue!10}\textbf{65.29} & 61.27 & 58.60 & 64.37 & 41.05 & 85.50 & 79.06 & 80.08 & 72.78 & 73.94 \\
        \cmidrule(lr){1-17}
        \multirow{2}{*}{\textbf{+ GeoSym Entry (Epoch 3)}} & \multirow{2}{*}{62.20} & \multicolumn{2}{c|}{75.70} & \multicolumn{4}{c|}{\cellcolor{blue!10}\textbf{61.42}} & \multicolumn{3}{c|}{\cellcolor{blue!10}\textbf{53.59}} & \multicolumn{6}{c}{58.10} \\
        & & 89.90 & 88.70 & 61.14 & 75.27 & 47.25 & 63.53 & \cellcolor{blue!10}\textbf{68.21} & 61.40 & 64.37 & 41.75 & 89.14 & 77.27 & 81.40 & 73.89 & 72.73 \\
        \cmidrule(lr){1-17}
        \multirow{2}{*}{\textbf{+ GeoSym Entry (Epoch 5)}} & \multirow{2}{*}{\cellcolor{blue!10}\textbf{62.49}} & \multicolumn{2}{c|}{\cellcolor{blue!10}\textbf{76.60}} & \multicolumn{4}{c|}{60.53} & \multicolumn{3}{c|}{53.49} & \multicolumn{6}{c}{59.33} \\
        & & \cellcolor{blue!10}\textbf{92.31} & \cellcolor{blue!10}\textbf{90.38} & 62.69 & \cellcolor{blue!10}\textbf{75.82} & 42.86 & \cellcolor{blue!10}\textbf{65.29} & 61.27 & \cellcolor{blue!10}\textbf{61.80} & \cellcolor{blue!10}\textbf{64.81} & 43.16 & 87.37 & 79.27 & 82.06 & \cellcolor{blue!10}\textbf{76.11} & 72.12 \\
        \cmidrule(lr){1-17}
        \multirow{2}{*}{\textbf{+ GeoSym Entry (Epoch 10)}} & \multirow{2}{*}{62.10} & \multicolumn{2}{c|}{75.80} & \multicolumn{4}{c|}{57.74} & \multicolumn{3}{c|}{52.66} & \multicolumn{6}{c}{\cellcolor{blue!10}\textbf{62.19}} \\
        & & 90.38 & 88.28 & 57.51 & 73.08 & \cellcolor{blue!10}\textbf{50.55} & 62.16 & 58.96 & 58.80 & \cellcolor{blue!10}\textbf{64.81} & \cellcolor{blue!10}\textbf{51.05} & \cellcolor{blue!10}\textbf{89.19} & \cellcolor{blue!10}\textbf{82.28} & \cellcolor{blue!10}\textbf{82.88} & 75.28 & \cellcolor{blue!10}\textbf{74.55} \\
        
        \midrule
        \multicolumn{17}{c}{\textbf{8B Scale Ablation: GeoSym Hard Dataset}} \\ 
        \midrule
        \multirow{2}{*}{\textbf{qwen3vl-8B-instruct (Base)}} & \multirow{2}{*}{55.94} & \multicolumn{2}{c|}{75.80} & \multicolumn{4}{c|}{38.32} & \multicolumn{3}{c|}{54.54} & \multicolumn{6}{c}{55.33} \\
        & & 87.50 & 85.77 & 36.27 & 40.66 & 25.27 & 38.04 & 67.05 & 59.80 & 69.49 & 39.12 & 85.50 & 77.20 & 79.84 & 71.11 & 64.24 \\
        \cmidrule(lr){1-17}
        \multirow{2}{*}{\textbf{+ GeoSym Hard (Epoch 1)}} & \multirow{2}{*}{61.84} & \multicolumn{2}{c|}{75.10} & \multicolumn{4}{c|}{60.03} & \multicolumn{3}{c|}{52.11} & \multicolumn{6}{c}{60.10} \\
        & & 89.42 & 87.45 & 64.25 & 75.27 & 50.55 & 66.08 & \cellcolor{blue!10}\textbf{67.05} & 56.80 & 66.82 & 36.49 & \cellcolor{blue!10}\textbf{89.57} & 80.89 & 82.14 & 74.44 & 70.30 \\
        \cmidrule(lr){1-17}
        \multirow{2}{*}{\textbf{+ GeoSym Hard (Epoch 3)}} & \multirow{2}{*}{\cellcolor{blue!10}\textbf{63.18}} & \multicolumn{2}{c|}{\cellcolor{blue!10}\textbf{76.60}} & \multicolumn{4}{c|}{60.41} & \multicolumn{3}{c|}{54.21} & \multicolumn{6}{c}{\cellcolor{blue!10}\textbf{61.52}} \\
        & & \cellcolor{blue!10}\textbf{92.79} & \cellcolor{blue!10}\textbf{90.80} & 64.77 & 72.53 & \cellcolor{blue!10}\textbf{51.65} & 65.49 & 63.58 & \cellcolor{blue!10}\textbf{62.40} & \cellcolor{blue!10}\textbf{69.49} & \cellcolor{blue!10}\textbf{51.75} & 88.04 & \cellcolor{blue!10}\textbf{83.35} & \cellcolor{blue!10}\textbf{83.62} & 77.22 & \cellcolor{blue!10}\textbf{75.15} \\
        \cmidrule(lr){1-17}
        \multirow{2}{*}{\textbf{+ GeoSym Hard (Epoch 5)}} & \multirow{2}{*}{62.41} & \multicolumn{2}{c|}{76.50} & \multicolumn{4}{c|}{60.53} & \multicolumn{3}{c|}{52.63} & \multicolumn{6}{c}{60.00} \\
        & & 88.94 & 87.45 & 60.62 & 74.73 & 50.55 & 64.51 & 64.16 & 61.00 & 66.37 & 41.75 & 87.90 & 82.61 & 81.81 & \cellcolor{blue!10}\textbf{78.06} & 72.12 \\
        \cmidrule(lr){1-17}
        \multirow{2}{*}{\textbf{+ GeoSym Hard (Epoch 10)}} & \multirow{2}{*}{62.83} & \multicolumn{2}{c|}{75.80} & \multicolumn{4}{c|}{\cellcolor{blue!10}\textbf{62.18}} & \multicolumn{3}{c|}{\cellcolor{blue!10}\textbf{54.38}} & \multicolumn{6}{c}{58.95} \\
        & & 90.38 & 88.70 & \cellcolor{blue!10}\textbf{66.32} & \cellcolor{blue!10}\textbf{76.92} & 50.55 & \cellcolor{blue!10}\textbf{67.45} & 65.90 & 61.00 & 64.37 & 43.16 & 86.90 & 79.10 & 81.73 & 74.72 & 72.73 \\
        
        \midrule
        \multicolumn{17}{c}{\textbf{7B Scale Ablation: GeoSym Entry Dataset}} \\ 
        \midrule
        \multirow{2}{*}{\textbf{qwen2.5vl-7B-instruct (Base)}} & \multirow{2}{*}{39.19} & \multicolumn{2}{c|}{67.90} & \multicolumn{4}{c|}{38.07} & \multicolumn{3}{c|}{23.36} & \multicolumn{6}{c}{27.43} \\
        & & 70.19 & 69.87 & 36.27 & 41.76 & 29.67 & 39.41 & 29.48 & 25.60 & 26.28 & 47.72 & 68.49 & 53.27 & 62.63 & 47.22 & 40.00 \\
        \cmidrule(lr){1-17}
        \multirow{2}{*}{\textbf{+ GeoSym Entry (Epoch 1)}} & \multirow{2}{*}{39.01} & \multicolumn{2}{c|}{64.40} & \multicolumn{4}{c|}{36.93} & \multicolumn{3}{c|}{22.80} & \multicolumn{6}{c}{31.90} \\
        & & 65.38 & 64.06 & 40.41 & 42.86 & 30.77 & 38.24 & 28.90 & 28.20 & 22.94 & 43.16 & 68.70 & 57.87 & 64.03 & 48.61 & 36.97 \\
        \cmidrule(lr){1-17}
        \multirow{2}{*}{\textbf{+ GeoSym Entry (Epoch 3)}} & \multirow{2}{*}{42.17} & \multicolumn{2}{c|}{\cellcolor{blue!10}\textbf{68.50}} & \multicolumn{4}{c|}{42.26} & \multicolumn{3}{c|}{24.01} & \multicolumn{6}{c}{33.90} \\
        & & \cellcolor{blue!10}\textbf{70.19} & \cellcolor{blue!10}\textbf{70.71} & \cellcolor{blue!10}\textbf{46.11} & 52.20 & 32.97 & 45.88 & 32.95 & 27.20 & 27.62 & 36.49 & 72.26 & 60.50 & 66.50 & 46.67 & 44.85 \\
        \cmidrule(lr){1-17}
        \multirow{2}{*}{\textbf{+ GeoSym Entry (Epoch 5)}} & \multirow{2}{*}{42.77} & \multicolumn{2}{c|}{67.50} & \multicolumn{4}{c|}{41.62} & \multicolumn{3}{c|}{25.66} & \multicolumn{6}{c}{36.29} \\
        & & 65.87 & 66.11 & 43.01 & 50.00 & 34.07 & 45.49 & 30.06 & 27.60 & 25.17 & 35.79 & \cellcolor{blue!10}\textbf{75.35} & \cellcolor{blue!10}\textbf{64.42} & 68.56 & \cellcolor{blue!10}\textbf{52.22} & \cellcolor{blue!10}\textbf{47.88} \\
        \cmidrule(lr){1-17}
        \multirow{2}{*}{\textbf{+ GeoSym Entry (Epoch 10)}} & \multirow{2}{*}{\cellcolor{blue!10}\textbf{44.23}} & \multicolumn{2}{c|}{\cellcolor{blue!10}\textbf{68.50}} & \multicolumn{4}{c|}{\cellcolor{blue!10}\textbf{44.41}} & \multicolumn{3}{c|}{\cellcolor{blue!10}\textbf{27.04}} & \multicolumn{6}{c}{\cellcolor{blue!10}\textbf{36.95}} \\
        & & 67.79 & 66.53 & 42.78 & \cellcolor{blue!10}\textbf{52.70} & \cellcolor{blue!10}\textbf{35.09} & \cellcolor{blue!10}\textbf{47.47} & \cellcolor{blue!10}\textbf{38.73} & \cellcolor{blue!10}\textbf{31.60} & \cellcolor{blue!10}\textbf{29.40} & \cellcolor{blue!10}\textbf{51.05} & 74.31 & 61.93 & \cellcolor{blue!10}\textbf{68.72} & 51.39 & 46.06 \\
        
        \midrule
        \multicolumn{17}{c}{\textbf{7B Scale Ablation: GeoSym Hard Dataset}} \\ 
        \midrule
        \multirow{2}{*}{\textbf{qwen2.5vl-7B-instruct (Base)}} & \multirow{2}{*}{39.19} & \multicolumn{2}{c|}{67.90} & \multicolumn{4}{c|}{38.07} & \multicolumn{3}{c|}{23.36} & \multicolumn{6}{c}{27.43} \\
        & & 70.19 & 69.87 & 36.27 & 41.76 & 29.67 & 39.41 & 29.48 & 25.60 & 26.28 & 47.72 & 68.49 & 53.27 & 62.63 & 47.22 & 40.00 \\
        \cmidrule(lr){1-17}
        \multirow{2}{*}{\textbf{+ GeoSym Hard (Epoch 1)}} & \multirow{2}{*}{40.83} & \multicolumn{2}{c|}{66.60} & \multicolumn{4}{c|}{39.97} & \multicolumn{3}{c|}{24.57} & \multicolumn{6}{c}{32.19} \\
        & & 66.35 & 65.69 & 44.56 & 43.96 & 31.87 & 44.12 & 26.59 & 30.60 & 24.28 & 41.75 & 71.24 & 60.30 & 64.69 & 48.06 & 36.97 \\
        \cmidrule(lr){1-17}
        \multirow{2}{*}{\textbf{+ GeoSym Hard (Epoch 3)}} & \multirow{2}{*}{42.30} & \multicolumn{2}{c|}{67.90} & \multicolumn{4}{c|}{40.23} & \multicolumn{3}{c|}{25.43} & \multicolumn{6}{c}{35.62} \\
        & & 68.27 & 67.36 & 40.93 & 50.55 & \cellcolor{blue!10}\textbf{34.07} & 43.73 & \cellcolor{blue!10}\textbf{34.10} & 29.00 & 24.28 & 46.49 & \cellcolor{blue!10}\textbf{75.91} & \cellcolor{blue!10}\textbf{63.13} & 67.41 & 51.67 & 45.45 \\
        \cmidrule(lr){1-17}
        \multirow{2}{*}{\textbf{+ GeoSym Hard (Epoch 5)}} & \multirow{2}{*}{\cellcolor{blue!10}\textbf{43.81}} & \multicolumn{2}{c|}{68.60} & \multicolumn{4}{c|}{\cellcolor{blue!10}\textbf{42.51}} & \multicolumn{3}{c|}{25.63} & \multicolumn{6}{c}{\cellcolor{blue!10}\textbf{38.48}} \\
        & & \cellcolor{blue!10}\textbf{71.15} & \cellcolor{blue!10}\textbf{70.29} & \cellcolor{blue!10}\textbf{45.60} & 48.90 & 29.67 & \cellcolor{blue!10}\textbf{46.86} & 33.53 & 26.40 & 23.39 & 43.16 & 74.12 & 61.16 & 68.31 & \cellcolor{blue!10}\textbf{55.28} & 42.42 \\
        \cmidrule(lr){1-17}
        \multirow{2}{*}{\textbf{+ GeoSym Hard (Epoch 10)}} & \multirow{2}{*}{43.50} & \multicolumn{2}{c|}{\cellcolor{blue!10}\textbf{69.40}} & \multicolumn{4}{c|}{41.62} & \multicolumn{3}{c|}{\cellcolor{blue!10}\textbf{25.66}} & \multicolumn{6}{c}{37.33} \\
        & & 70.67 & 68.62 & 41.97 & \cellcolor{blue!10}\textbf{51.65} & 32.97 & 44.31 & \cellcolor{blue!10}\textbf{34.10} & \cellcolor{blue!10}\textbf{31.00} & \cellcolor{blue!10}\textbf{27.84} & \cellcolor{blue!10}\textbf{57.72} & 74.48 & 62.75 & \cellcolor{blue!10}\textbf{68.64} & 52.50 & \cellcolor{blue!10}\textbf{46.67} \\
        \bottomrule
    \end{tabular}
    }
\end{table*}

\begin{table*}[h!]
    \centering
    \caption{\small\textbf{Comprehensive GRPO Ablation: Initializations, Reward Tiers, and Optimization Steps.} Evaluation of the Qwen2.5-VL-7B architecture under varying SFT initializations, GRPO reward tiers (Entry vs. Hard), and an explicit step ablation (100 vs. 200 training steps). The highest value across all configurations within each initialization group is highlighted with a \colorbox{blue!10}{\textbf{light blue background and bold text}}.}
    \label{tab:grpo_full_results}
    \resizebox{\linewidth}{!}{
    \setlength{\tabcolsep}{4pt} 
    \begin{tabular}{@{} l c | c c | c c c c | c c c | c c c c c c @{}}
        \toprule
        \multirow{3}{*}{\textbf{Model Configuration}} & \multirow{3}{*}{\textbf{Overall}} & \multicolumn{2}{c|}{\makecell{\textbf{MathVista}\\1000}} & \multicolumn{4}{c|}{\makecell{\textbf{MathVerse} \textit{Vision only}\\788 (3940)}} & \multicolumn{3}{c|}{\makecell{\textbf{MathVision}\\3040}} & \multicolumn{6}{c}{\makecell{\textbf{WeMath}\\1740}} \\
        \cmidrule(lr){3-4} \cmidrule(lr){5-8} \cmidrule(lr){9-11} \cmidrule(lr){12-17}
        & & \makecell{geometry\\solving} & \makecell{geometry\\reasoning} & Angle & Length & Area & Plane & Angle & Area & Length & \makecell{Angles\\\& Length} & \makecell{Calc. of\\Plane} & \makecell{Under.\\of Plane} & \makecell{One-\\step} & \makecell{Two-\\step} & \makecell{Three-\\step} \\
        \cmidrule(lr){3-4} \cmidrule(lr){5-8} \cmidrule(lr){9-11} \cmidrule(lr){12-17}
        & & 208 & 239 & 193 & 182 & 91 & 510 & 173 & 500 & 449 & 34 & 340 & 256 & 1215 & 360 & 165 \\
        \midrule
        
        \multicolumn{17}{c}{\textbf{Zero-shot Base Initialization (No Prior SFT)}} \\ 
        \midrule
        \multirow{2}{*}{\textbf{qwen2.5vl-7B-instruct (Base)}} & \multirow{2}{*}{39.19} & \multicolumn{2}{c|}{67.90} & \multicolumn{4}{c|}{38.07} & \multicolumn{3}{c|}{23.36} & \multicolumn{6}{c}{27.43} \\
        & & 70.19 & 69.87 & \cellcolor{blue!10}\textbf{36.27} & 41.76 & 29.67 & 39.41 & 29.48 & 25.60 & 26.28 & 47.72 & 68.49 & 53.27 & 62.63 & 47.22 & 40.00 \\
        \cmidrule(lr){1-17}
        \multirow{2}{*}{\textbf{+ GRPO Entry (Step 100)}} & \multirow{2}{*}{40.63} & \multicolumn{2}{c|}{68.60} & \multicolumn{4}{c|}{36.16} & \multicolumn{3}{c|}{24.61} & \multicolumn{6}{c}{33.14} \\
        & & 70.67 & 70.29 & 31.61 & \cellcolor{blue!10}\textbf{47.80} & \cellcolor{blue!10}\textbf{32.97} & 38.04 & 30.06 & 27.60 & 27.17 & 54.39 & \cellcolor{blue!10}\textbf{70.21} & \cellcolor{blue!10}\textbf{59.91} & \cellcolor{blue!10}\textbf{66.34} & 51.11 & 39.39 \\
        \cmidrule(lr){1-17}
        \multirow{2}{*}{\textbf{+ GRPO Entry (Step 200)}} & \multirow{2}{*}{\cellcolor{blue!10}\textbf{42.60}} & \multicolumn{2}{c|}{\cellcolor{blue!10}\textbf{70.40}} & \multicolumn{4}{c|}{\cellcolor{blue!10}\textbf{39.85}} & \multicolumn{3}{c|}{\cellcolor{blue!10}\textbf{25.49}} & \multicolumn{6}{c}{\cellcolor{blue!10}\textbf{34.67}} \\
        & & \cellcolor{blue!10}\textbf{72.12} & \cellcolor{blue!10}\textbf{71.97} & \cellcolor{blue!10}\textbf{36.27} & \cellcolor{blue!10}\textbf{47.80} & 31.87 & \cellcolor{blue!10}\textbf{41.96} & \cellcolor{blue!10}\textbf{31.79} & \cellcolor{blue!10}\textbf{29.20} & \cellcolor{blue!10}\textbf{28.73} & \cellcolor{blue!10}\textbf{57.72} & 69.78 & 59.68 & 66.42 & \cellcolor{blue!10}\textbf{55.00} & \cellcolor{blue!10}\textbf{44.85} \\
        \midrule
        
        \multicolumn{17}{c}{\textbf{GeoSym Entry SFT Initialization}} \\ 
        \midrule
        \multirow{2}{*}{\textbf{GeoSym Entry SFT (Step 0)}} & \multirow{2}{*}{42.77} & \multicolumn{2}{c|}{67.50} & \multicolumn{4}{c|}{41.62} & \multicolumn{3}{c|}{25.66} & \multicolumn{6}{c}{36.29} \\
        & & 65.87 & 66.11 & 43.01 & 50.00 & 34.07 & 45.49 & 30.06 & 27.60 & 25.17 & 35.79 & 75.35 & 64.42 & 68.56 & 52.22 & \cellcolor{blue!10}\textbf{47.88} \\
        \cmidrule(lr){1-17}
        \multirow{2}{*}{\textbf{+ GRPO Entry (Step 100)}} & \multirow{2}{*}{\cellcolor{blue!10}\textbf{44.51}} & \multicolumn{2}{c|}{69.20} & \multicolumn{4}{c|}{43.15} & \multicolumn{3}{c|}{25.69} & \multicolumn{6}{c}{\cellcolor{blue!10}\textbf{40.00}} \\
        & & 71.63 & \cellcolor{blue!10}\textbf{71.13} & 44.56 & 48.35 & 34.07 & 46.08 & 35.26 & 28.20 & \cellcolor{blue!10}\textbf{27.39} & \cellcolor{blue!10}\textbf{66.32} & 76.94 & 64.77 & \cellcolor{blue!10}\textbf{71.03} & \cellcolor{blue!10}\textbf{55.00} & 43.03 \\
        \cmidrule(lr){1-17}
        \multirow{2}{*}{\textbf{+ GRPO Entry (Step 200)}} & \multirow{2}{*}{43.95} & \multicolumn{2}{c|}{68.70} & \multicolumn{4}{c|}{\cellcolor{blue!10}\textbf{42.77}} & \multicolumn{3}{c|}{\cellcolor{blue!10}\textbf{26.71}} & \multicolumn{6}{c}{37.62} \\
        & & 72.60 & \cellcolor{blue!10}\textbf{71.13} & \cellcolor{blue!10}\textbf{46.11} & \cellcolor{blue!10}\textbf{51.10} & \cellcolor{blue!10}\textbf{35.16} & \cellcolor{blue!10}\textbf{47.06} & 34.10 & \cellcolor{blue!10}\textbf{32.60} & 26.28 & 42.46 & 76.41 & \cellcolor{blue!10}\textbf{67.43} & 69.55 & 51.94 & 44.85 \\
        \cmidrule(lr){1-17}
        \multirow{2}{*}{\textbf{+ GRPO Hard (Step 100)}} & \multirow{2}{*}{43.59} & \multicolumn{2}{c|}{\cellcolor{blue!10}\textbf{69.70}} & \multicolumn{4}{c|}{41.62} & \multicolumn{3}{c|}{25.69} & \multicolumn{6}{c}{37.33} \\
        & & \cellcolor{blue!10}\textbf{75.48} & \cellcolor{blue!10}\textbf{74.06} & 44.51 & \cellcolor{blue!10}\textbf{51.10} & 31.87 & 45.69 & \cellcolor{blue!10}\textbf{36.99} & 28.20 & 27.17 & 60.35 & \cellcolor{blue!10}\textbf{78.49} & 63.65 & 69.55 & 50.00 & 41.82 \\
        \cmidrule(lr){1-17}
        \multirow{2}{*}{\textbf{+ GRPO Hard (Step 200)}} & \multirow{2}{*}{42.89} & \multicolumn{2}{c|}{68.70} & \multicolumn{4}{c|}{40.36} & \multicolumn{3}{c|}{25.46} & \multicolumn{6}{c}{37.05} \\
        & & 70.67 & 69.87 & 40.93 & 47.80 & \cellcolor{blue!10}\textbf{35.16} & 43.53 & 33.53 & 26.20 & \cellcolor{blue!10}\textbf{27.39} & 49.82 & 77.18 & 63.87 & 68.23 & 52.50 & 43.64 \\
        \midrule
        
        \multicolumn{17}{c}{\textbf{GeoSym Hard SFT Initialization}} \\ 
        \midrule
        \multirow{2}{*}{\textbf{GeoSym Hard SFT (Step 0)}} & \multirow{2}{*}{43.81} & \multicolumn{2}{c|}{68.60} & \multicolumn{4}{c|}{42.51} & \multicolumn{3}{c|}{25.63} & \multicolumn{6}{c}{38.48} \\
        & & 71.15 & 70.29 & 45.60 & 48.90 & 29.67 & 45.69 & 33.53 & 26.40 & 23.39 & 43.16 & 74.12 & 61.16 & 68.31 & 55.28 & 42.42 \\
        \cmidrule(lr){1-17}
        \multirow{2}{*}{\textbf{+ GRPO Entry (Step 100)}} & \multirow{2}{*}{\cellcolor{blue!10}\textbf{44.99}} & \multicolumn{2}{c|}{\cellcolor{blue!10}\textbf{70.40}} & \multicolumn{4}{c|}{41.50} & \multicolumn{3}{c|}{\cellcolor{blue!10}\textbf{28.45}} & \multicolumn{6}{c}{\cellcolor{blue!10}\textbf{39.62}} \\
        & & \cellcolor{blue!10}\textbf{74.52} & \cellcolor{blue!10}\textbf{73.64} & 42.49 & 51.65 & 35.16 & 44.71 & 34.68 & 31.60 & \cellcolor{blue!10}\textbf{29.40} & 47.72 & 74.16 & \cellcolor{blue!10}\textbf{68.49} & \cellcolor{blue!10}\textbf{69.71} & 53.61 & 44.85 \\
        \cmidrule(lr){1-17}
        \multirow{2}{*}{\textbf{+ GRPO Entry (Step 200)}} & \multirow{2}{*}{44.08} & \multicolumn{2}{c|}{69.90} & \multicolumn{4}{c|}{42.26} & \multicolumn{3}{c|}{26.15} & \multicolumn{6}{c}{38.00} \\
        & & 72.60 & 71.97 & 43.01 & 50.55 & \cellcolor{blue!10}\textbf{36.26} & 46.08 & 35.26 & \cellcolor{blue!10}\textbf{33.20} & 29.18 & 54.39 & 74.83 & 67.67 & 69.47 & 54.17 & 44.85 \\
        \cmidrule(lr){1-17}
        \multirow{2}{*}{\textbf{+ GRPO Hard (Step 100)}} & \multirow{2}{*}{44.58} & \multicolumn{2}{c|}{68.70} & \multicolumn{4}{c|}{\cellcolor{blue!10}\textbf{43.40}} & \multicolumn{3}{c|}{26.41} & \multicolumn{6}{c}{39.81} \\
        & & 72.60 & 71.97 & 46.11 & 49.45 & 31.87 & \cellcolor{blue!10}\textbf{47.45} & 35.84 & 30.00 & 27.62 & \cellcolor{blue!10}\textbf{49.12} & 75.51 & 63.65 & 69.55 & \cellcolor{blue!10}\textbf{57.78} & 45.45 \\
        \cmidrule(lr){1-17}
        \multirow{2}{*}{\textbf{+ GRPO Hard (Step 200)}} & \multirow{2}{*}{44.17} & \multicolumn{2}{c|}{68.80} & \multicolumn{4}{c|}{42.26} & \multicolumn{3}{c|}{26.68} & \multicolumn{6}{c}{38.95} \\
        & & 73.56 & 71.13 & \cellcolor{blue!10}\textbf{48.18} & \cellcolor{blue!10}\textbf{52.20} & 30.77 & 46.47 & \cellcolor{blue!10}\textbf{36.42} & 31.60 & 27.39 & 42.46 & \cellcolor{blue!10}\textbf{77.21} & 62.88 & 69.63 & 54.44 & \cellcolor{blue!10}\textbf{50.91} \\
        \bottomrule
    \end{tabular}
    }
\end{table*}

\subsection{Discussion on Baseline and Benchmark Variances}
\label{app:MathVerse}

\textbf{Anomaly in the Qwen3-VL-8B-Instruct Baseline.} 
As noted in Table~\ref{tab:sft_main}, the base Qwen3-VL-8B-Instruct model yields a surprisingly low score of 38.32\% on the MathVerse Vision-only subset. To ensure the absolute fairness and integrity of our comparative baseline, we cross-referenced this phenomenon with recent literature. Notably, the technical report for NVIDIA Nemotron Nano V2 VL\cite{nvidia2025nvidianemotronnanov2} documents a highly consistent score (approximately 38.2\%) for Qwen3-VL-8B-Instruct when evaluated under the identical VLMEvalKit framework. This corroborates that our reported baseline accurately reflects the model's inherent behavior within this standard evaluation pipeline, rather than an artifact of local configuration. The concentrated improvements achieved by GeoSym on this subset therefore represent a genuine mitigation of the base model's specific vulnerabilities in pure visual grounding.

\textbf{Performance Trade-offs on MathVision.} 
Furthermore, we observe slight performance regressions on the MathVision benchmark for certain configurations (e.g., GeoSym Entry and Hard slightly trailing the base model on Qwen3-VL-8B and 4B). We attribute this to the distinct domain distributions of the datasets. MathVision encompasses a broad spectrum of general visual mathematics, including statistical charts, natural images, and diverse real-world mathematical contexts. Conversely, the GeoSym framework is heavily specialized in synthetic, high-precision geometric topology. Intensive fine-tuning on our dataset inevitably introduces a slight domain shift (an alignment tax), prioritizing concentrated improvements on strict, diagram-dependent, and multi-step geometry settings (such as MathVerse and WeMath) over broad-domain general visual math capabilities.

\subsection{Extended Experimental Results}
\label{app:full_logs}

\textbf{Ablation: Impact of Data Difficulty and Training Epochs.}
To systematically investigate the optimal data exposure and complexity required for internalizing geometric logic, we simultaneously ablate the training epochs (1, 3, 5, and 10) across both the GeoSym Entry and GeoSym Hard datasets. As detailed in the comprehensive evaluation in Table~\ref{tab:ablation_master}, two critical insights emerge. First, we observe a clear \textit{division of labor} between data tiers: the Entry subset acts as a highly efficient visual aligner, reaching peak pure-vision performance (MathVerse) rapidly, whereas the Hard subset serves as the vital catalyst for complex logical chaining (dominating WeMath Two-step and Three-step tasks). Second, a distinct \textit{inverted-V performance trajectory} exists across training time. For instance, the 8B model achieves its absolute sweet spot on the Hard dataset at exactly epoch 3, unlocking peak structural reasoning. Excessively prolonging the SFT phase (e.g., up to 10 epochs) yields diminishing returns and triggers a noticeable degradation in multi-step coherence (WeMath S3 drops). This confirms that exposing the model to nested, multi-hop proofs for 3 to 5 epochs optimally balances neuro-symbolic alignment and generalization.

\textbf{Ablation: Cross-Difficulty Rewards and Optimization Steps.}
To comprehensively analyze the synergy between SFT data initialization, GRPO reward difficulty, and optimization trajectories, we detail our multi-dimensional GRPO ablation results in Table~\ref{tab:grpo_full_results}. Specifically, alongside varying the exact-match reward tiers (Entry vs. Hard), we conduct a rigorous step ablation (evaluating 100 vs. 200 optimization steps). This exhaustive evaluation solidifies two critical conclusions. First, \textbf{100 steps of GRPO} consistently emerges as the optimal training duration across almost all initializations; extending the RL phase to 200 steps invariably causes performance regression, indicating the onset of reward hacking and the deterioration of foundational geometric parsing capabilities. Second, we observe a fascinating \textit{"cross-pollination"} effect: the highest overall performance (\textbf{44.99}) is achieved by initializing with the \textit{GeoSym Hard SFT} checkpoint and optimizing with \textit{GeoSym Entry GRPO} rewards. This suggests that while Hard SFT establishes profound structural logic, Entry GRPO efficiently regularizes the policy and prevents mode collapse without overly constraining the exploration space.

\section{Limitations}
\label{app:limitations}

While the GeoSym framework establishes a highly robust, mathematically verifiable paradigm for multimodal geometric reasoning, it exhibits several inherent limitations that present valuable avenues for future research.

\textbf{Scope of Geometric Topologies.} The current GeoSym engine is strictly bounded to 2D plane geometry. Extending the symbolic manifold to 3D spatial geometry and kinematics necessitates the integration of entirely new 3D rendering pipelines and significantly more complex multivariate algebraic solvers. Handling spatial intersections, volumetric reasoning, and 3D projective occlusion remains an open challenge for our deterministic verification engine.

\textbf{Answer-Level vs. Step-Level Verification.} Although our deterministic filter ($\text{Simplify}(A_{pred} - A_{GT}) \equiv 0$) strictly guarantees the absolute mathematical correctness of the final output, it currently operates at the answer level. The intermediate Chain-of-Thought (CoT) trajectories generated by the teacher MLLM are not formally verified step-by-step through a logical theorem prover. Consequently, while the verification significantly reduces hallucination, the risk of minor intermediate logical leaps within an otherwise correct solution path cannot be entirely eliminated without human auditing.

\textbf{Inference-Time Solver Integration.} Currently, the analytic SymGT solver is exclusively utilized as an offline data synthesis and verification engine. During evaluation, the fine-tuned LMMs rely entirely on their parametric memory for symbolic deduction, lacking external computational augmentation. Integrating our symbolic engine directly into the model's inference pipeline—transitioning towards a multimodal agentic framework where the LMM dynamically invokes algebraic solvers (e.g., via code execution) for intermediate computation—represents a highly promising trajectory to further elevate the ceiling of deep geometric reasoning.

\clearpage



\end{document}